\documentclass{article} %
\usepackage[preprint]{neurips_2026}

\usepackage[utf8]{inputenc}
\usepackage[T1]{fontenc}
\usepackage{url}
\usepackage{booktabs}
\usepackage{amsfonts}
\usepackage{amsmath}
\usepackage{amssymb}
\usepackage{nicefrac}
\usepackage{microtype}
\usepackage{xcolor}
\usepackage{graphicx}
\usepackage{tabularx}
\usepackage{algorithm}
\usepackage{algpseudocode}
\usepackage{amsthm}
\usepackage{float}
\usepackage{placeins}
\usepackage{hyperref}
\usepackage{bookmark}

\newtheorem{proposition}{Proposition}

\newcommand{\Hset}{\mathcal{H}}
\newcommand{\Sset}{\mathcal{S}}
\newcommand{\Aset}{\mathcal{A}}
\newcommand{\pimix}{\pi_{\text{mix}}}
\newcommand{\Dmix}{D_{\text{mix}}}
\newcommand{\fth}{f_\theta}
\newcommand{\rth}{r_\theta}
\newcommand{\Phipsi}{\Phi}
\newcommand{\Lrouter}{\mathcal{L}_{\text{router}}}

\newcommand{\xmark}{\ensuremath{\boldsymbol{\times}}}

\title{Inverse Reinforcement Learning with State-Dependent Foresight for Human Decision-Making}

\author{
    \textbf{Jinil Kim}$^{1,2,3}$
    \quad
    \textbf{Jinwoo Jeong}$^{2,4}$
    \quad
    \textbf{Hyeongjun Yun}$^{5}$
    \\
    \textbf{Min-hwan Oh}$^{5}$
    \quad
    \textbf{Woo-Young Ahn}$^{2,6,7}$
    \\[2pt]
    \normalfont
    $^{1}$Dept of Computer Science and Engineering, Seoul National University
    \\
    \normalfont
    $^{2}$Dept of Psychology, Seoul National University
    \\
    \normalfont
    $^{3}$Dept of Linguistics, Seoul National University
    \\
    \normalfont
    $^{4}$School of Computer Science, Georgia Institute of Technology
    \\
    \normalfont
    $^{5}$Graduate School of Data Science, Seoul National University
    \\
    \normalfont
    $^{6}$Dept of Brain and Cognitive Sciences, Seoul National University
    \\
    \normalfont
    $^{7}$AI Institute, Seoul National University
    \\[2pt]
    \normalfont\texttt{\{haba6030, jwjeong16, yunst97, minoh, wahn55\}@snu.ac.kr}
}

\begin{document}
\maketitle

\begin{abstract}Computational modeling of behavior represents individuals with a model fitted to their observed behavior and characterizes them through its parameters.
Naturalistic tasks incorporate the complexity of real environments and yield high-dimensional behavioral trajectories.
Inverse reinforcement learning (IRL) infers the latent reward function underlying these trajectories.
Standard IRL, however, assumes effectively infinite-horizon planning, whereas human foresight is both bounded and state-dependent.
Here we introduce \textbf{SC-AIRL} (\textbf{S}tate-\textbf{C}onditional Adversarial IRL), which builds these two cognitive properties into the planner.
Per-depth policies share one reward but differ in how far ahead they sum it explicitly.
A state-conditional router mixes them according to the current state.
Across held-out episodes in pedestrian crossing and highway overtaking tasks, SC-AIRL improves imitation over Standard AIRL, raising top-1 accuracy from 0.549 to 0.647 and from 0.629 to 0.718, respectively, while more closely reproducing participant behavior in rollouts. Inferred planning depth correlates with self-reported anxiety and motor impulsivity. SC-AIRL thus reproduces human behavior more closely than Standard AIRL while providing an interpretable, model-derived measure of state-dependent foresight.
\end{abstract}

\section{Introduction}
\label{sec:intro}

Computational models characterize individuals through model parameters estimated from their observed choices. The parameters quantify latent constructs such as impulsivity and reward sensitivity~\citep{montague2012computational, huys2016computational}. This approach has largely been developed in controlled laboratory tasks. Yet parameters estimated from such tasks often correlate weakly with self-report questionnaires and show limited stability across sessions~\citep{enkavi2019testretest, dang2020weakly, karvelis2023individual}. Naturalistic reinforcement learning has been proposed to address this limitation by incorporating the complexity of real environments into task design~\citep{wise2024naturalistic}. In these tasks, participants make decisions unconstrained in time, and these decisions form high-dimensional behavioral trajectories~\citep{mobbs2021ethology}. For example, \citet{lee2024bridging} recorded behavioral trajectories in a real-time driving task and showed that task performance correlated more strongly with self-reported impulsivity than traditional laboratory measures of impulsivity did. In the high-dimensional state space of such tasks, however, the reward is difficult to define for every state~\citep{ng2000irl, arora2021survey}. Characterizing individual differences in these tasks therefore motivates inferring reward functions from observed behavior.

Inverse reinforcement learning (IRL) infers the reward function that best explains observed trajectories~\citep{ng2000irl, ziebart2008maxent, fu2018airl} and has been applied to behavioral data from mice and humans~\citep{ashwood2022dirl, felso2022planningdepth, lee2024bridging}. IRL assumes that behavior is generated by a reward and a planning process that accumulates reward along a trajectory. The same behavior can be generated by multiple pairs of reward function and planning process~\citep{armstrong2018occam, skalse2023misspecification, skalse2025nonexponential}. Inferring the reward therefore requires an assumption about how the participant plans. Standard IRL assumes soft-optimal planning over the full horizon. Under this assumption, the participant sums reward over all future steps and selects each action with probability proportional to the exponential of that cumulative sum~\citep{ziebart2008maxent, fu2018airl}.

However, two properties of human planning conflict with this assumption. First, human foresight is bounded. Bounded rationality treats computational limits as a foundational premise~\citep{simon1955behavioral, lieder2020resource}, and \citet{mattar2022planning} suggest that planning in the brain is focused, depth-limited, and serial. Second, planning depth varies across states within the same individual. Time pressure can lead to shallower planning~\citep{keramati2016habits}, large losses can stop search along a branch~\citep{huys2012bonsai}, and people adjust planning depth based on task demands~\citep{eluchans2025planningdepth}. Planning depth also varies across individuals~\citep{felso2022planningdepth, opheusden2023nature}. Existing work has relaxed the standard planning assumption in two main ways. One line retains full-horizon planning and explains deviations through the discount rate~\citep{yao2024multiplehorizons} or through noise and bias in the choice rule~\citep{laidlaw2022bpd, beliaev2025irleed}. The other fixes a single depth parameter for an individual~\citep{felso2022planningdepth} or across all demonstrations~\citep{barnes2024rhip}. Jointly inferring the reward and a state-dependent planning depth from behavior remains underexplored.

Here we introduce \textbf{SC-AIRL} (\textbf{S}tate-\textbf{C}onditional AIRL), an adversarial IRL framework that infers a reward and a state-dependent depth of foresight from behavior in two stages (§\ref{sec:method}). In the first stage, SC-AIRL learns a single depth-invariant reward $r_\theta$ together with a family of per-depth policies $\{\pi^h\}_{h\in\Hset}$ that share it. Each policy $\pi^h$ is trained on this reward summed explicitly over $h$ steps, and value beyond these $h$ steps is estimated by a depth-specific critic.
In the second stage, a state-conditional router $p_\phi(h|s)$ is fitted to the training data with the converged policies frozen.
The policy of SC-AIRL at each state is then a mixture of the per-depth policies, weighted by the probability $p_\phi(h|s)$ that the router assigns to each depth.
On held-out episodes of two real-time tasks, pedestrian crossing ($N{=}52$) and highway overtaking ($N{=}47$), SC-AIRL predicts participants' actions more accurately than Standard AIRL (§\ref{sec:exp-imitation}). 
The shared reward is consistent with the reward learned by Standard AIRL, and its action preferences differ from those of the policy it trains (§\ref{sec:exp-reward}). 
Finally, two measures derived from the inferred depth correlate with self-reported anxiety and motor impulsivity, respectively (§\ref{sec:exp-clinical}). 
Together, these results indicate that introducing the bounded, state-dependent structure of human planning into IRL improves imitation fidelity and yields depth measures associated with individual traits.

\section{Related Work}
\label{sec:related}

\paragraph{S1. Bounded planning horizon.}
Models that limit computation predict human choices more accurately than unconstrained models~\citep{binz2022resourcerational}.
In large-scale route prediction, IRL with a receding ten-step horizon predicts routes at least as well as an infinite horizon~\citep[RHIP,][]{barnes2024rhip}.
Other cognitive properties were likewise introduced into inverse models of human behavior.
Attentional biases have been inferred from behavior within inverse planning~\citep{banerjee2025attention}, and pedestrian crossing has been modeled as bounded-optimal control under noisy perception~\citep{wang2025pedestrian}.
Whereas RHIP fixes a single horizon, SC-AIRL lets the depth vary with the state within adversarial IRL.

\paragraph{S2. Discount adaptation in IRL.}
Prior IRL has represented planning depth through the discount rate.
\citet{yao2024multiplehorizons} infer a per-agent exponential discount together with a shared reward in tabular settings.
Exponential discount adaptation rescales an infinite-lookahead computation rather than truncating it.
In SC-AIRL, each planner $\pi^h$ instead sums the shared reward explicitly over at most $h$ steps and evaluates the remainder with its own learned critic, as in heuristic-search models of human planning~\citep{opheusden2023nature}. The depth $h$ varies with the state.
Proposition~\ref{prop:bounded-support} formalizes the difference between truncation and exponential discount rescaling.

\paragraph{S3. State-conditional routing and multi-policy imitation.}
Multi-policy imitation prevents averaging across behavior modes by gating several sub-policies.
InfoGAIL selects one sub-policy per trajectory through a latent code~\citep{li2017infogail}, whereas Information Maximizing Curriculum gates its sub-policies at each state~\citep{blessing2023imc}.
Mixture-of-experts imitation gates expert controllers by the recent state history~\citep{wang2025moeil}, and Option-GAIL switches sub-policies through a latent option variable inferred from the trajectory~\citep{jing2021optiongail}.
\citet{tian2021latentintelligence} infer a cognitive bound as a latent variable from demonstrations but hold it constant within each trajectory, a limitation the authors identify.
SC-AIRL routes over the planning depth $h$ with a state-conditional router $p_\phi(h|s)$, so the latent variable is cognitively motivated and varies within a trajectory.

\section{Preliminaries \& Problem Formulation}
\label{sec:prelim}

\paragraph{Preliminaries.}
We consider a Markov decision process (MDP)
$(\Sset,\Aset,P,r,\gamma)$ with state space $\Sset$, action space $\Aset$, transition kernel $P(s'|s,a)$, reward function $r:\Sset\!\times\!\Aset\!\to\!\mathbb{R}$, and discount $\gamma\in(0,1)$.
Expert trajectories $\tau=(s_0,a_0,s_1,\dots)$ are sampled from an unknown expert policy $\pi_E$.
We build on the AIRL framework of \citet{fu2018airl}, which recovers a reward function from expert demonstrations via a discriminator--generator game.
Under maximum-entropy optimality $\pi^\star(a|s)\propto\exp Q^\star(s,a)$ \citep{ziebart2008maxent}, the discriminator takes the form $D_\theta(s,a,s')=\sigma\!\big(\fth(s,a,s')-\log\pi(a|s)\big)$.
At convergence under deterministic dynamics, $\fth$ recovers a reward up to potential-based shaping \citep{fu2018airl,ng1999shaping}.
The modeling assumptions that extend this framework to bounded planning depth follow in §\ref{sec:assumptions}.
A full notation table is provided in Appendix~\ref{sec:appendix-notation}.

\subsection{Assumptions}
\label{sec:assumptions}

\paragraph{(A1) Shared reward.}
In a three-stage choice task, a weighted combination of strategies that accumulate the same reward to different planning depths fit human choices better than any single strategy~\citep{keramati2016habits}.
Accordingly, we assume that a set of depth-indexed policies $\{\pi^h\}_{h\in\Hset}$ shares a single reward $\rth$ without potential shaping ($\Phi\!\equiv\!0$; §\ref{sec:method-reward}), so the policies differ only in $h$.

\paragraph{(A2) Bounded foresight.}
In a sequential decision task, tree-search models with truncated planning horizons explained human choices better than a model with a full horizon~\citep{huys2012bonsai}.
Accordingly, we assume that each policy $\pi^h$ in (A1) is trained on the shared reward accumulated explicitly over at most $h$ steps, with $h$ in the finite depth set $\Hset$.
Value beyond these steps enters only through its own critic $V^h$ (§\ref{sec:method-policy}), as in a model of human planning with a limited depth that evaluates the rest with values learned from past outcomes~\citep{keramati2016habits}.

\paragraph{(A3) State-conditional depth.}
Planning depth varies with immediate needs within a single task and subject~\citep{mattar2022planning}.
Accordingly, we assume that each expert decision at state $s$ mixes the policies in (A1) with state-conditional weights $p(h|s)$, which serve as a proxy for the expert's planning depth.

\subsection{Problem: Imitation under state-dependent bounded foresight}
\label{sec:problem}

Given a dataset $\mathcal{D}=\{\tau_i\}_{i=1}^{N}$ of expert trajectories with per-state planning depths \emph{unobserved}, SC-AIRL infers per-depth policies $\{\pi^h\}_{h\in\Hset}$ and a shared reward $\rth$ via AIRL under Assumptions~(A1)--(A3), then fits a router $p_\phi(h|s)$ post-hoc on the converged policies.
Standard AIRL, whether infinite-horizon discounted or episode-terminated, assigns credit without truncation.
When experts plan with bounded foresight, its policy can therefore produce action frequencies that differ from the expert's.
SC-AIRL models a per-state bounded depth from demonstrations to reduce this mismatch (§\ref{sec:exp-imitation}).

\section{State-Conditional AIRL (SC-AIRL)}
\label{sec:method}

\begin{figure}[t]
  \centering
  \includegraphics[width=0.70\linewidth]{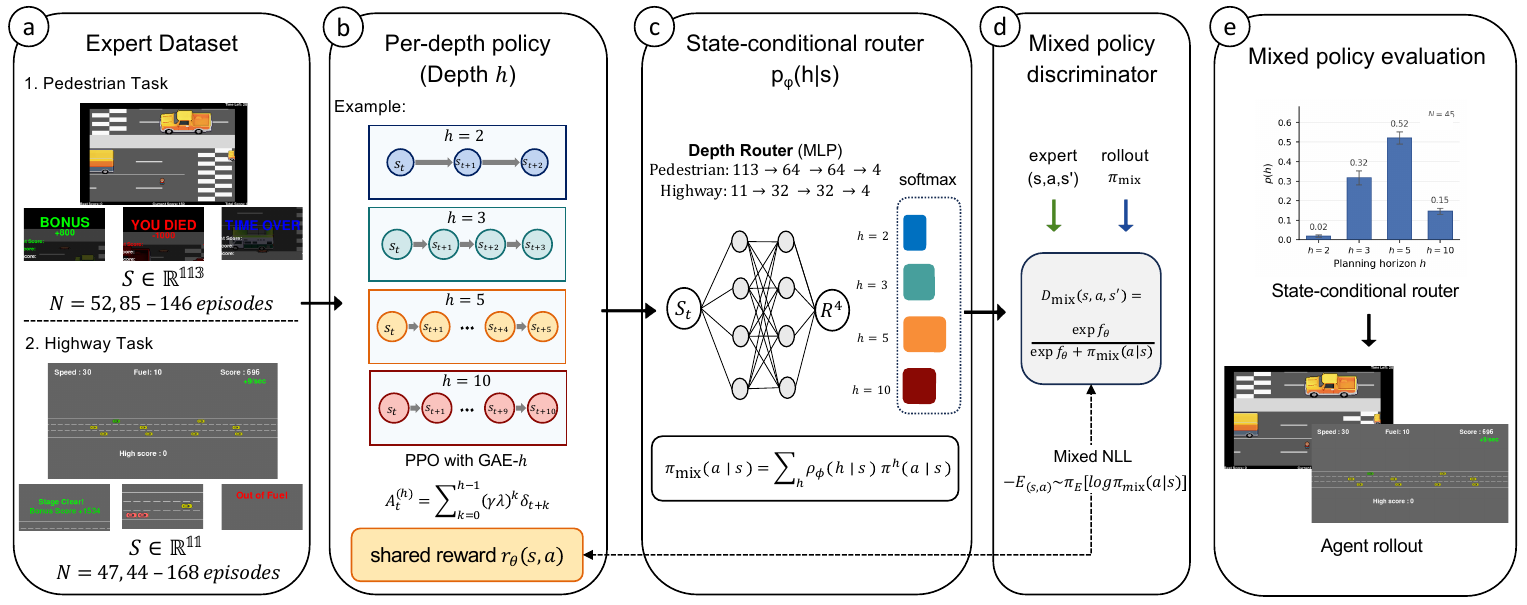}
  \caption{SC-AIRL pipeline. Expert transitions interact with the discriminator
    $\Dmix$ whose denominator is the mixed policy $\pimix$, assembled
    from per-depth actors $\{\pi^h\}$ via the state-dependent router
    $p_\phi(h|s)$. Per-depth policies and shared reward $\rth$ are
    trained via AIRL; the router is fit post-hoc on the converged
    policies (no-loop default; Appendix~\ref{sec:appendix-loop-validation}).
    Panel (b) shows the pedestrian depth grid; the highway grid is
    $\Hset=\{12,16,20,24\}$, or $2.4$--$4.8$~s at the task's $0.2$~s step.}
  \label{fig:system}
\end{figure}

\subsection{Overview}
\label{sec:method-overview}
SC-AIRL (Figure~\ref{fig:system}) departs from Standard AIRL~\citep{fu2018airl} along three structural replacements, each corresponding to one modeling assumption in §\ref{sec:assumptions} (Table~\ref{tab:three-axes}).
The replacements are detailed in §\ref{sec:method-reward}--§\ref{sec:method-dmix}.
Algorithm~\ref{alg:scairl} summarizes the training procedure.

\begin{table}[h]
  \centering
  \small
  \begin{tabular}{@{}p{2.2cm}p{4.2cm}p{6.7cm}@{}}
    \toprule
    & Standard AIRL & SC-AIRL \\
    \midrule
    No shaping    & $\fth = \rth + \gamma\Phipsi(s') - \Phipsi(s)$ & $\fth = \rth$; $\Phipsi\!\equiv\!0$; no $\gamma^h\Phi(s_{t+h})$ boundary term \\
    Explicit sum  & $G_t = \sum_{k=0}^{\infty}\gamma^k r$        & $G_t^{(h)} = \sum_{k=0}^{h-1}\gamma^k \rth$; beyond $h$: critic $V^h$ \\
    Multi-policy  & single $\pi$                                  & $\{\pi^h\}_{h\in\Hset}$ mixed by $p_\phi(h|s)$ \\
    \bottomrule
  \end{tabular}
  \caption{The three structural replacements in SC-AIRL.}
  \label{tab:three-axes}
\end{table}

\subsection{Depth-invariant reward parameterization}
\label{sec:method-reward}

\paragraph{Shared reward with no shaping.}
Standard AIRL \citep{fu2018airl} admits a potential-based shaping term $\fth(s,a,s')=\rth(s,a)+\gamma\Phi(s')-\Phi(s)$ that preserves the optimal policy \citep{ng1999shaping}. Summed along an $h$-step window, this contribution collapses to a pure boundary term at the truncation horizon,
\begin{equation}
  \sum_{k=0}^{h-1}\gamma^k\!\left[\fth(s_{t+k},a_{t+k},s_{t+k+1})
                                     - \rth(s_{t+k},a_{t+k})\right]
  \;=\; \gamma^h\,\Phi(s_{t+h}) - \Phi(s_t),
  \label{eq:telescope}
\end{equation}
which places a learned boundary value $\gamma^h\Phi(s_{t+h})$ inside the shared reward, and hence inside the explicit sum of every depth.
As (A2) requires, SC-AIRL keeps value beyond the window in the per-depth critics $V^h$ rather than in the shared reward, and therefore fixes
\begin{equation}
  \fth(s,a,s') \;=\; \rth(s,a), \qquad \Phi \equiv 0,
  \label{eq:shared-reward}
\end{equation}
so $\rth$ carries only the depth-invariant value signal.
All depth heterogeneity is absorbed by the planner family $\{\pi^h, V^h\}$ (§\ref{sec:method-policy}).

\subsection{Multi-policy architecture}
\label{sec:method-policy}

\paragraph{Per-depth policies.}
Under Assumption~(A2), we instantiate one per-depth policy $\pi^h_\omega$ with its own critic $V^h_\xi$, each trained with GAE-$h$, which limits explicit reward credit to $h$ steps.
The depths form a finite set $\Hset$ chosen per task (§\ref{sec:setup-data}).

\paragraph{$h$-step truncated return and GAE-$h$.}
For depth $h$, the return is the $h$-step discounted sum
\begin{equation*}
  G_t^{(h)} \;=\; \sum_{k=0}^{h-1}\gamma^k\,\rth(s_{t+k},a_{t+k}).
\end{equation*}
The per-depth critic $V^h_\xi$ is fitted to the GAE-$h$ return target by minimizing
\begin{equation}
  \mathcal{L}_{V}^h \;=\; \mathbb{E}_{\pi^h}\!\Bigl[\bigl(V^h(s_t) - \hat{R}_t^{(h)}\bigr)^2\Bigr],
  \quad \hat{R}_t^{(h)} \;=\; A_t^{(h)} + V^h_{\mathrm{old}}(s_t),
  \label{eq:critic-loss}
\end{equation}
and the advantage uses GAE-$h$, a truncated form of GAE~\citep{schulman2016gae}
\begin{align}
  \delta_t^{(h)} &\;=\; \rth(s_t,a_t) + \gamma\,V^h(s_{t+1}) - V^h(s_t),
  \label{eq:delta-h}\\
  A_t^{(h)}      &\;=\; \sum_{k=0}^{\min(h,T-t)-1}(\gamma\lambda)^k\,\delta_{t+k}^{(h)},
  \label{eq:gae-h}
\end{align}
with $\gamma=0.99$ and GAE parameter $\lambda=0.95$. Only the per-step TD residual $\delta_t^{(h)}$ bootstraps through $V^h$.
The advantage sum is truncated at $k=\min(h,T-t)-1$, so the reward $r_{t+k}$ enters $A_t^{(h)}$ with weight $(\gamma\lambda)^k\mathbf{1}[k<h]$ and explicit reward credit spans at most $h$ steps.
This kernel has bounded support and is therefore not a geometric discount rescaling (Proposition~\ref{prop:bounded-support}, Appendix~\ref{sec:appendix-theory}).
As in depth-limited search, $V^h$ approximates the value beyond the window. Since it is fitted from finite rollouts, $h$ sets how much of the advantage rests on observed reward rather than on $V^h$ (Appendix~\ref{sec:appendix-theory}).
In the fitted models, the per-depth policies act differently in both tasks (Appendices~\ref{sec:appendix-depth-profile} and~\ref{sec:appendix-highway-depth}).

\subsection{State-dependent depth router}
\label{sec:method-router}

\paragraph{Router overview.}
The router $p_\phi(h\mid s)$ maps a state $s\in\Sset$ to a distribution over planning depths $h\in\Hset$ and composes the per-depth family into the \emph{mixed policy}
\begin{equation}
  \pimix(a|s) \;=\; \sum_{h\in\Hset} p_\phi(h|s)\,\pi^h(a|s).
  \label{eq:pimix}
\end{equation}
It is fit by maximum likelihood on expert demonstrations:
\begin{equation}
  \Lrouter(\phi)
  \;=\;
  -\,\mathbb{E}_{(s,a)\sim\pi_E}\!\bigl[\log\pimix(a|s)\bigr]
  \;=\;
  -\,\mathbb{E}_{(s,a)\sim\pi_E}\!
  \biggl[\log\sum_{h\in\Hset} p_\phi(h|s)\,\pi^h(a|s)\biggr].
  \label{eq:router-nll}
\end{equation}
In the loop variants (Appendix~\ref{sec:appendix-loop-validation}), $p_\phi(h|s)$ also enters $\Dmix$ as the depth prior.

\paragraph{Router output distribution.}
We consider three parameterization families: \textbf{categorical} (unconstrained logits), \textbf{stick-breaking} (cumulative ordering), and \textbf{Gaussian kernel}
(unimodal over the depth index).
All share the same backbone (two hidden layers; Appendix~\ref{sec:appendix-router-families}).
Formulations and ablations are in Appendix~\ref{sec:appendix-router-variants}.

\subsection{Mixed-policy discriminator}
\label{sec:method-dmix}

The discriminator is trained against the mixture $\pimix$.
During training the mixture uses uniform weights (§\ref{sec:method-algo}).
Following the AIRL formulation \citep{fu2018airl}, the discriminator takes the log-ratio form
\begin{equation}
  \Dmix(s,a,s') \;=\;
     \frac{\exp \fth(s,a,s')}%
          {\exp \fth(s,a,s') + \pimix(a|s)},
  \label{eq:dmix}
\end{equation}
and is trained with a standard binary cross-entropy loss:
\begin{equation}
  \mathcal{L}_D \;=\;
  \mathbb{E}_{\pi_E}\!\bigl[-\log \Dmix\bigr]
  + \mathbb{E}_{\pimix}\!\bigl[-\log(1-\Dmix)\bigr].
  \label{eq:dmix-obj}
\end{equation}

\paragraph{Entropy bonus on the depth responsibility.}
We add an entropy bonus $-\lambda_{\text{ent}}\,\mathbb{E}_{\pi_E}[\mathbb{H}(q_D)]$ to Eq.~\ref{eq:dmix-obj}.
Here $q_D$ is the depth responsibility
\begin{equation}
  q_D(h\mid s,a) \;\propto\; p_\phi(h\mid s)\,\pi^h(a\mid s)\,D_h(s,a,s'),
  \label{eq:qD}
\end{equation}
where $D_h(s,a,s')=\exp\fth(s,a,s')\,/\!\bigl(\exp\fth(s,a,s')+\pi^h(a|s)\bigr)$ is the per-depth discriminator score.
Implementation details ($\lambda_{\text{ent}}$ and its effect on gradients) are in Appendix~\ref{sec:appendix-entropy}.

\subsection{Training algorithm}
\label{sec:method-algo}

SC-AIRL updates two parameter groups per AIRL round: reward $\rth$ and per-depth policy--value pairs $\{(\omega_h,\xi_h)\}_{h\in\Hset}$, with the router held at uniform $p_\phi(h|s)=1/|\Hset|$ throughout.
Training the reward and the router separately gives higher state-level imitation than training them jointly (Appendix~\ref{sec:appendix-loop-validation}).
Each round consists of one discriminator step and one PPO~\citep{schulman2017ppo} update per depth.
After AIRL convergence, the router $\phi$ is fit on the converged policies by maximizing $\log\pimix(a|s)$ at expert states.
We select the router fit length using a two-stage criterion based on imitation performance and depth-measure reliability.
First, we retain lengths whose NLL on a validation split (a random 20\% of training-split transitions withheld from the router fit) is below that of a uniform mixture of the same per-depth policies.
Second, among these, we select the length with the highest cross-seed reliability of the depth measures of \S\ref{sec:exp-clinical} ($I(s;h)$ and $\Delta h$ on the pedestrian task, $\Delta h$ on the highway task; ICC(A,k) across three training seeds).
The rule selects $500$ steps on the pedestrian task and $600$ steps on the highway task, and the reported routers are fitted at these lengths on the full training split.
Full pseudocode is in Appendix~\ref{sec:appendix-algo} (Algorithm~\ref{alg:scairl}).

\section{Experimental Setup}
\label{sec:setup}

\subsection{Naturalistic real-time pedestrian dataset}
\label{sec:setup-data}

\paragraph{Task environment.}
The first dataset comes from a pedestrian-crossing task, a naturalistic reinforcement learning task implemented in a Gymnasium-based simulator~\citep{towers2025gymnasium}.
An agent must reach the end point within a 30-second time limit.
The score rewards forward progress and finishing early, and penalizes collisions by severity.

\paragraph{State space, actions, and demonstrations.}
The state is $\mathbb{R}^{113}$ and varies within and between episodes as vehicle configurations, crosswalk presence, and other contextual factors change stochastically.
The action set is discrete $\Aset=\{\text{NOTHING},\text{UP},\text{DOWN}, \text{LEFT},\text{RIGHT}\}$.
We use $N{=}52$ participants (6 excluded).
The depth grid $\Hset=\{2,3,5,10\}$ spans about one lane width at $h{=}2$ and a wide lane block at $h{=}10$.
The state space, the score, the exclusion criteria, and the demonstration counts are detailed in Appendix~\ref{sec:appendix-data}.
A second grid, $\Hset=\{2,4,6,8,10\}$, is tested in Appendix~\ref{sec:appendix-uniform5}.

\subsection{Real-time highway overtaking dataset}
\label{sec:setup-highway}

\paragraph{Task environment.}
The second dataset is the public behavioral data from the highway task of \citet{lee2024bridging}, a real-time driving task built on \texttt{highway-env}~\citep{leurent2018highwayenv} and originally analyzed with AIRL.
Participants drive one car along a three-lane road, scoring at a rate that grows with the square of their speed.
An episode ends when the car collides, when the fuel runs out, or when the participant has overtaken every car on the road.

\paragraph{State space, actions, and demonstrations.}
The state is $\mathbb{R}^{11}$: own speed and lane, and the lead speed, lead gap, and rear gap of each of the three lanes.
The action set is $\Aset=\{\text{LEFT},\text{IDLE},\text{RIGHT},\text{FASTER},\text{SLOWER}\}$.
We use all $N{=}47$ participants of the original study.
The depth grid $\Hset=\{12,16,20,24\}$ spans $2.4$--$4.8$~s at the task's $0.2$~s step.
The state space, the score, and the demonstration counts are detailed in Appendix~\ref{sec:appendix-highway}.

\subsection{Common setting}
\label{sec:setup-common}

\paragraph{Held-out protocol.}
\label{sec:setup-holdout}
All pedestrian participants share $40$ map seeds, and all highway participants share $22$ traffic scenarios.
We hold out $8$ seeds and $4$ scenarios respectively, chosen by stratified selection within quartiles of expert success rate, balanced on episode length (and, on the highway task, on crash and fuel-out rates), and fixed before any evaluation.
The reward, per-depth policies, and router are trained on the remaining episodes only, and the router fit length is chosen on a validation split drawn from those episodes (\S\ref{sec:method-algo}).
All imitation metrics use the held-out episodes, and rollouts are run on the held-out seeds or scenarios (in-sample results in Appendix~\ref{sec:appendix-insample}).

\paragraph{Clinical battery.}
\label{sec:setup-battery}
Pedestrian participants completed the Delay Discounting Task (DDT), Choice Under Risk and Ambiguity (CRA), Barratt Impulsiveness Scale (BIS-11), Short UPPS-P Impulsive Behavior Scale, Center for Epidemiologic Studies Depression Scale (CES-D), and State-Trait Anxiety Inventory (STAI).
Highway participants completed the BIS-11, a delay-discounting task, and a go/no-go task~\citep{lee2024bridging}.
Full details of the pedestrian battery are in Appendix~\ref{sec:appendix-clinical}.

\subsection{Baseline models}
\label{sec:setup-baselines}

We compare against six learned baselines and one heuristic (Appendix Table~\ref{tab:baselines}).
A \textbf{TTC} heuristic following TTC-based pedestrian decision modeling~\citep{lobjois2007age} serves as a non-ML cognitive reference.
\textbf{BC} and \textbf{GAIL} imitate without learning a reward.
\textbf{Option-GAIL} adds state-conditional routing without reward learning or truncation.
\textbf{Standard AIRL} learns a reward with potential shaping over an infinite horizon, without truncation or routing.
\textbf{Single-$h$ AIRL} uses one per-depth policy of SC-AIRL alone, the depth with the lowest training-split NLL ($h{=}5$ on the pedestrian task and $h{=}12$ on the highway task).
\textbf{Episode-AIRL} uses the same per-depth policies but holds the depth weights fixed within each episode, so it matches SC-AIRL in parameter count and training cost and varies only the routing, per episode rather than per state.
Hyperparameters are listed in Appendix~\ref{sec:appendix-baseline-hp}.
SC-AIRL, Episode-AIRL, Single-$h$, Standard AIRL, BC, GAIL, and Option-GAIL are evaluated on three training runs per task (pedestrian: seeds $42$, $7$, and $123$; highway: seeds $7$ and $123$ and one run without a fixed seed). On the highway task, Standard AIRL has only the two seeded runs, and BC, GAIL, and Option-GAIL use seeds $42$, $7$, and $123$.

\section{Results}
\label{sec:experiments}

\subsection{Imitation fidelity}
\label{sec:exp-imitation}

We evaluate on held-out episodes of both tasks (§\ref{sec:setup-holdout}) at two levels (Table~\ref{tab:imitation}).
At the state level, we score the mixed policy on expert $(s,a)$ pairs by mixture negative log-likelihood (NLL; $-\log\sum_h p_\phi(h|s)\pi^h(a|s)$), Top-1 accuracy, and KL divergence between marginal action distributions.
At the rollout level, we report per-participant absolute deviations of agent rates from expert rates, for the five action rates and for the episode outcome.

\begin{table}[h]\centering\footnotesize\setlength{\tabcolsep}{4pt}
\resizebox{\linewidth}{!}{%
\begin{tabular}{@{}lccccc@{}}
\toprule
& \multicolumn{3}{c}{\textit{State-level}} & \multicolumn{2}{c}{\textit{Rollout}} \\
\cmidrule(lr){2-4}\cmidrule(lr){5-6}
Model & NLL$\downarrow$ & KL$\downarrow$ & Top-1$\uparrow$ & mean $|\Delta\mathrm{action}|\downarrow$ & outcome gap$\downarrow$ \\
\midrule
\multicolumn{6}{l}{\emph{Pedestrian crossing ($N{=}52$; mean [min, max] over training seeds; outcome = success rate)}} \\
BC$^\dagger$ & 0.948 [0.940, 0.955] & 0.037 [0.036, 0.037] & 0.689 [0.688, 0.691] & 0.024 [0.024, 0.025] & 0.190 [0.185, 0.198] \\
TTC$^\dagger$ & 1.170 & 0.016 & 0.485 & 0.028 & 0.514 \\
Standard AIRL & 15.67 [14.94, 16.81] & 0.2286 [0.2234, 0.2313] & 0.549 [0.544, 0.553] & 0.0658 [0.0646, 0.0664] & 0.280 [0.272, 0.285] \\
GAIL & 14.22 [12.63, 15.32] & 0.266 [0.261, 0.271] & 0.460 [0.455, 0.462] & 0.093 [0.089, 0.097] & 0.506 [0.503, 0.508] \\
Option-GAIL & 15.54 [14.31, 17.44] & 0.622 [0.591, 0.639] & 0.363 [0.356, 0.373] & 0.119 [0.118, 0.120] & 0.507 [0.505, 0.509] \\
Single-$h$, $h{=}5$ & 2.109 [2.059, 2.163] & 0.116 [0.112, 0.120] & 0.610 [0.607, 0.611] & 0.037 [0.034, 0.038] & 0.191 [0.184, 0.195] \\
Episode-AIRL & 1.656 [1.626, 1.679] & 0.116 [0.113, 0.120] & 0.617 [0.614, 0.620] & 0.045 [0.043, 0.046] & 0.188 [0.182, 0.194] \\
\textbf{SC-AIRL (ours)} & \textbf{1.579 [1.549, 1.594]} & \textbf{0.083 [0.081, 0.085]} & \textbf{0.647 [0.643, 0.651]} & \textbf{0.033 [0.033, 0.033]} & \textbf{0.178 [0.172, 0.190]} \\
\midrule
\multicolumn{6}{l}{\emph{Highway overtaking ($N{=}47$; mean [min, max] over training runs; outcome = cleared rate)}} \\
BC$^\dagger$ & 0.617 [0.613, 0.619] & 0.0065 [0.0063, 0.0070] & 0.794 [0.794, 0.794] & 0.023 [0.023, 0.024] & 0.141 [0.141, 0.142] \\
TTC$^\dagger$ & 0.674 & 0.006 & 0.794 & 0.021 & 0.146 \\
Standard AIRL & 4.422 [4.385, 4.459] & 0.102 [0.097, 0.107] & 0.629 [0.624, 0.633] & 0.031 [0.031, 0.032] & 0.123 [0.122, 0.124] \\
GAIL & 5.761 [5.675, 5.881] & 0.086 [0.082, 0.089] & 0.611 [0.608, 0.614] & 0.074 [0.071, 0.078] & 0.146 [0.146, 0.147] \\
Option-GAIL & 15.35 [14.97, 15.79] & 1.856 [1.794, 1.967] & 0.172 [0.166, 0.179] & 0.264 [0.261, 0.266] & 0.147 [0.145, 0.149] \\
Single-$h$, $h{=}12$ & 2.681 [2.663, 2.692] & 0.107 [0.099, 0.111] & 0.654 [0.651, 0.659] & 0.034 [0.033, 0.036] & 0.131 [0.129, 0.133] \\
Episode-AIRL & 1.742 [1.676, 1.788] & 0.070 [0.069, 0.072] & 0.675 [0.674, 0.676] & 0.028 [0.026, 0.029] & 0.128 [0.122, 0.131] \\
\textbf{SC-AIRL (ours)} & \textbf{1.505 [1.442, 1.576]} & \textbf{0.046 [0.042, 0.050]} & \textbf{0.718 [0.715, 0.720]} & \textbf{0.024 [0.023, 0.026]} & \textbf{0.119 [0.108, 0.127]} \\
\bottomrule
\end{tabular}}
\caption{Held-out imitation fidelity on both tasks. Bold marks the best reward-learning model. Mean $|\Delta\mathrm{action}|$ averages the absolute deviations of the five action rates; outcome gap is $|\Delta\mathrm{SR}|$ (pedestrian) or $|\Delta\mathrm{cleared}|$ (highway). Training runs per model are listed in \S\ref{sec:setup-baselines}. Details are in Appendix~\ref{sec:appendix-holdout}. $^\dagger$Reference rows without reward learning.}
  \label{tab:imitation}
\end{table}

\paragraph{State-level results.}
Among reward-learning models, SC-AIRL attains the best held-out state-level fidelity on both tasks.
On the pedestrian task, SC-AIRL reaches NLL $1.58$ and Top-1 $0.647$ (mean over seeds), against $1.66$ and $0.617$ for Episode-AIRL and $15.7$ and $0.549$ for Standard AIRL.
On the highway task, it reaches NLL $1.505$ and Top-1 $0.718$, against $1.742$ and $0.675$ for Episode-AIRL and $4.422$ and $0.629$ for Standard AIRL.
SC-AIRL has lower NLL than both baselines for every participant in every training run of both tasks, with $45$ exceptions among the $156$ pedestrian participant runs against Episode-AIRL (paired Wilcoxon, $p{<}.002$).
BC achieves the lowest NLL and the highest or tied-highest Top-1 accuracy on both tasks, as it is trained directly on the expert action likelihood.

\paragraph{Rollout-level results.}
Among reward-learning models, SC-AIRL also imitates participants most closely in rollouts on both tasks.
On the pedestrian task, SC-AIRL closes the NOTHING-rate gap to the expert data from $-15.5$ (Standard AIRL) to $-4.9$ percentage points, a $68\%$ reduction.
On the highway task, SC-AIRL has the smallest mean action-rate deviation among reward-learning models ($0.024$) and the smallest clearing-rate gap ($0.119$); against Standard AIRL at seed $7$ these hold for $32$ and $28$ of $47$ participants (paired Wilcoxon, $p{=}.008$ and $p{=}.055$).
On both tasks, BC and TTC track the expert action rates at least as closely as SC-AIRL (pedestrian $0.024$ and $0.028$ vs.\ $0.033$; highway $0.023$ and $0.021$ vs.\ $0.024$), while TTC leaves a larger success-rate gap on the pedestrian task ($0.514$ vs.\ $0.178$).

\subsection{Shared reward consistency}
\label{sec:exp-reward}

To assess shared-reward consistency, we compare the rewards learned by SC-AIRL and Standard AIRL from the training episodes of both tasks.
We measure the per-participant Pearson correlation $r_{\mathrm{cross}}$ between the two methods' raw rewards $r_\theta(s,a)$ over all actions at the training expert states.

\begin{figure}[htbp]
  \centering
  \includegraphics[width=0.75\linewidth]{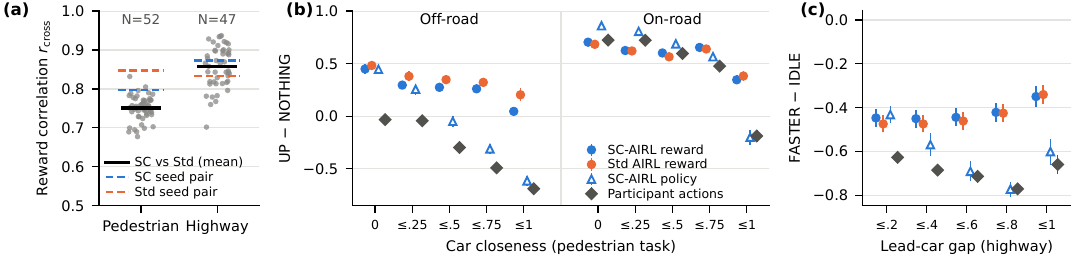}
  \caption{\textbf{Reward correlation between SC-AIRL and Standard AIRL, and reward and policy preferences.}
  (a) Per-participant correlation $r_{\mathrm{cross}}$ between SC-AIRL and Standard AIRL rewards (pedestrian: mean over seeds $42$, $7$, and $123$, $N{=}52$; highway: mean over seeds $7$ and $123$, $N{=}47$). Dashed lines show the between-run correlation of each model (mean over run pairs).
  (b) UP$-$NOTHING preference of the rewards (softmax over actions), the SC-AIRL policy, and participants' actions in road-zone $\times$ car-closeness cells of the pedestrian task (mean $\pm$ 95\% CI over participants, seed 42).
  (c) FASTER$-$IDLE preference of the same quantities in five lead-car-gap cells of the highway task (mean $\pm$ 95\% CI over participants, averaged over seeds $7$ and $123$).}
  \label{fig:reward-policy}
\end{figure}

\paragraph{Reward consistency across methods.}
SC-AIRL learns a reward that correlates with that of Standard AIRL: $r_{\mathrm{cross}}$ is $0.748$--$0.754$ on the pedestrian task ($N{=}52$) and $0.857$ on the highway task ($N{=}47$), with per-participant minima of $0.65$ and $0.54$ (mean over participants per run; sign test, $p{<}10^{-13}$; Figure~\ref{fig:reward-policy}a).
Between-run correlations within each method are comparable to this between-method value: $0.796$--$0.799$ (pedestrian) and $0.868$--$0.876$ (highway) for SC-AIRL, and $0.843$--$0.850$ (pedestrian) and $0.833$ (highway) for Standard AIRL.
Details are in Appendix~\ref{sec:appendix-reward}.

\paragraph{Reward and policy.}
In both tasks, the learned reward is distinct from the policy it trains.
In cells of a road-zone $\times$ car-closeness grid, $r_\theta$ prefers UP over NOTHING in all $10$ cells for both methods and all three training seeds, whereas the SC-AIRL policy chooses NOTHING more often in $4$ cells and participants do so in $6$ (Figure~\ref{fig:reward-policy}b).
The highway task shows the same separation across five lead-car-gap cells.
Both rewards penalize FASTER relative to IDLE less than participants do in every cell ($43$ to $47$ of $47$ participants per cell), whereas the policy penalizes it more than both rewards once the gap exceeds $0.4$ ($47$ of $47$; Figure~\ref{fig:reward-policy}c).
The reward still reflects participants' action preferences within a state.
The reward difference between UP and NOTHING correlates with participants' log-odds of choosing UP over NOTHING, with median Spearman correlations of $0.55$--$0.61$ across methods and the three training seeds; the correlation is positive for every participant.

\subsection{Correlation of the inferred depth with psychometric measures}
\label{sec:exp-clinical}

To test whether the inferred depth is related to independent psychometric measures, we use anxiety (STAI total score), whose trait component has been linked to weaker adjustment of learning to the current environment~\citep{browning2015anxious}, and impulsivity, measured by the motor subscale of the BIS-11 and associated with aberrant and risky driving in most reviewed studies~\citep{bicaksiz2016impulsivity}.
We test two correlations, anxiety with the variability of the inferred depth and motor impulsivity with how the depth shifts near a vehicle.

Each of the two hypotheses is tested at $\alpha{=}.05$ with Spearman correlations of the three-seed mean measures, raw and with the mean task score controlled (bootstrap 95\% CIs in brackets).

\begin{figure}[htbp]
  \centering
  \includegraphics[width=0.75\linewidth]{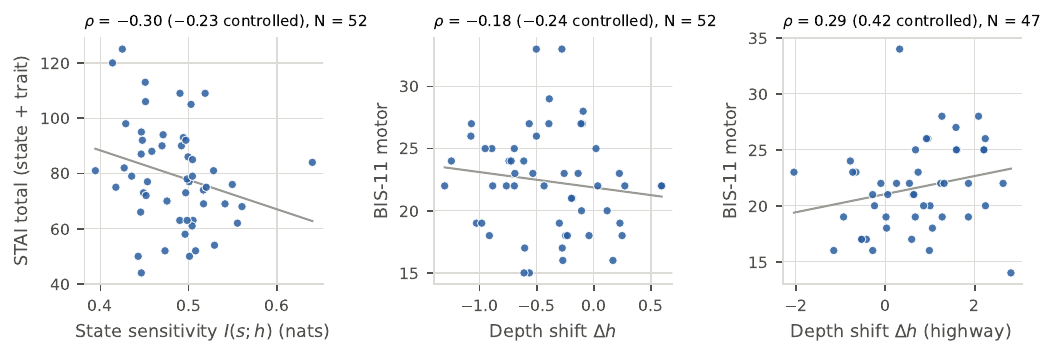}
  \caption{\textbf{Inferred depth and questionnaire scores (held-out-protocol runs, averaged over three training seeds).}
\emph{Left}: pedestrian depth variability $I(s;h)$ vs.\ STAI total score.
\emph{Middle}: pedestrian depth shift $\Delta h$ vs.\ BIS-11 motor subscale.
\emph{Right}: the same on the highway task.
Panel titles give $r_s$ without and with the mean task score controlled.}
  \label{fig:router-scatter-clinical}
\end{figure}

\paragraph{(i) --- Depth variability $I(s;h)$ $\times$ STAI anxiety.}
\emph{Depth variability} is the mutual information between state and inferred depth,
\begin{equation}
  I(s;h) \;=\; H\!\bigl(\bar{p}(h)\bigr) - \mathbb{E}_{s}\!\bigl[H\!\left(p_\phi(h|s)\right)\bigr], \qquad \bar{p}(h){=}\mathbb{E}_s[p_\phi(h|s)], \label{eq:mi-state-depth}
\end{equation}
computed over a participant's expert states.
It rises with the spread of the participant's marginal depth distribution $\bar{p}(h)$ and with how much the routing differs across states.
Depth variability correlates negatively with anxiety ($r_s{=}{-}0.295$, 95\% CI $[-0.53,-0.02]$, $p{=}.033$; $N{=}52$; Figure~\ref{fig:router-scatter-clinical}, left), and with the task score controlled the correlation is $r_s{=}{-}0.225$ ($[-0.48,+0.05]$, $p{=}.108$).
Parameter recovery at the same fit length shows that this correlation can arise from individual differences in the marginal depth distribution $\bar{p}(h)$ alone (Appendix~\ref{sec:appendix-recovery}).
On held-out states the raw correlation is $r_s{=}{-}0.28$ ($p{=}.047$).

\paragraph{(ii) --- Depth shift $\Delta h$ $\times$ BIS-11 motor impulsivity.}
With the task score controlled, the depth difference near a vehicle correlates with motor impulsivity on the highway task ($p{=}.003$) and, at $p{=}.080$, on the pedestrian task.
The \emph{depth shift} $\Delta h=\mathbb{E}[h|\mathrm{danger}]-\mathbb{E}[h|\mathrm{safe}]$ contrasts on-road states near a vehicle with off-road states (pedestrian) and states near the vehicle ahead with those without one (highway; Appendix~\ref{sec:appendix-depth-profile}).
Higher motor impulsivity accompanies a lower depth shift on the pedestrian task ($r_s{=}{-}0.175$, $[-0.39,+0.06]$, $p{=}.214$; controlled $r_s{=}{-}0.245$, $[-0.46,+0.01]$, $p{=}.080$; Figure~\ref{fig:router-scatter-clinical}, middle) and a higher depth shift on the highway task ($r_s{=}{+}0.290$, $p{=}.048$; controlled $r_s{=}{+}0.423$, $p{=}.003$; $N{=}47$; Figure~\ref{fig:router-scatter-clinical}, right).
With the per-depth policies fixed, the 500-step router recovers the participant ranking of $\Delta h$ from behavior generated with as many episodes as each participant ($\rho{=}0.90$; Appendix~\ref{sec:appendix-recovery}).

In both tasks, the correlations point in the direction that more impulsive participants shift their depth near a vehicle in the same direction as the cohort, but by a larger amount.
On the pedestrian task, the shallower policies gain weight in danger states, making fewer back or lateral moves (Appendix~\ref{sec:appendix-depth-profile}).
On the highway task, the deeper policies gain weight there, decelerating and changing lanes more (Appendix~\ref{sec:appendix-highway-depth}).

Together, (i) and (ii) indicate that the inferred depth can carry information about individual traits.

\section{Conclusion}
\label{sec:conclusion}
Standard IRL assumes infinite-horizon planning, whereas human foresight is bounded and varies with the state.
We introduced SC-AIRL, which incorporates these two properties into the planner through per-depth policies that share one reward and a state-conditional router that mixes them.
On held-out episodes of two real-time tasks, SC-AIRL narrows the gap to participants' behavior that Standard AIRL leaves.
The inferred depth correlates with independent psychometric measures.
These results indicate that incorporating the properties of human planning into IRL yields closer imitation of human behavior and an interpretable, model-derived measure of those properties.

\paragraph{Limitations and future work.}
Both tasks are simulator-based road tasks.
Training $|\Hset|$ policy--critic pairs costs $4.0$--$4.4$ times the CPU time of a Standard AIRL round (Appendix~\ref{sec:appendix-compute}).
On the highway task, the depth shift is weakly reproducible across training seeds ($0.17$--$0.28$).
The inferred depth is a model-derived proxy for foresight, testable by manipulating foresight within a participant, for example through time pressure~\citep{keramati2016habits}, and by recording ventromedial prefrontal and hippocampal activity~\citep{mattar2018prioritized,holton2025vmpfc}.

\section*{AI Use Statement}
We used generative AI tools (Anthropic's Claude Opus 5, Claude Opus 5.5, and Claude Fable 5.1) for four tasks.
First, the tools assisted in implementing the training, evaluation, and analysis code, including the code that cleans and reformats the behavioral data.
The authors reviewed the AI-assisted code and checked the reported numbers against the saved outputs of the analysis scripts.
Second, the tools provided feedback on the design of experiments and analyses, including proposing control analyses and robustness checks.
The authors decided which analyses to run and report.
Third, the tools assisted in interpreting experimental results, and the authors verified each claim against the underlying data.
Fourth, the tools edited text written by the authors for readability and checked the reference list against DOI and arXiv records.
We did not use generative AI tools to generate synthetic datasets, to develop theoretical models or conceptual frameworks, to formulate or prove mathematical claims, to propose hypotheses, to translate text, or to find related work.
Qualitative or thematic data analysis does not apply to this work.
We take responsibility for the final content of this work, including text, claims, and artifacts produced with the aid of generative AI.

\section*{Ethics Statement}
The pedestrian data are the publicly accessible behavioral data of \citet{jeong2026crosswalk}.
The highway data are the publicly accessible behavioral data of \citet{lee2024bridging}.
Participants were excluded only for task performance or CES-D scores beyond three standard deviations, which does not exclude clinical populations by design.
The inferred depth and its correlations with questionnaire scores are cohort-level associations and are not diagnostic or screening measures; limits on their use are discussed in Appendix~\ref{sec:appendix-broader-impact}.

\section*{Reproducibility Statement}
The Supplementary Material contains the training and evaluation code for both tasks, the configuration files, the result tables behind every table and figure, and the scripts that regenerate the tables and figures from them; its README maps each table and figure to the script and result file that produce it.
The highway data are available from the public repository of \citet{lee2024bridging} (\url{https://github.com/CCS-Lab/project_highway_irl_public}), and the Supplementary Material includes the script that builds our expert trajectories from it.
The pedestrian data are available from the public repository of \citet{jeong2026crosswalk} (\url{https://github.com/CCS-Lab/crosswalk-task-public}), and the trained checkpoints will be released upon publication.
Hyperparameters are listed in Appendix~\ref{sec:appendix-baseline-hp} (pedestrian) and Appendix~\ref{sec:appendix-highway-setup} (highway), the held-out split in \S\ref{sec:setup-holdout}, and training seeds in \S\ref{sec:setup-baselines}.
Of the three highway training runs, the first did not fix the random seed and the other two used seeds 7 and 123; Appendix~\ref{sec:appendix-highway} marks the seed-unfixed run.
On the pedestrian task the seed fixes network initialization and training randomness but not the order of environment episodes, so reruns with the same seed agree in distribution rather than bit for bit.

\bibliographystyle{iclr2027_conference}
\bibliography{references}

\appendix
\raggedbottom
\section{Full notation table}\label{sec:appendix-notation}
\begin{table}[h]
  \centering\small
  \begin{tabular}{@{}llp{8cm}@{}}
    \toprule
    Symbol & Type & Description \\
    \midrule
    $\mathcal{S},\mathcal{A}$ & sets & State and action spaces \\
    $P(s'|s,a)$ & kernel & Transition dynamics \\
    $r$ & function & Ground-truth (expert) reward \\
    $\gamma$ & scalar & Discount factor ($=0.99$) \\
    $\pi_E$ & policy & Expert (demonstrator) policy \\
    $\tau$ & trajectory & Trajectory $(s_0,a_0,s_1,\dots)$ (expert, or a rollout $\tau^h_k$ of $\pi^h$) \\
    \addlinespace[2pt]
    $\mathcal{H}$ & set & Discrete planning-depth set (pedestrian $\{2,3,5,10\}$; highway $\{12,16,20,24\}$) \\
    $h$ & scalar & Planning depth $h \in \mathcal{H}$ \\
    $\pi^h$ & policy & Per-depth policy trained with depth-$h$ explicit reward lookahead \\
    $V^h$ & function & Per-depth value function \\
    $A_t^{(h)}$ & scalar & Per-depth GAE advantage (Eq.~\ref{eq:gae-h}) \\
    $G_t^{(h)}$ & scalar & $h$-step truncated return \\
    $\delta_t^{(h)}$ & scalar & Per-step TD residual under depth $h$ \\
    $\omega_h,\xi_h$ & params & Policy and value parameters of depth $h$ \\
    \addlinespace[2pt]
    $r_\theta$ & function & Learned (shared) reward network \\
    $f_\theta(s,a,s')$ & scalar & Discriminator advantage $r_\theta(s,a)+\gamma\Phi(s')-\Phi(s)$; equals $r_\theta(s,a)$ in SC-AIRL ($\Phi\!\equiv\!0$) \\
    $\Phi$ & function & Potential-shaping term (fixed at $0$ in SC-AIRL; learned in Standard AIRL) \\
    $\theta$ & params & Reward network parameters \\
    \addlinespace[2pt]
    $p_\phi(h|s)$ & distribution & State-conditional planning-depth router \\
    $\phi$ & params & Router network parameters \\
    $\pi_{\text{mix}}(a|s)$ & policy & Mixed policy $\sum_h p_\phi(h|s)\pi^h(a|s)$ \\
    $D_{\text{mix}}(s,a,s')$ & scalar & Mixed-policy discriminator \\
    $D_h(s,a,s')$ & scalar & Per-depth discriminator score (Eq.~\ref{eq:qD}) \\
    $q_D(h|s,a)$ & distribution & Depth responsibility (Eq.~\ref{eq:qD}) \\
    $\lambda_{\text{ent}}$ & scalar & Weight of the entropy bonus on $q_D$ \\
    \addlinespace[2pt]
    $\lambda$ & scalar & GAE trace-decay parameter ($=0.95$) \\
    $\mathcal{L}_{\text{router}}$ & scalar & Router NLL objective (Eq.~\ref{eq:router-nll}) \\
    $K,R$ & integers & Numbers of AIRL rounds and post-hoc router rounds (Algorithm~\ref{alg:scairl}) \\
    \addlinespace[2pt]
    $I(s;h)$ & scalar & Depth variability (Eq.~\ref{eq:mi-state-depth}) \\
    $\Delta h$ & scalar & Depth shift $\mathbb{E}[h|\mathrm{danger}]-\mathbb{E}[h|\mathrm{safe}]$ (\S\ref{sec:exp-clinical}) \\
    \bottomrule
  \end{tabular}
  \caption{Notation used throughout the paper.}
  \label{tab:notation}
\end{table}

\section{Theoretical Analysis}
\label{sec:appendix-theory}
\label{sec:theory}

Assumption (A2) raises a formal question. Can changing the geometric discount rate reproduce the explicit reward weighting of the per-depth planners? Proposition~\ref{prop:bounded-support} shows that it cannot. The proof is in Appendix~\ref{sec:appendix-proof}. Remark~1 describes the reward learned under $\Phi\!\equiv\!0$ and how consistently two methods that share this parameterization learn it.

\begin{proposition}[Bounded-support weighting is not discount-rescaling]
\label{prop:bounded-support}
Let a reward-weighting kernel $w_k\ge 0$ satisfy $w_0>0$ and $w_k=0$ for all $k\ge h$. No scale $c>0$ and geometric discount $\gamma'\!\in\!(0,1)$ give $w_k=c\,\gamma'^{\,k}$ for all $k$. This covers the $h$-step return ($w_k=\gamma^k\mathbf{1}[k<h]$) and the explicit reward weights of GAE-$h$ ($w_k=(\gamma\lambda)^k\mathbf{1}[k<h]$, Eq.~\ref{eq:gae-h}).
\end{proposition}

Depth and discount play different roles in the SC-AIRL planner. The depth $h$ sets how many steps of reward the planner sums explicitly before it relies on its critic. The discount $\gamma$ weights reward geometrically at every step, both within the $h$-step window and inside $V^h$. If $V^h$ equals the on-policy value $V^{\pi^h}$, then $\mathbb{E}[\delta_{t+k}^{(h)}\mid s_t,a_t]=0$ for $k\ge1$, so $\mathbb{E}[A_t^{(h)}\mid s_t,a_t]=A^{\pi^h}(s_t,a_t)$ for every $h\ge1$. Because $V^h$ is fitted from finite rollouts, $h$ sets how much of the advantage rests on observed reward rather than on the critic's estimate.

\paragraph{Remark 1 --- Reward consistency across methods.}
Each per-depth planner is trained toward the MaxEnt condition $\pi^h(a|s)\!\propto\!\exp Q^h(s,a)$ \citep{ziebart2008maxent,fu2018airl} under its own depth $h$ (Assumption~A2), so depth heterogeneity stays in the planners $\{\pi^h,V^h\}$ (§\ref{sec:method-reward}).
We interpret $\rth$ as a per-participant, model-derived value descriptor within one environment.
We do not treat it as an identifiable ground-truth reward or transfer it across dynamics.
SC-AIRL and AIRL (non-shaping; Appendix~\ref{sec:appendix-baseline-hp}) share the parameterization $\Phi\!\equiv\!0$ and differ in the planner and in the entropy bonus on $q_D$ (Eq.~\ref{eq:qD}).
Their agreement therefore measures the cross-method consistency of the learned reward rather than the effect of truncation alone.
On the held-out-protocol runs, the per-participant correlation between them is $0.778\pm0.047$ on training expert states (seed 42; Appendix~\ref{sec:appendix-reward}).

\subsection{Proof of Proposition~\ref{prop:bounded-support}}\label{sec:appendix-proof}
\begin{proof}
For any $c>0$ and $\gamma'\!\in\!(0,1)$, $c\,\gamma'^{\,k}>0$ for every $k\ge 0$, whereas $w_k=0$ for every $k\ge h$.
The two kernels therefore differ at every $k\ge h$, and no geometric-discount rescaling reproduces a bounded-support kernel.
\end{proof}

\section{State-conditional depth router}
\label{sec:appendix-router-variants}

\subsection{Output distribution families}
\label{sec:appendix-router-families}

Let $k\in\{0,\dots,|\Hset|-1\}$ index the depths in $\Hset$.

\paragraph{Categorical.}
\begin{equation}
  p_\phi(h_k|s) \;=\; \mathrm{softmax}(g_\phi(s))_k,
  \label{eq:router-categorical}
\end{equation}
with unconstrained logits $g_\phi(s)\in\mathbb{R}^{|\Hset|}$, which place no ordering or shape constraint on $p_\phi(h|s)$.

\paragraph{Stick-breaking.}
\begin{equation}
  p_\phi(h_k|s) \;=\; v_k(s)\prod_{j<k}(1-v_j(s)),
  \qquad v_k(s) = \sigma(g_\phi(s)_k),
  \label{eq:router-stickbreaking}
\end{equation}
for $k<|\Hset|-1$, with $p_\phi(h_{|\Hset|-1}|s)=\prod_{j<|\Hset|-1}(1-v_j(s))$, where $\sigma$ denotes the sigmoid, with $g_\phi(s)\in\mathbb{R}^{|\Hset|-1}$. Cumulative ordering is imposed, admitting both unimodal and multimodal profiles over $\Hset$.

\paragraph{Gaussian kernel.}
\begin{equation}
  p_\phi(h_k|s) \;\propto\; \exp\!\Bigl(-\tfrac{(k - \mu_\phi(s))^2}{2\sigma^2}\Bigr),
  \label{eq:router-gaussian}
\end{equation}
where $\mu_\phi(s)\in[0,|\Hset|-1]$ is a bounded scalar output and $\sigma$ is a fixed width. This family is unimodal over the ordered depth index.
Besides this fixed-width kernel on the depth index, we use two other Gaussian kernels: a kernel on the depth values $h$ with a learned state-dependent width $\sigma(s)\in[0.3,10]$ (H-space), and a kernel on $\log h$ with a state-dependent width. Appendix~\ref{sec:appendix-router-robustness} states which kernel each analysis uses.

On the pedestrian task the three families share one backbone (two hidden layers of width 64, LeakyReLU, dropout 0.1) and differ only in the output head. The highway routers have two hidden layers of width 32 and no dropout (Table~\ref{tab:highway-hparams}).

\subsection{Training schedule \texorpdfstring{$\times$}{x} distribution family}

SC-AIRL varies two design choices (the loop schedule was run for Categorical and Stick-breaking only, Appendix~\ref{sec:appendix-loop-validation}):

\begin{enumerate}\itemsep=0pt
  \item \textbf{Output distribution.} The router $p_\phi(h|s)$ is parameterized as one of the three families defined above (§\ref{sec:appendix-router-families}).
  \item \textbf{Router integration in training (loop vs.\ no-loop).} In the \emph{loop} condition, $p_\phi$ is updated end-to-end within the AIRL training loop and enters the discriminator both as the depth prior and through the depth responsibility $q_D$ (Eq.~\ref{eq:qD}). In the \emph{no-loop} condition, all per-depth policies are trained with a uniform prior $p_\phi(h|s)=1/|\Hset|$ throughout, and the router is fit post-hoc on the converged policies.
\end{enumerate}

Table~\ref{tab:imitation-insample} reports the no-loop imitation metrics of every family on the full-data runs, and Appendix~\ref{sec:appendix-loop-validation} compares the two schedules.

\subsection{Loop variants validation}\label{sec:appendix-loop-validation}

We trained end-to-end loop variants for Categorical and Stick-breaking: the two families with completed full-length AIRL runs under the loop schedule. The loop variant of Algorithm~\ref{alg:scairl} interleaves router updates with AIRL training. After a warmup of $T_\mathrm{warm}=10{,}000$ rounds (uniform prior), $\phi$ is updated every $\Delta_\mathrm{router}=10$ rounds via Eq.~\ref{eq:router-nll} just before the discriminator step, and the post-hoc fit (lines~18--22 of Algorithm~\ref{alg:scairl}) is omitted. Both loop variants share the SC-AIRL training configuration of Appendix~\ref{sec:appendix-baseline-hp} (optimizer, capacity, batch size, $K$). We performed no per-family re-tuning of the loop schedule. Table~\ref{tab:loop-validation} reports per-variant imitation metrics with rollouts under each variant's router (in-loop or post-hoc), as in Table~\ref{tab:imitation-insample}.

\begin{table}[h]
  \centering
  \footnotesize
  \setlength{\tabcolsep}{4pt}
  \resizebox{\linewidth}{!}{%
  \begin{tabular}{@{}llcccccccc@{}}
    \toprule
    & & \multicolumn{3}{c}{\textit{State-level (expert states)}}
      & \multicolumn{5}{c}{\textit{Rollout-level}} \\
    \cmidrule(lr){3-5}\cmidrule(lr){6-10}
    Output dist. & Train mode
      & NLL$\downarrow$ & KL$\downarrow$ & Top-1$\uparrow$
      & $|\Delta\mathrm{SR}|\downarrow$ & $|\Delta\mathrm{CR}|\downarrow$
      & $|\Delta\mathrm{TO}|\downarrow$ & $|\Delta\mathrm{UP}|\downarrow$
      & $|\Delta\mathrm{NOTHING}|\downarrow$ \\
    \midrule
    Categorical    & Loop (end-to-end)  & $1.499$ & $0.023$ & $0.726$ & $0.086$ & $0.084$ & $0.013$ & $0.028$ & $0.062$ \\
    Categorical    & No-loop (post-hoc) & $1.410$ & $0.018$ & $0.768$ & $0.091$ & $0.084$ & $0.013$ & $0.028$ & $0.043$ \\
    \addlinespace[2pt]
    Stick-breaking & Loop (end-to-end)  & $1.437$ & $0.024$ & $0.719$ & $0.082$ & $0.076$ & $0.013$ & $0.028$ & $0.058$ \\
    Stick-breaking & No-loop (post-hoc) & $1.369$ & $0.019$ & $0.767$ & $0.100$ & $0.092$ & $0.012$ & $0.028$ & $0.047$ \\
    \bottomrule
  \end{tabular}}
  \caption{Loop vs.\ no-loop imitation metrics for Categorical and Stick-breaking ($N{=}52$, full-data runs, paired by participant). No-loop attains higher state-level imitation in both families (loop $-$ no-loop for Categorical and Stick-breaking: $\Delta\mathrm{Top}\text{-}1$ $-0.042$ and $-0.048$, $\Delta\mathrm{NLL}$ $+0.09$ and $+0.07$, $\Delta\mathrm{KL}\!=\!+0.005$ each) and a smaller NOTHING-rate gap. Rollout $|\Delta\mathrm{SR}|$ and $|\Delta\mathrm{CR}|$ differences are small ($\leq 0.018$) and within paired participant noise ($\sigma_\mathrm{PP}\!\approx\!0.05$).}
  \label{tab:loop-validation}
\end{table}

\paragraph{Relation to factored latent-variable imitation.}
The state-level deficit of the loop variants (Table~\ref{tab:loop-validation}) aligns with prior latent-variable imitation findings. \citet{seo2024idil} report that the joint variant IDIL-J underperforms the factored IDIL in intent-driven domains and qualitatively fails to recover interpretable intent-driven behavior. \citet{seo2025dtil} carry the factored design into hierarchical multi-agent settings. Our setting differs (state-conditional depth router; AIRL retained at the reward layer), but the loop variant, like IDIL-J, trains the depth router jointly with the imitation objective. In the loop regime, $\{\pi^h\}$ and $p_\phi$ co-evolve through the discriminator prior, so drift in $p_\phi$ can shift the discriminator signal against which each $\pi^h$ is trained. The no-loop regime decouples these dynamics by training $\{\pi^h\}$ under a fixed uniform prior and fitting $p_\phi$ post-hoc on the converged $\{\pi^h\}$.

\section{Training Algorithm}
\label{sec:appendix-algo}

\paragraph{Per-round structure.}
Each AIRL round of §\ref{sec:method-algo} executes four steps.
(1) It collects an on-policy rollout $\tau^h_k$ from every per-depth policy $\pi^h_{\omega_h}$.
(2) It updates $\theta$ on the mixed-policy AIRL objective (Eq.~\ref{eq:dmix-obj}), with the generator density $\pimix$ assembled from the current $\{\pi^h\}$ and the uniform prior.
(3) It recomputes the learned rewards $\rth(s_{\cdot},a_{\cdot})$ on each $\tau^h_k$, normalizes them, and adds the environment's terminal reward at episode ends to form the PPO reward.
(4) It updates each $(\omega_h,\xi_h)$ with standard PPO~\citep{schulman2017ppo} on the $h$-step advantage $A_t^{(h)}$ (Eq.~\ref{eq:gae-h}). PPO truncates GAE at the end of each fixed-length trajectory segment, whereas GAE-$h$ truncates it $h$ steps after each time step.
On the pedestrian task the terminal reward is $+100$ scaled by the remaining fraction of the $30$\,s limit on success, $-100$ scaled by the collision penalty divided by $1{,}000$ on collision, and $-10$ on timeout. On the highway task it is $-50$ on collision. State-normalization statistics are frozen during steps (1)--(3) so that collection, the discriminator update, and reward recomputation operate on identical normalization; they are updated once per round, before step (4).

\paragraph{Entropy bonus.}\label{sec:appendix-entropy}
The entropy bonus on $q_D$ (Eq.~\ref{eq:qD}) is computed on expert transitions with the per-depth policies held fixed and the depth prior uniform, as during AIRL training, so its gradient reaches only the reward network.
Its weight is $\lambda_{\text{ent}}=10^{-2}$.

\begin{algorithm}[h]
  \caption{SC-AIRL training (no-loop default; loop variant in Appendix~\ref{sec:appendix-loop-validation}; post-hoc fit for the held-out evaluation and the trait measures: $R{=}100$ rounds of $5$ gradient steps ($500$ steps) on the pedestrian task and $600$ steps on the highway task, chosen by the rule of \S\ref{sec:method-algo}, with $\{\rth,\pi^h\}$ frozen)}
  \label{alg:scairl}
  \begin{algorithmic}[1]
    \State \textbf{Input:} expert transitions $\mathcal{D}_E$; depth set $\Hset$; total AIRL rounds $K$; post-hoc router rounds $R$
    \State \textbf{Init:} reward $\theta$; per-depth policy--value $\{(\omega_h,\xi_h)\}_{h\in\Hset}$
    \State \textbf{Set} $p_\phi(h|s) \gets 1/|\Hset|$ \Comment{uniform prior throughout AIRL}
    \For{$k = 0, 1, \dots, K-1$} \Comment{AIRL training}
      \For{$h \in \Hset$}
        \State collect on-policy rollout $\tau^h_k$ from $\pi^h_{\omega_h}$ \Comment{step 1}
      \EndFor
      \State sample expert batch from $\mathcal{D}_E$ and generator batches from $\{\tau^h_k\}_{h\in\Hset}$
      \State update $\theta$ by gradient step on Eq.~\ref{eq:dmix-obj} \Comment{step 2: discriminator}
      \For{$h \in \Hset$}
        \State compute $h$-step advantages $A^{(h)}$ on $\tau^h_k$ via $\rth$ and Eq.~\ref{eq:gae-h} \Comment{step 3}
      \EndFor
      \State update state-normalization statistics on $\bigcup_h \tau^h_k$
      \For{$h \in \Hset$}
        \State update $(\omega_h,\xi_h)$ with PPO on $(\tau^h_k, A^{(h)})$ \Comment{step 4}
      \EndFor
    \EndFor
    \State \textbf{Init} router $\phi$
    \For{$r = 0, 1, \dots, R-1$} \Comment{post-hoc router fit on converged $\{\pi^h\}$}
      \State sample expert batch from $\mathcal{D}_E$
      \State update $\phi$ by $5$ gradient steps on Eq.~\ref{eq:router-nll} \Comment{NLL minimization}
    \EndFor
    \State \Return $\rth,\ \{\pi^h\}_{h\in\Hset},\ p_\phi$
  \end{algorithmic}
\end{algorithm}

\paragraph{Router trained alongside AIRL.}
In the full-data runs, a router was also trained alongside AIRL.
It never enters the discriminator, and the post-hoc router (lines~18--22 of Algorithm~\ref{alg:scairl}) replaces it in every reported number.
Its training started after $100$ rounds at seed $42$ ($10{,}000$ rounds for two participants) and after $10{,}000$ rounds at seeds $7$ and $123$.

\section{Dataset Details}
\label{sec:appendix-data}

\paragraph{Participants and demonstrations.}
Six participants were excluded before any modeling, for task performance more than three standard deviations below the cohort mean ($z{<}{-3}$) or a CES-D score more than three standard deviations above it ($z{>}3$).
The remaining $N{=}52$ participants contribute $5{,}803$ episodes in total, $85$--$146$ each (median $112.5$).
Episodes from the practice session that preceded the task are excluded throughout.
For training, each participant's episodes are augmented with left--right mirrored copies, and the NOTHING steps before the first movement of each episode are removed; all evaluations use the unaugmented episodes.

\paragraph{What the task keeps and what it simplifies.}
The task keeps real-time control under a $30$\,s deadline, polled every $100$\,ms, with a standing collision risk, and a state that varies within and between episodes (vehicle speed and size, crosswalk distance, remaining time).
It simplifies the action set to five discrete keyboard actions, uses a designed score in which the cost of delay is linear, announces collision severity by vehicle color, and moves vehicles by fixed rules.
Each episode has one agent and no other pedestrians.
The task therefore sits between trial-based risk paradigms, which control the situation but do not elicit continuous decisions under time pressure, and observation in real traffic, where the experimenter sets neither the initial conditions nor the number of observations per participant~\citep{wise2024naturalistic,mobbs2021ethology}.

\paragraph{Task reward and depth grid.}
The score has two components: a per-road progress bonus of $+5$ per grid unit of forward advancement, and an episode-completion bonus of $+50 \times t_{\mathrm{remaining}}$.
Collisions incur graded penalties of $-100$, $-500$, or $-1000$ by vehicle color, encoding collision severity.
The depth grid $\Hset=\{2,3,5,10\}$ follows the task's spatial structure.
Because UP actions make up $43{\pm}6\%$ of steps (range $34$--$59\%$), $h{=}2,3$ steps advance about one lane width, and $h{=}5$ and $h{=}10$ a narrow and a wide lane block.
A step lasts $100$\,ms for NOTHING and $400$\,ms for each movement action, so the grid does not map to a fixed interval in seconds.

\paragraph{Ethics, consent, and compensation.}
The study was approved by the institutional review board of the authors' institution.
All participants provided written informed consent after the procedure and risks of the simulator-based crossing task were explained; risks were assessed as minimal.
Participants received a base honorarium of approximately USD~3.6 per 30~minutes of laboratory time, calculated proportionally and rounded up to the next half-hour.
Upon completion of the full protocol, participants received an additional performance-based bonus of up to approximately USD~3.6 determined by task score.
Total compensation typically ranged from approximately USD~11 to USD~18 for the 1.5--2~hour session (converted from the local currency).

\subsection{Recruitment, Instructions, and Screen Captures}
\label{sec:appendix-protocol}

\paragraph{Recruitment.}
Participants were recruited online through a public study advertisement.
Eligibility: adults aged 18 or above; individuals who had previously participated in this task in another study were excluded.
The advertisement disclosed the study purpose (modeling decision-making behavior under varied real-time situations using machine learning), the expected time commitment (1.5--2~hours, single visit), the personal information collected (name, account number, contact, gender, age), the compensation structure, and the IRB number.
Sign-up was completed through an online scheduling channel.

\paragraph{Pre-task procedure.}
Upon arrival, participants received a printed study-information sheet to read at their own pace, signed three consent forms (one returned to the participant), and recorded the information needed for honorarium payment (institution, name, signature).
Mobile phones were silenced and set aside.
Verbal task instructions were delivered in the participants' native language using a standardized script accompanied by a 7-slide visual aid.

\paragraph{Pedestrian-task instructions (English translation of the standardized script).}
\begin{itemize}\itemsep=2pt
  \item \textbf{Game and goal.} The task, called ``Crossing the Street,'' asks participants to reach the top of a top-down map as quickly as possible.
  \item \textbf{Controls.} Movement uses the arrow keys or WASD. The simulator polls the held key every 100~ms and moves the avatar in that direction; participants are instructed to \emph{hold} the desired direction rather than tap.
  \item \textbf{View.} The visible region is fixed around the start position; the dark left and right side regions are unreachable.
  \item \textbf{Scoring.} Each round has a 30~s time limit. Reaching the top yields a base score plus a bonus proportional to the remaining time. If time expires before reaching the top, a partial score is awarded based on the vertical distance traveled.
  \item \textbf{Road penalties.} Dark tiles are roads on which cars travel. Collision with a car ends the round and applies a colored penalty: red (highest), yellow (medium), green (lowest), following traffic-light semantics.
  \item \textbf{Car sizes.} Small cars (15 variants) occupy a single tile. Large cars (4 variants) occupy two vertically adjacent tiles, so participants must judge hitbox extent before sharing a road column.
  \item \textbf{Crosswalks.} Each road contains at most one crosswalk, placed probabilistically. Cars approaching the crosswalk yield to a waiting player; cars already overlapping the crosswalk continue to clear so as not to block the player.
\end{itemize}

\paragraph{Comprehension quiz.}
Participants completed a 10-item multiple-choice quiz testing the rules above.
Any incorrect items were reviewed verbally with the experimenter before proceeding.

\paragraph{Practice and main task.}
A 5-minute supervised practice block preceded the main task; additional practice was offered on request and excluded from all analyses.
The main task comprised two 20-minute sessions separated by a 5-minute break.
Sessions could not be paused; participants were instructed to use the restroom and address any other needs before each session began.

\begin{figure}[htbp]
  \centering
  \includegraphics[width=0.80\linewidth]{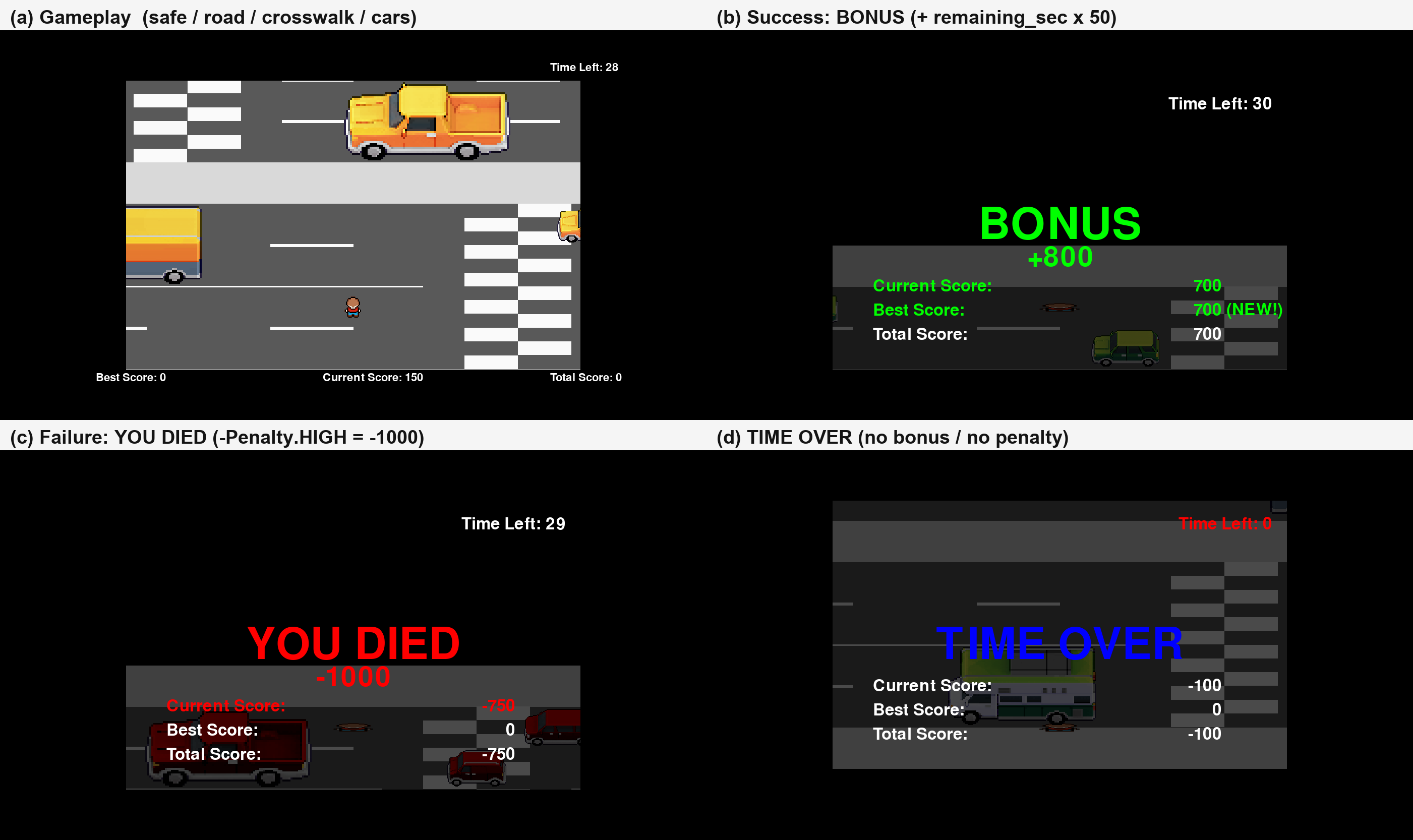}
  \caption{Representative screen captures of the pedestrian-crossing task.
  \textit{Top-left}: active gameplay; the head-up display shows current/best/total score and remaining time.
  \textit{Top-right}: ``BONUS'' end-state when the player reaches the top with time remaining.
  \textit{Bottom-left}: ``YOU DIED'' end-state on collision with a car.
  \textit{Bottom-right}: ``TIME OVER'' end-state when the 30~s round timer expires before the player reaches the top.}
  \label{fig:env-screenshots}
\end{figure}

\paragraph{Screen captures.}
Figure~\ref{fig:env-screenshots} shows representative screen captures of the four task states.

\subsection{State Space}

Table~\ref{tab:state-schema} groups the 113-dimensional observation vector by functional role. All values are normalized to $[0,1]$; Boolean channels are $\{0,1\}$-valued (indices: 5--13, 33--40).

\begin{table}[h]
  \centering\small
  \begin{tabularx}{\linewidth}{@{}llcX@{}}
    \toprule
    Group & Indices & Dims & Description \\
    \midrule
    Meta & 0--4 & 5 & Time remaining (1); closeness to unreachable area, 4 directions (4) \\
    Tile types & 5--12 & 8 & Dangerous-row flag: current/up/down (3, Boolean); crosswalk flag: current/up/down/right/left (5, Boolean) \\
    Target road & 13--23 & 11 & Crosswalk visible (1, Boolean); car penalty level (1); distance to road end/start (2); total cars visible (1); approaching cars by quadrant NW/NE/SE/SW/left/right (6) \\
    Crosswalk proximity & 24--32 & 9 & Closeness to 3 nearest crosswalks $\times$ 3 directions (up/right/left) \\
    Vehicle channels & 33--112 & 80 & Car existence in $3{\times}3$ grid (8, Boolean); head closeness $2{\times}6{\times}2$ (24); tail closeness (24); tail distance $6{\times}2$ (12); car speed (12) \\
    \bottomrule
  \end{tabularx}
  \caption{Grouped breakdown of the 113-dimensional state vector.}
  \label{tab:state-schema}
\end{table}

\subsection{Clinical Battery}
\label{sec:appendix-clinical}

Behavioral tasks and self-report surveys were administered following task completion ($\approx$25 min). The \textbf{Delay Discounting Task} (DDT; \citealt{green2004discounting}; 20 trials) required choices between a smaller immediate and a larger delayed monetary reward; individual discount rates ($\log k$) were estimated via Adaptive Design Optimization (ADO; \citealt{ahn2020ado}). The \textbf{Choice under Risk and Ambiguity} task (CRA; \citealt{levy2010cra}; 30 trials) required choices between a risky or ambiguous gamble and a safe option, and risk preference ($\alpha$) was estimated with ADO in the same way. Impulsivity was self-reported on the \textbf{Barratt Impulsiveness Scale} (BIS-11; \citealt{patton1995bis}; Korean version: \citealt{lee2012bis_korean}; 30 items, 4-point Likert; attentional, motor, and non-planning subscales) and the \textbf{Short UPPS-P Impulsive Behavior Scale} (S-UPPS-P; \citealt{cyders2014supps}; Korean version: \citealt{lim2018supps_korean}; 20 items, 4-point Likert). Depression and anxiety were self-reported on the \textbf{Center for Epidemiologic Studies Depression Scale} (CES-D; \citealt{radloff1977cesd}; Korean version: \citealt{chon2001cesd}) and the \textbf{State-Trait Anxiety Inventory} (STAI; \citealt{spielberger1970stai}; sum of the 20-item state and 20-item trait scales, range 40--160).

\section{Highway task details}\label{sec:appendix-highway}

This appendix gives the settings of the highway task (\S\ref{sec:setup-highway}), the per-run values behind the highway block of Table~\ref{tab:imitation}, paired tests per run, the routing ablation against a uniform mixture, and the depth gradients cited in \S\ref{sec:exp-clinical}.
Router-family results are in Tables~\ref{tab:router-robust-imitation} and~\ref{tab:router-robust-traits}.
SC-AIRL, Episode-AIRL, and Single-$h$ have three training runs (unfixed, $7$, $123$), Standard AIRL two ($7$, $123$), and AIRL (non-shaping) two (unfixed, $7$). The seed-$123$ cells of AIRL (non-shaping) are marked ``---''.
The unfixed run is the first cohort run, whose seed was not controlled (its configuration records seed $42$ as a placeholder).
The routing ablation, the depth gradients, and the router-family comparisons use the unfixed and seed-$7$ runs.

\subsection{Task, observation, and training settings}\label{sec:appendix-highway-setup}

Table~\ref{tab:highway-task} lists the task constants and demonstration data, Table~\ref{tab:highway-state} the $11$ observation dimensions, and Table~\ref{tab:highway-hparams} the training and evaluation settings against the pedestrian task.
The two training configurations differ in the depth grid and the number of rounds.

\paragraph{What the task keeps and what it simplifies.}
The task keeps real-time control at a $0.2$\,s step with a standing collision risk, and a state that changes as the participant's car approaches, passes, and changes lanes among other cars.
It simplifies driving to five discrete actions on a three-lane road, locks the participant's control for $0.2$\,s during a lane change, limits each episode by fuel, and scores progress by a designed rule that grows with the square of speed.
Each episode has one participant car among $35$ other cars and no other human drivers.

\begin{table}[h]\centering\footnotesize\setlength{\tabcolsep}{5pt}
\resizebox{\linewidth}{!}{%
\begin{tabular}{@{}ll@{}}
\toprule
Quantity & Value \\
\midrule
Time step & $0.2$~s (state and action logged every step) \\
Road & three lanes; $35$ other cars ($11$, $11$, and $13$ in lanes 0, 1, and 2) \\
Speeds & other cars $20$--$40$, participant $20$--$120$ distance units per second \\
Visibility & $105$ distance units \\
Lane change & $0.2$~s, during which the participant cannot control the car (agents are not locked) \\
Fuel & starts at $60$ units, drains $1$ per second \\
Score per step & $0.2\times(\mathrm{speed}/10)^2$ (at most $144$ per second) \\
Collision penalty & $200$ points \\
Clear bonus & $36\times$remaining seconds on screen; $144\times$ in the released pipeline (used for evaluation scoring) \\
Training reward & learned reward plus a terminal $-50$ on collision; no clear bonus, and episodes do not end on clear \\
Episode end & collision, fuel exhausted, or every car overtaken \\
\midrule
Participants & $N{=}47$; practice session excluded \\
Demonstrations & $4{,}089$ episodes, $44$--$168$ per participant \\
Outcome shares (all episodes) & collision $72.4\%$, cleared $16.0\%$, fuel $10.4\%$, other $1.1\%$ \\
Expert action rates (all episodes) & LEFT $0.033$, IDLE $0.784$, RIGHT $0.033$, FASTER $0.093$, SLOWER $0.057$ \\
Held-out reference (4 of 22 scenarios) & $8$ episodes per participant; cleared $0.146$, crash $0.729$, IDLE $0.794$ \\
Held-out / training states per participant & median $1{,}006$ (range $225$--$1{,}955$) / median $8{,}275$ (range $7{,}370$--$9{,}277$) \\
\bottomrule
\end{tabular}}
\caption{\textbf{Task constants and demonstration data of the highway task.}
Constants follow \citet{lee2024bridging} and the released pipeline.
The held-out reference is the participant's own episodes on the four held-out scenarios (\S\ref{sec:setup-holdout}), and every rollout-level $|\Delta|$ in this appendix is measured against it.}
\label{tab:highway-task}
\end{table}

\begin{table}[h]\centering\footnotesize\setlength{\tabcolsep}{5pt}
\begin{tabular}{@{}clll@{}}
\toprule
Index & Feature & Scale & Note \\
\midrule
0 & own speed & normalized & \\
1 & own lane & $y/8$ (continuous) & lane index $=\mathrm{clip}(\mathrm{round}(2s_1),0,2)$ \\
2--4 & lead-vehicle speed, lanes 0--2 & normalized & \\
5--7 & forward gap, lanes 0--2 & $/105$ & equals $1.0$ when no vehicle within $105$ units \\
8--10 & rear gap, lanes 0--2 & $/46$ & equals $1.0$ when none within $46$ units behind \\
\bottomrule
\end{tabular}
\caption{\textbf{The $11$ observation dimensions.}
The lead-vehicle gap of the participant's own lane is $s_{5+\mathrm{lane}}$; the danger and safe state sets of \S\ref{sec:exp-clinical} are defined on this quantity.}
\label{tab:highway-state}
\end{table}

\begin{table}[h]\centering\footnotesize\setlength{\tabcolsep}{5pt}
\resizebox{\linewidth}{!}{%
\begin{tabular}{@{}lll@{}}
\toprule
Setting & Highway & Pedestrian \\
\midrule
Depth grid $\Hset$ & $\{12,16,20,24\}$ ($2.4$--$4.8$~s) & $\{2,3,5,10\}$ \\
Training rounds & $3{,}000$ & $20{,}000$ \\
Steps per round & $512$ & $512$ \\
Reward network $r_\theta$ & hidden $8$, $\Phi\equiv0$ & same \\
Policy / value networks & $2\times32$ / $32$, one pair per depth & same \\
PPO & $10$ epochs, batch $64$, clip $0.2$, $\gamma{=}0.99$, $\lambda{=}0.95$, entropy $0.01$, lr $3\times10^{-4}$ & same \\
Discriminator & lr $10^{-3}$, weight decay $10^{-4}$, replay of $5$ rounds & same \\
State / reward normalization & EMA momentum $0.99$ & same \\
Post-hoc router (reported) & categorical MLP, hidden $32$, $600$ Adam steps, lr $10^{-3}$, wd $10^{-4}$ & $500$ Adam steps, lr $10^{-3}$, batch $512$ (imitation and traits) \\
Rollouts & $150$ episodes on the $4$ held-out scenarios, rollout seed $0$, collision reward $-200$, stop on clear & Appendix~\ref{sec:appendix-holdout} \\
\bottomrule
\end{tabular}}
\caption{\textbf{Training and evaluation settings of the highway task against the pedestrian task.}
``Same'' marks fields with identical values in the two configurations.}
\label{tab:highway-hparams}
\end{table}

\paragraph{Depth grid.}
For reference against the grid's $2.4$--$4.8$~s, \citet{montgomery2014ttc} report group means of per-driver minimum time-to-collision at braking of $2.5$--$6.4$~s.
The grid was chosen on three pilot participants, who are part of the cohort, before the cohort was trained; excluding them leaves the score-controlled highway depth-shift correlation at $+0.396$ ($p{=}.008$, $N{=}44$).

\subsection{Per-seed imitation fidelity}\label{sec:appendix-highway-perseed}

Table~\ref{tab:highway-perseed} gives every model at each training run; the mean rows of the models in Table~\ref{tab:imitation} equal its highway rows.
SC-AIRL uses the categorical post-hoc router of \S\ref{sec:exp-imitation}; Episode-AIRL is the same router evaluated at the first state of each episode and held fixed; Single-$h$ is $\pi^{h=12}$ alone.

\begin{table}[h]\centering\footnotesize\setlength{\tabcolsep}{3pt}
\resizebox{\linewidth}{!}{%
\begin{tabular}{@{}llccccccccccc@{}}
\toprule
& & \multicolumn{3}{c}{\textit{State-level}} & \multicolumn{8}{c}{\textit{Rollout (held-out scenarios vs.\ held-out reference)}} \\
\cmidrule(lr){3-5}\cmidrule(lr){6-13}
Model & Run & NLL$\downarrow$ & KL$\downarrow$ & Top-1$\uparrow$ & $|\Delta\mathrm{LEFT}|$ & $|\Delta\mathrm{IDLE}|$ & $|\Delta\mathrm{RIGHT}|$ & $|\Delta\mathrm{FASTER}|$ & $|\Delta\mathrm{SLOWER}|$ & mean $|\Delta\mathrm{action}|$ & $|\Delta\mathrm{cleared}|$ & $|\Delta\mathrm{crash}|$ \\
\midrule
BC$^\dagger$  & seed 42  & 0.619 & 0.006 & 0.794 & 0.006 & 0.046 & 0.006 & 0.042 & 0.013 & 0.023 & 0.141 & 0.257 \\
              & seed 7   & 0.619 & 0.007 & 0.794 & 0.006 & 0.051 & 0.006 & 0.045 & 0.013 & 0.024 & 0.142 & 0.258 \\
              & seed 123 & 0.613 & 0.006 & 0.794 & 0.006 & 0.048 & 0.006 & 0.041 & 0.012 & 0.023 & 0.141 & 0.255 \\
              & mean     & 0.617 & 0.007 & 0.794 & 0.006 & 0.048 & 0.006 & 0.043 & 0.013 & 0.023 & 0.141 & 0.257 \\
TTC$^\dagger$ & ---     & 0.674 & 0.006 & 0.794 & 0.009 & 0.038 & 0.008 & 0.034 & 0.016 & 0.021 & 0.146 & 0.268 \\
\addlinespace[2pt]
Standard AIRL & seed 7   & 4.385 & 0.107 & 0.624 & 0.023 & 0.072 & 0.018 & 0.023 & 0.023 & 0.032 & 0.124 & 0.160 \\
              & seed 123 & 4.459 & 0.097 & 0.633 & 0.021 & 0.067 & 0.018 & 0.023 & 0.025 & 0.031 & 0.122 & 0.167 \\
              & mean     & 4.422 & 0.102 & 0.629 & 0.022 & 0.069 & 0.018 & 0.023 & 0.024 & 0.031 & 0.123 & \textbf{0.163} \\
\addlinespace[2pt]
AIRL (non-shaping) & unfixed  & 4.375 & 0.094 & 0.642 & 0.020 & 0.070 & 0.019 & 0.022 & 0.023 & 0.031 & 0.115 & 0.182 \\
                         & seed 7   & 4.375 & 0.091 & 0.646 & 0.019 & 0.070 & 0.017 & 0.026 & 0.027 & 0.032 & 0.131 & 0.162 \\
                         & seed 123 & \multicolumn{11}{c}{---} \\
                         & mean     & 4.375 & 0.092 & 0.644 & 0.020 & 0.070 & 0.018 & 0.024 & 0.025 & 0.031 & 0.123 & 0.172 \\
\addlinespace[2pt]
GAIL & seed 42 & 5.675 & 0.089 & 0.614 & 0.026 & 0.186 & 0.025 & 0.122 & 0.031 & 0.078 & 0.147 & 0.271 \\
     & seed 7 & 5.881 & 0.086 & 0.608 & 0.032 & 0.171 & 0.030 & 0.106 & 0.028 & 0.073 & 0.146 & 0.270 \\
     & seed 123 & 5.726 & 0.082 & 0.612 & 0.024 & 0.162 & 0.024 & 0.118 & 0.027 & 0.071 & 0.146 & 0.269 \\
     & mean & 5.761 & 0.086 & 0.611 & 0.027 & 0.173 & 0.026 & 0.115 & 0.029 & 0.074 & 0.146 & 0.270 \\
\addlinespace[2pt]
Option-GAIL & seed 42 & 15.793 & 1.967 & 0.170 & 0.104 & 0.665 & 0.106 & 0.276 & 0.181 & 0.266 & 0.147 & 0.269 \\
            & seed 7 & 15.273 & 1.794 & 0.166 & 0.104 & 0.658 & 0.094 & 0.266 & 0.195 & 0.263 & 0.149 & 0.268 \\
            & seed 123 & 14.970 & 1.805 & 0.179 & 0.113 & 0.652 & 0.111 & 0.252 & 0.179 & 0.261 & 0.145 & 0.269 \\
            & mean & 15.345 & 1.856 & 0.172 & 0.107 & 0.658 & 0.104 & 0.265 & 0.185 & 0.264 & 0.147 & 0.269 \\
\addlinespace[2pt]
Single-$h$, $h{=}12$ & unfixed  & 2.663 & 0.099 & 0.659 & 0.021 & 0.071 & 0.021 & 0.029 & 0.022 & 0.033 & 0.132 & 0.183 \\
                     & seed 7   & 2.692 & 0.111 & 0.651 & 0.024 & 0.078 & 0.026 & 0.029 & 0.025 & 0.036 & 0.129 & 0.200 \\
                     & seed 123 & 2.687 & 0.111 & 0.651 & 0.022 & 0.073 & 0.022 & 0.029 & 0.019 & 0.033 & 0.133 & 0.200 \\
                     & mean     & 2.681 & 0.107 & 0.654 & 0.022 & 0.074 & 0.023 & 0.029 & 0.022 & 0.034 & 0.131 & 0.194 \\
\addlinespace[2pt]
Episode-AIRL & unfixed  & 1.762 & 0.069 & 0.674 & 0.016 & 0.058 & 0.014 & 0.025 & 0.023 & 0.027 & 0.130 & 0.178 \\
             & seed 7   & 1.676 & 0.069 & 0.674 & 0.017 & 0.064 & 0.015 & 0.024 & 0.025 & 0.029 & 0.122 & 0.181 \\
             & seed 123 & 1.788 & 0.072 & 0.676 & 0.017 & 0.057 & 0.015 & 0.021 & 0.022 & 0.026 & 0.131 & 0.183 \\
             & mean     & 1.742 & 0.070 & 0.675 & 0.017 & 0.060 & 0.015 & 0.023 & 0.023 & 0.028 & 0.128 & 0.181 \\
\addlinespace[2pt]
\textbf{SC-AIRL (ours)} & unfixed  & 1.499 & 0.042 & 0.719 & 0.017 & 0.052 & 0.014 & 0.019 & 0.017 & 0.024 & 0.123 & 0.178 \\
                     & seed 7   & 1.442 & 0.046 & 0.720 & 0.018 & 0.057 & 0.016 & 0.019 & 0.020 & 0.026 & 0.108 & 0.163 \\
                     & seed 123 & 1.576 & 0.050 & 0.715 & 0.018 & 0.051 & 0.015 & 0.017 & 0.016 & 0.023 & 0.127 & 0.168 \\
                     & mean     & \textbf{1.505} & \textbf{0.046} & \textbf{0.718} & 0.017 & 0.053 & 0.015 & 0.018 & 0.018 & \textbf{0.024} & \textbf{0.119} & 0.170 \\
\bottomrule
\end{tabular}}
\caption{\textbf{Held-out imitation fidelity on the highway task by training run ($N{=}47$).}
State-level metrics are computed on the held-out expert states of each participant.
Rollout columns are participant means of the absolute deviation between the agent's rate over $150$ rollout episodes on the four held-out scenarios and the participant's rate over their eight held-out episodes; mean $|\Delta\mathrm{action}|$ averages the five action columns.
``mean'' rows average each model's runs (three, or two for Standard AIRL and AIRL (non-shaping)) and, for the models in Table~\ref{tab:imitation}, equal its highway entries.
Bold marks the best mean among reward-learning models in the NLL, KL, Top-1, mean $|\Delta\mathrm{action}|$, $|\Delta\mathrm{cleared}|$, and $|\Delta\mathrm{crash}|$ columns. In the per-action columns Episode-AIRL has the smaller unrounded LEFT and RIGHT deviations and SC-AIRL the smaller IDLE, FASTER, and SLOWER deviations.
$^\dagger$Reference rows without reward learning: BC is trained with seeds $42$, $7$, and $123$, and TTC is deterministic.}
\label{tab:highway-perseed}
\end{table}

\subsection{Paired comparisons per seed}\label{sec:appendix-highway-paired}

Table~\ref{tab:highway-paired} tests SC-AIRL against each reward-learning baseline within participants, per run.
SC-AIRL has lower held-out NLL than every reward-learning baseline for $47$ of $47$ participants in every run.
The action-rate deviation separates SC-AIRL from AIRL (non-shaping), Single-$h$, and Standard AIRL in every run, and from Episode-AIRL at seed $123$ only.
The crash-rate gap does not separate SC-AIRL from AIRL (non-shaping) ($24$ of $47$, $p{\ge}.86$), and the clearing-rate gap separates them in the seed-$7$ run only.
Standard AIRL has the smallest crash gap among the reward-learning models ($0.163$ against $0.170$ for SC-AIRL; Table~\ref{tab:highway-perseed}).

\begin{table}[h]\centering\footnotesize\setlength{\tabcolsep}{2.5pt}
\begin{tabular}{@{}llcccc@{}}
\toprule
Run & SC-AIRL vs. & NLL & Top-1 & mean $|\Delta\mathrm{action}|$ & $|\Delta\mathrm{cleared}|$ \\
\midrule
unfixed & AIRL (non-shaping) & 47/47 ($1.4\times10^{-14}$) & 44/47 ($3.9\times10^{-11}$) & 35/47 ($7.6\times10^{-4}$) & 21/47 (.34) \\
        & Episode-AIRL            & 47/47 ($1.4\times10^{-14}$) & 42/47 ($3.6\times10^{-12}$) & 28/47 (.052) & 27/47 (.21) \\
        & Single-$h$, $h{=}12$    & 47/47 ($1.4\times10^{-14}$) & 42/47 ($9.4\times10^{-9}$)  & 39/47 ($4.4\times10^{-7}$) & 28/47 (.21) \\
\midrule
seed 7  & AIRL (non-shaping) & 47/47 ($1.4\times10^{-14}$) & 43/47 ($6.3\times10^{-9}$)  & 31/47 ($7.2\times10^{-3}$) & 31/47 (.021) \\
        & Episode-AIRL            & 47/47 ($2.4\times10^{-9}$)  & 45/47 ($5.3\times10^{-9}$)  & 30/47 (.054) & 26/47 (.10) \\
        & Single-$h$, $h{=}12$    & 47/47 ($1.4\times10^{-14}$) & 43/47 ($1.1\times10^{-11}$) & 41/47 ($2.0\times10^{-8}$) & 34/47 (.015) \\
        & Standard AIRL           & 47/47 ($1.4\times10^{-14}$) & 46/47 ($4.3\times10^{-14}$) & 32/47 (.008) & 28/47 (.055) \\
\midrule
seed 123 & Episode-AIRL           & 47/47 ($1.4\times10^{-14}$) & 42/47 ($1.2\times10^{-8}$) & 33/47 ($5.9\times10^{-3}$) & 20/47 (.65) \\
        & Single-$h$, $h{=}12$    & 47/47 ($1.4\times10^{-14}$) & 43/47 ($2.5\times10^{-11}$) & 37/47 ($5.5\times10^{-7}$) & 27/47 (.36) \\
        & Standard AIRL           & 47/47 ($1.4\times10^{-14}$) & 45/47 ($5.2\times10^{-9}$)  & 36/47 ($4.2\times10^{-5}$) & 19/47 (.39) \\
\bottomrule
\end{tabular}
\caption{\textbf{Within-participant comparisons of SC-AIRL against each reward-learning baseline, per training run ($N{=}47$).}
Each cell counts the participants for whom SC-AIRL is better (lower NLL, higher Top-1, or a smaller rollout deviation) and gives the two-sided $p$ of a Wilcoxon signed-rank test on the paired differences.
Ties ($0$ to $6$ participants in the $|\Delta\mathrm{cleared}|$ column and $1$ in two Top-1 cells) are excluded from the counts.}
\label{tab:highway-paired}
\end{table}

\subsection{Router against uniform mixture and episode router}\label{sec:appendix-highway-routing}

Table~\ref{tab:highway-routing} replaces the state-conditional router with two alternatives on the same frozen policies: equal weights and state-independent weights fitted on the training split.
The third alternative, the router evaluated once at the first state of each episode, is Episode-AIRL (Tables~\ref{tab:highway-perseed} and~\ref{tab:highway-paired}).
The state-conditional router has lower held-out NLL than each alternative for $40$ to $47$ of $47$ participants in both runs.

\begin{table}[h]\centering\footnotesize\setlength{\tabcolsep}{4pt}
\begin{tabular}{@{}llcccc@{}}
\toprule
Run & Depth weights & NLL$\downarrow$ & KL$\downarrow$ & Top-1$\uparrow$ & SC-AIRL lower NLL \\
\midrule
unfixed & state-conditional router (SC-AIRL) & \textbf{1.499} & \textbf{0.042} & \textbf{0.719} & --- \\
        & uniform                            & 1.591 & 0.070 & 0.697 & 43/47 ($1.1\times10^{-11}$) \\
        & state-independent, fitted          & 1.564 & 0.067 & 0.700 & 43/47 ($3.5\times10^{-10}$) \\
\midrule
seed 7  & state-conditional router (SC-AIRL) & \textbf{1.442} & \textbf{0.046} & \textbf{0.720} & --- \\
        & uniform                            & 1.534 & 0.077 & 0.688 & 43/47 ($1.1\times10^{-10}$) \\
        & state-independent, fitted          & 1.506 & 0.071 & 0.694 & 40/47 ($7.0\times10^{-9}$) \\
\bottomrule
\end{tabular}
\caption{\textbf{State-level fidelity of alternative depth weightings on the same frozen policies ($N{=}47$, held-out states).}
All three rows of a run share the per-depth policies and reward of SC-AIRL.
Bold marks the best value within each run.
The last column counts participants for whom SC-AIRL has lower NLL than the row, with the two-sided Wilcoxon $p$.}
\label{tab:highway-routing}
\end{table}

\subsection{Depth-to-action and depth-to-outcome gradients}\label{sec:appendix-highway-depth}

\begin{figure}[h]
  \centering
  \includegraphics[width=\linewidth]{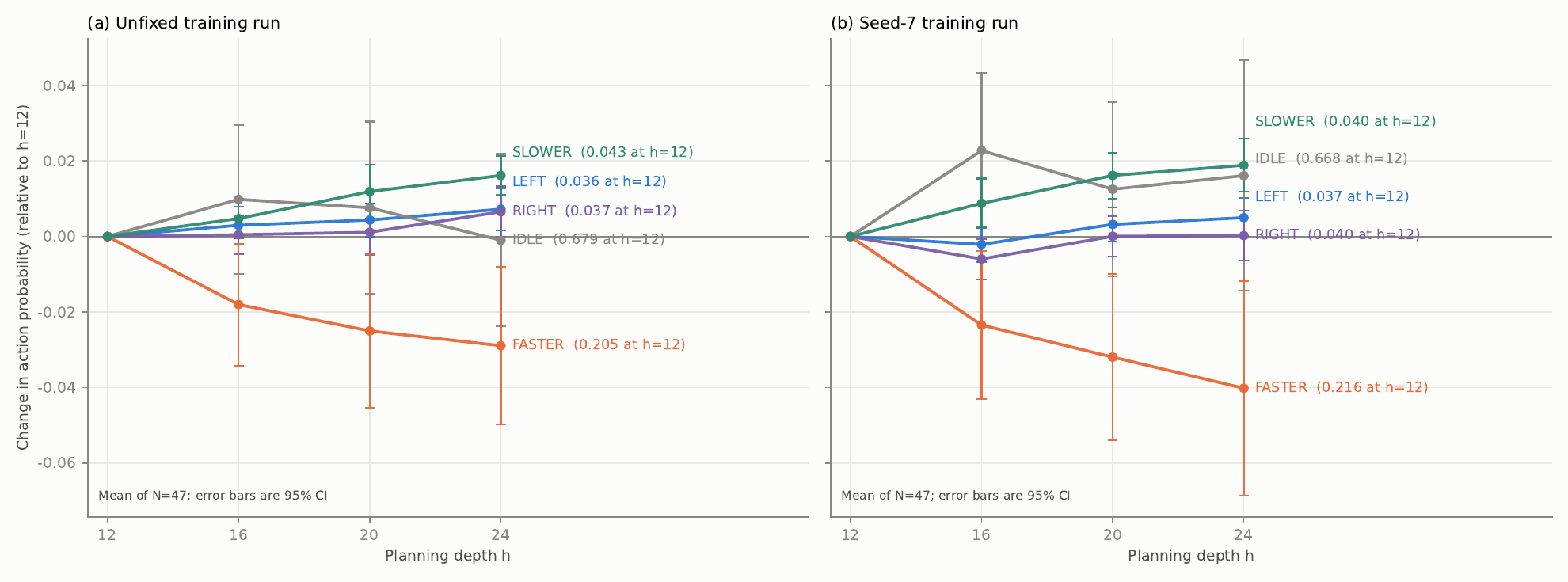}
  \caption{\textbf{Deeper per-depth policies decelerate more and accelerate less on the highway task, in both training runs.}
  Mean action probability of each frozen policy $\pi^h$ on held-out expert states, shown as the change from $h{=}12$; absolute probabilities at $h{=}12$ are given beside each label.
  (a) Unfixed training run; (b) seed-$7$ training run; the two runs share the held-out states.
  Error bars are $95\%$ confidence intervals across participants ($N{=}47$).
  Per-participant slopes are in Table~\ref{tab:highway-depth-action}.}
  \label{fig:highway-depth-mechanism}
\end{figure}

Figure~\ref{fig:highway-depth-mechanism} and Table~\ref{tab:highway-depth-action} give the action probabilities of each frozen per-depth policy on held-out expert states, and Table~\ref{tab:highway-depth-outcome} the outcomes of each policy run alone.
On expert states, the mean per-participant slope over $h$ is positive for SLOWER and negative for FASTER in both runs (Table~\ref{tab:highway-depth-action}).
The lane-change probability (LEFT$+$RIGHT) rises with depth in both runs ($34$ and $29$ of $47$; $p{=}.005$ and $.042$), driven by LEFT. RIGHT alone is significant in the unfixed run only, and the IDLE slope does not differ from zero in either run.
In rollouts, the crash rate falls with depth in both runs ($-0.0082$ and $-0.0068$ per unit depth; positive for $3$ and $6$ of $47$; $p{\le}1.1\times10^{-5}$). The cleared rate rises ($+0.0028$ and $+0.0036$; $34$ and $33$ of $47$; $p{\le}4\times10^{-4}$).

\begin{table}[h]\centering\footnotesize\setlength{\tabcolsep}{4.5pt}
\resizebox{\linewidth}{!}{%
\begin{tabular}{@{}llcccccccc@{}}
\toprule
Run & Action & Expert & Mixed & $h{=}12$ & $h{=}16$ & $h{=}20$ & $h{=}24$ & Slope per unit $h$ & Positive (Wilcoxon $p$) \\
\midrule
unfixed & LEFT   & 0.029 & 0.035 & 0.036 & 0.039 & 0.040 & 0.043 & $+0.0006$ & 32/47 (.007) \\
        & IDLE   & 0.794 & 0.724 & 0.679 & 0.689 & 0.687 & 0.678 & $-0.0001$ & 19/47 (.53) \\
        & RIGHT  & 0.030 & 0.034 & 0.037 & 0.038 & 0.038 & 0.044 & $+0.0005$ & 34/47 (.016) \\
        & FASTER & 0.091 & 0.157 & 0.205 & 0.187 & 0.180 & 0.176 & $-0.0023$ & 19/47 (.024) \\
        & SLOWER & 0.056 & 0.049 & 0.043 & 0.048 & 0.055 & 0.059 & $+0.0014$ & 39/47 ($9.1\times10^{-8}$) \\
\midrule
seed 7  & LEFT   & 0.029 & 0.036 & 0.037 & 0.035 & 0.041 & 0.042 & $+0.0005$ & 29/47 (.025) \\
        & IDLE   & 0.794 & 0.723 & 0.668 & 0.690 & 0.680 & 0.684 & $+0.0010$ & 23/47 (.68) \\
        & RIGHT  & 0.030 & 0.034 & 0.040 & 0.034 & 0.040 & 0.040 & $+0.0002$ & 30/47 (.25) \\
        & FASTER & 0.091 & 0.158 & 0.216 & 0.192 & 0.184 & 0.176 & $-0.0032$ & 22/47 (.035) \\
        & SLOWER & 0.056 & 0.049 & 0.040 & 0.048 & 0.056 & 0.059 & $+0.0016$ & 38/47 ($9.9\times10^{-7}$) \\
\bottomrule
\end{tabular}}
\caption{\textbf{Action probabilities of the per-depth policies on held-out expert states, per training run ($N{=}47$).}
Cells are participant means of the state-averaged $\pi^h(a|s)$; Expert is the empirical frequency in the same states (identical for the two runs, which share the held-out split) and Mixed is the router-weighted mixture.
Slope is the mean per-participant least-squares slope over $h\in\{12,16,20,24\}$, with the count of positive slopes and the two-sided Wilcoxon $p$.}
\label{tab:highway-depth-action}
\end{table}

\begin{table}[h]\centering\footnotesize\setlength{\tabcolsep}{5pt}
\begin{tabular}{@{}llcccccc@{}}
\toprule
Run & Policy & cleared & crash & fuel out & steps & FASTER & SLOWER \\
\midrule
unfixed & $h{=}12$ & 0.075 & 0.874 & 0.051 & 96.6 & 0.105 & 0.050 \\
        & $h{=}16$ & 0.120 & 0.814 & 0.065 & 119.4 & 0.086 & 0.059 \\
        & $h{=}20$ & 0.124 & 0.767 & 0.109 & 132.9 & 0.085 & 0.062 \\
        & $h{=}24$ & 0.111 & 0.780 & 0.108 & 131.6 & 0.086 & 0.061 \\
        & mixed (SC-AIRL) & 0.105 & 0.836 & 0.059 & 108.8 & 0.093 & 0.055 \\
\midrule
seed 7  & $h{=}12$ & 0.083 & 0.860 & 0.057 & 98.3 & 0.103 & 0.046 \\
        & $h{=}16$ & 0.123 & 0.794 & 0.083 & 119.8 & 0.086 & 0.057 \\
        & $h{=}20$ & 0.126 & 0.781 & 0.092 & 130.4 & 0.085 & 0.063 \\
        & $h{=}24$ & 0.130 & 0.773 & 0.097 & 133.7 & 0.085 & 0.062 \\
        & mixed (SC-AIRL) & 0.115 & 0.834 & 0.051 & 108.2 & 0.094 & 0.056 \\
\midrule
Expert  & all episodes & 0.160 & 0.724 & 0.104 & 126.2 & 0.093 & 0.057 \\
\bottomrule
\end{tabular}
\caption{\textbf{Outcomes of each per-depth policy run alone, per training run ($N{=}47$).}
Each per-depth row is the participant mean over $150$ rollout episodes on newly sampled traffic scenarios (rollout seed $0$); the mixed row uses the SC-AIRL router with two rollout seeds of $500$ episodes on the same scenario protocol.
These rollouts are on new scenarios rather than the held-out four, so the rows are compared with the all-episode expert reference and not with the held-out reference of Table~\ref{tab:highway-perseed}.
Steps is the mean episode length in $0.2$~s steps; FASTER and SLOWER are action rates over all rollout steps.}
\label{tab:highway-depth-outcome}
\end{table}

\subsection{Model size}\label{sec:appendix-highway-compute}
Table~\ref{tab:hw-params} lists the networks of each highway model, all of which take the $11$-dimensional state.
SC-AIRL has $4.1\times$ the parameters of Standard AIRL.
Training time is reported with the pedestrian task in Appendix~\ref{sec:appendix-compute}.

\begin{table}[h]
  \centering\small
  \begin{tabular}{@{}lcccrp{3.7cm}@{}}
    \toprule
    Model & Policy hid & Value hid & Reward/disc hid & Params & Notes \\
    \midrule
    BC                         & 32  & --- & --- & 1{,}605  & Supervised cross-entropy \\
    TTC                        & --- & --- & --- & 43       & 3 thresholds; $8\times5$ bucket action distribution \\
    GAIL                       & 32  & 32  & 8   & 3{,}295  & State--action discriminator \\
    Option-GAIL                & 32  & 32  & 8   & 13{,}789 & 4 modes; gating hidden 64; per-mode $D_c(s)$ \\
    Standard AIRL              & 32  & 32  & 8   & 3{,}472  & $h{=}\infty$; potential $\Phi$ hidden 8 \\
    AIRL (non-shaping)   & 32  & 32  & 8   & 3{,}295  & $h{=}\infty$; no shaping \\
    Single-$h$ ($h{=}12$)      & 32  & --- & --- & 1{,}605  & One SC-AIRL head at evaluation \\
    Episode-AIRL               & 32  & 32  & 8   & 14{,}101 & SC-AIRL heads and post-hoc router, evaluated once at the first state of each episode \\
    \textbf{SC-AIRL}           & 32  & 32  & 8   & 14{,}101 & 4 policy/value head pairs and reward net (12{,}529) + post-hoc router, hidden 32 (1{,}572) \\
    \bottomrule
  \end{tabular}
  \caption{Highway model sizes. Parameter counts are read from the trained checkpoints and are identical across participants within each model. Single-$h$ and Episode-AIRL reuse the SC-AIRL networks, so they add no training cost.}
  \label{tab:hw-params}
\end{table}

\section{Baseline hyperparameters and limitations}\label{sec:appendix-baseline-hp}

Table~\ref{tab:baselines} places the models on three design axes; AIRL (non-shaping) is Standard AIRL with $\Phi\!\equiv\!0$ (Appendix~\ref{sec:appendix-shaping}).

\begin{table}[htbp]
  \centering
  \small
  \begin{tabular}{@{}lcccc@{}}
    \toprule
    Model & Reward (IRL) & Horizon & State-cond.\ latent & Multi-policy \\
    \midrule
    BC~\citep{pomerleau1991bc}                          & \xmark & ---         & \xmark & \xmark \\
    GAIL~\citep{ho2016gail}                             & \xmark & $\infty$    & \xmark & \xmark \\
    Option-GAIL~\citep{jing2021optiongail}              & \xmark & $\infty$    & \checkmark\ (options) & \checkmark \\
    Standard AIRL~\citep{fu2018airl}                    & \checkmark & $\infty$ & \xmark & \xmark \\
    AIRL (non-shaping)                            & \checkmark & $\infty$ & \xmark & \xmark \\
    Single-$h$ AIRL ($h{=}5$; highway $h{=}12$)        & \checkmark & fixed $h$ & \xmark & \xmark \\
    Episode-AIRL                                              & \checkmark & mixture $h$ & fixed per episode & \checkmark \\
    \textbf{SC-AIRL (ours)}                             & \checkmark & \textbf{routed $h$} & \checkmark\ $p_\phi(h|s)$ & \checkmark \\
    \bottomrule
  \end{tabular}
  \caption{Model comparison across three design axes: reward learning, depth truncation, and latent routing granularity.}
  \label{tab:baselines}
\end{table}

Unless Table~\ref{tab:baseline-hp} states otherwise, every model uses the settings of Table~\ref{tab:highway-hparams}. BC and GAIL use batch size $128$ on the pedestrian task, BC trains for $200$ epochs, and GAIL and Option-GAIL train their discriminators without replay. All AIRL variants receive the same terminal reward (Appendix~\ref{sec:appendix-algo}).

\begin{table}[htbp]
  \centering\small
  \begin{tabular}{@{}lcccp{4.5cm}@{}}
    \toprule
    Model & Policy hid & Value hid & Reward/disc hid & Notes \\
    \midrule
    BC            & 32 & --- & --- & Supervised cross-entropy; no discriminator \\
    Standard AIRL & 32 & 32 & 8   & Untruncated ($h{=}\infty$); potential net $\Phi$ (hidden 8); single policy \\
    AIRL (non-shaping) & 32 & 32 & 8 & Untruncated ($h{=}\infty$); $\Phi\!\equiv\!0$; single policy \\
    GAIL          & 256 & 256 & 256 & State-action discriminator; no reward extraction; entropy $0.02$ \\
    Single-$h$    & 32 & 32 & 8   & SC-AIRL $\pi^{h=5}$ proxy; no mixture \\
    Episode-AIRL  & 32 & 32 & 8   & SC-AIRL policies; depth weights fixed per episode \\
    Option-GAIL   & 32 & 32 & 256 & Per-mode $D_c(s)$; entropy $0.02$; see capacity note below \\
    \textbf{SC-AIRL} & 32 & 32 & 8   & State-conditional router; GAE-$h$ truncation \\
    \bottomrule
  \end{tabular}
  \caption{Network architecture summary (pedestrian task). All per-depth/per-mode policy and value networks use two hidden layers with hidden width 32 (256 for GAIL, its own configuration) and LeakyReLU activation.}
  \label{tab:baseline-hp}
\end{table}

\paragraph{Option-GAIL discriminator capacity.}
On the pedestrian task, Option-GAIL's per-mode discriminator $D_c(s)$ uses $256$ hidden units, the value in the released Option-GAIL code~\citep{jing2021optiongail}.
With SC-AIRL's width $8$ its accuracy stayed near chance in a pilot.

\paragraph{Baseline sensitivity.}
GAIL and Option-GAIL use their own configurations (Table~\ref{tab:baseline-hp}).
To test whether their gap to the reward-learning models reflects under-tuning, we retrained both on the full-data runs ($N{=}52$). The retrained runs use the policy, value, and discriminator widths of SC-AIRL ($32$, $32$, and $8$), its learning rates and batch size, and its $20{,}000$-round budget.
With this matched configuration GAIL reaches NLL $12.27$, Top-1 $0.570$, and KL $0.092$, and Option-GAIL reaches $11.27$, $0.412$, and $0.217$ (own-configuration values in Table~\ref{tab:imitation-insample}).
SC-AIRL has lower NLL and higher Top-1 than each matched baseline for all $52$ participants (paired Wilcoxon, $p{=}3.5\times10^{-10}$).

\paragraph{Option-GAIL on the highway task.}
The highway Option-GAIL row has the worst state-level scores in Table~\ref{tab:imitation} (NLL $15.35$, Top-1 $0.172$).
Its discriminator does not separate expert states from generated ones: final discriminator accuracy has median $0.549$ over the $47$ seed-$7$ runs, against $0.656$ for GAIL on the same participants.
GAIL is the higher of the two for $45$ of the $47$ participants (paired Wilcoxon, $p{=}8.8\times10^{-9}$).
The two configurations differ in the discriminator (a per-mode state-only $D_c(s)$ against GAIL's state--action discriminator), in the state-conditional routing over modes, and in the number of modes ($4$ against $1$).
Their discriminator learning rate ($10^{-3}$), discriminator width ($8$), and entropy coefficient ($0.01$) are identical.
We then ran three pilots of $800$ rounds on three participants: a single mode, an entropy coefficient of $0.002$, and a discriminator learning rate of $3\times10^{-3}$.
The last two leave the final accuracy at $0.518$--$0.564$, within the range of the main runs.
With a single mode, all three participants exceed $0.8$ within the first $70$ rounds; at round $790$ one is above $0.8$ ($0.688$, $0.539$, and $0.812$).
Width $8$ alone does not explain the near-chance accuracy of the four-mode runs, because a single-mode $D_c(s)$ of the same width exceeds $0.8$ in the pilot. Splitting the width-$8$ discriminator over four modes is the more likely cause, and the pedestrian runs, which use width $256$, do not show this failure.
We therefore report the row as measured and draw no conclusion about Option-GAIL beyond the configurations we ran.

\subsection{Compute resources}\label{sec:appendix-compute}
All training runs were performed on CPU only; no GPU was used.
Hardware was a heterogeneous internal cluster of five Linux nodes (Intel Xeon, 72--112 logical cores, 125--503\,GB RAM per node; representative model: Xeon Gold 6152 @ 2.10\,GHz) and a single Apple~M4~Max workstation (14 cores, 36\,GB RAM).
On the pedestrian task, per-participant wall-clock time on a cluster node is up to 75\,hours for SC-AIRL.
On the same host, one AIRL round of SC-AIRL takes 3.9 to 4.5 times as long as one round of AIRL (non-shaping) (M4: 4.08 vs.\ 1.04\,s; Host~B: 14.0 vs.\ 3.23\,s; Host~C: 15.5 vs.\ 3.46\,s).
The shaping term changes the per-round time by a factor of 0.84 to 1.07.
A matched sequential measurement on the M4 workstation (participant P1000 on the pedestrian task, P302 on the highway task; one run each) gives 4.4 and 4.0 times the per-round time of Standard AIRL.
Placing the per-depth policies on separate cores lowers the ratio to Standard AIRL to 1.10 on the highway task and 0.68 on the pedestrian task (same M4 measurement) and leaves the CPU time unchanged.
A shared-trunk policy or fewer depths would reduce this cost.
The highway cohort was trained with this placement.
On the three cluster hosts where SC-AIRL and AIRL (non-shaping) were trained in the same launch, a round of SC-AIRL takes $1.20$--$1.22\times$ as long as a round of AIRL (non-shaping) (Table~\ref{tab:hw-time}).
This corresponds to $15$--$22$ hours per participant for SC-AIRL and $12$--$18$ hours for AIRL (non-shaping).
On the pedestrian task, GAIL and Option-GAIL take approximately 0.5--2.5\,hours per participant.

\begin{table}[h]
  \centering\small
  \begin{tabular}{@{}lccc@{}}
    \toprule
    Host & SC-AIRL (s/round) & AIRL (non-shaping) (s/round) & Ratio \\
    \midrule
    A ($n{=}10$) & 26.4 [26.0, 26.9] & 22.0 [20.9, 22.7] & 1.20 \\
    B ($n{=}16$) & 20.5 [19.1, 20.8] & 17.1 [15.8, 17.7] & 1.20 \\
    C ($n{=}21$) & 18.0 [17.2, 18.3] & 14.7 [14.2, 15.3] & 1.22 \\
    \bottomrule
  \end{tabular}
  \caption{Training time per round on the highway task (median over participants [interquartile range]; $n$ participants per host, both models trained in the same launch on each host). SC-AIRL runs its four per-depth policies on separate cores, so its ratio is a wall-clock ratio at four times the core count. Times from different hosts are not compared because concurrent load was not recorded.}
  \label{tab:hw-time}
\end{table}

\section{Held-out evaluation details}\label{sec:appendix-holdout}

This appendix expands the pedestrian block of Table~\ref{tab:imitation}.
The held-out unit is the episode: $8$ of the $40$ shared map seeds are withheld (\S\ref{sec:setup-holdout}). Every model is trained on the other $32$ shared seeds together with each participant's non-shared episodes, and scores are computed on the $8$ withheld seeds for $N{=}52$ participants.
The SC-AIRL router is the $500$-step fit of \S\ref{sec:method-algo} on training-split states and is not refitted for any column.
Rollout rates average $200$ closed-loop episodes per participant against that participant's $8$ held-out expert episodes.

Table~\ref{tab:heldout-full} lists the eight columns.
With $8$ expert episodes per participant, the expected $|\Delta\mathrm{SR}|$ of a model whose true success rate equals the participant's is $0.122$ from the expert-side sampling alone, so we do not read $|\Delta\mathrm{SR}|$ or $|\Delta\mathrm{CR}|$ as differences between models.
Among the rollout gaps, SC-AIRL is closer to the expert than Episode-AIRL on the NOTHING rate at every seed (three-seed means $0.069$ vs.\ $0.101$; paired Wilcoxon $p{=}5\times10^{-7}$, $43$ of $52$ participants; $p{<}10^{-4}$ at each seed).
The timeout and UP gaps do not differ ($p{\ge}.41$).

The validation split of the fit-length rule (\S\ref{sec:method-algo}), which the early-stopped router of Appendix~\ref{sec:appendix-router-robustness-imitation} also uses, is a random $20\%$ of the training-split transitions; the early-stopped fit stops once the validation mixture NLL has not improved for $20$ epochs and keeps the weights of the best epoch, so the withheld map seeds enter neither the fit nor the stopping rule.
Table~\ref{tab:router-early-stop} gives the validation and training NLL by epoch at the three training seeds.
Validation NLL is lowest near epoch $10$ and rises afterwards while training NLL keeps falling.
The best epoch has a median of $12$ to $13$ (range $4$ to $29$), and no fit reaches the $200$-epoch limit.

\begin{table}[h]\centering\footnotesize\setlength{\tabcolsep}{4pt}
\begin{tabular}{@{}rccccccc@{}}
\toprule
& \multicolumn{3}{c}{Validation NLL} & \multicolumn{3}{c}{Training NLL} & \\
\cmidrule(lr){2-4}\cmidrule(lr){5-7}
Epoch & seed $42$ & seed $7$ & seed $123$ & seed $42$ & seed $7$ & seed $123$ & $N$ fitting \\
\midrule
$1$  & 1.439 & 1.448 & 1.429 & 1.486 & 1.500 & 1.478 & $52$/$52$/$52$ \\
$5$  & 1.362 & 1.369 & 1.353 & 1.359 & 1.375 & 1.352 & $52$/$52$/$52$ \\
$10$ & \textbf{1.351} & \textbf{1.358} & \textbf{1.343} & 1.319 & 1.337 & 1.316 & $52$/$52$/$52$ \\
$15$ & 1.353 & 1.360 & 1.344 & 1.288 & 1.300 & 1.282 & $52$/$52$/$52$ \\
$20$ & 1.360 & 1.367 & 1.351 & 1.260 & 1.273 & 1.252 & $52$/$52$/$52$ \\
$30$ & 1.397 & 1.391 & 1.356 & 1.230 & 1.248 & 1.214 & $39$/$42$/$37$ \\
\bottomrule
\end{tabular}
\caption{\textbf{Validation and training NLL of the early-stopped router fit by epoch (pedestrian, held-out-protocol runs).}
Cells are means over the participants still fitting at that epoch, whose number is given in the last column (seeds $42$/$7$/$123$).
Validation NLL is computed on the random $20\%$ of training-split transitions held out from the fit, never on the withheld map seeds.
Bold marks the lowest mean validation NLL among the listed epochs.
All three training seeds use the same fitting procedure, split, and initialization seed.}
\label{tab:router-early-stop}
\end{table}

\begin{table}[h]\centering\footnotesize\setlength{\tabcolsep}{3pt}
\resizebox{\linewidth}{!}{%
\begin{tabular}{@{}llcccccccc@{}}
\toprule
& & \multicolumn{3}{c}{\textit{State-level}} & \multicolumn{5}{c}{\textit{Rollout: $|$agent rate $-$ expert rate$|$}} \\
\cmidrule(lr){3-5}\cmidrule(lr){6-10}
Model & Seeds & NLL$\downarrow$ & KL$\downarrow$ & Top-1$\uparrow$ & $|\Delta\mathrm{SR}|$ & $|\Delta\mathrm{CR}|$ & $|\Delta\mathrm{TO}|$ & $|\Delta\mathrm{UP}|$ & $|\Delta\mathrm{NOTHING}|$ \\
\midrule
BC$^\dagger$ & 42, 7, 123 & 0.948 & 0.037 & 0.689 & 0.190 [0.185, 0.198] & 0.201 [0.193, 0.213] & 0.031 [0.031, 0.033] & 0.031 [0.030, 0.031] & 0.048 [0.047, 0.048] \\
TTC$^\dagger$ & 42 & 1.170 & 0.016 & 0.485 & 0.514 & 0.545 & 0.031 & 0.030 & 0.047 \\
Standard AIRL & 42, 7, 123 & 15.674 & 0.229 & 0.549 & 0.280 [0.272, 0.285] & 0.267 [0.256, 0.275] & 0.032 [0.028, 0.037] & 0.045 [0.044, 0.047] & 0.157 [0.155, 0.159] \\
AIRL (non-shaping) & 42 & 17.589 & 0.209 & 0.551 & 0.275 & 0.259 & 0.035 & 0.046 & 0.149 \\
GAIL & 42, 7, 123 & 14.220 & 0.266 & 0.460 & 0.506 [0.503, 0.508] & 0.534 [0.530, 0.536] & 0.032 [0.030, 0.033] & 0.078 [0.077, 0.081] & 0.162 [0.153, 0.170] \\
Option-GAIL & 42, 7, 123 & 15.543 & 0.622 & 0.363 & 0.507 [0.505, 0.509] & 0.529 [0.529, 0.531] & 0.034 [0.032, 0.036] & 0.066 [0.061, 0.072] & 0.243 [0.237, 0.253] \\
Single-$h$, $h{=}5$ & 42, 7, 123 & 2.109 & 0.116 & 0.610 & 0.191 [0.184, 0.195] & 0.182 [0.178, 0.190] & 0.031 [0.031, 0.032] & 0.035 [0.031, 0.036] & 0.073 [0.069, 0.077] \\
Episode-AIRL & 42, 7, 123 & 1.656 & 0.116 & 0.617 & 0.188 [0.182, 0.194] & 0.178 [0.170, 0.185] & 0.030 [0.029, 0.031] & 0.036 [0.033, 0.039] & 0.101 [0.099, 0.102] \\
\textbf{SC-AIRL (ours)} & 42, 7, 123 & 1.579 & 0.083 & 0.647 & 0.178 [0.172, 0.190] & 0.167 [0.160, 0.178] & 0.031 [0.031, 0.031] & 0.038 [0.037, 0.038] & 0.069 [0.067, 0.071] \\
\bottomrule
\end{tabular}}
\caption{Full pedestrian held-out table ($N{=}52$, mean over the listed training seeds, with [min, max] for the rollout columns; single-run rows have no range; per-seed state-level values are in Table~\ref{tab:heldout-seeds}).
Rollout columns are per-participant absolute deviations of the agent rate from the held-out expert rate, and lower is closer to the expert.
SR, CR, and TO are the success, collision, and timeout rates, and UP and NOTHING are action rates.
Held-out expert rates are SR $0.519$, CR $0.450$, TO $0.031$, UP $0.414$, NOTHING $0.391$.
$^\dagger$Reference rows without reward learning; BC is the sampling-mode policy with fixed seeds.}
\label{tab:heldout-full}
\end{table}

Table~\ref{tab:heldout-signed} reports the signed deviations.
The reward-learning models exceed the expert success rate ($+0.078$ to $+0.270$), and BC (sampling), TTC, GAIL, and Option-GAIL fall below it.
BC changes sign with the decoding mode ($-0.136$ under sampling, $+0.075$ under argmax) while its $|\Delta\mathrm{SR}|$ stays at $0.190$ and $0.211$, so the BC bias is visible only in the signed columns.

\begin{table}[h]\centering\footnotesize\setlength{\tabcolsep}{4pt}
\resizebox{\linewidth}{!}{%
\begin{tabular}{@{}llcccccc@{}}
\toprule
Model & Seeds & $\Delta\mathrm{SR}$ (signed) & $\Delta\mathrm{CR}$ (signed) & $|\Delta\mathrm{SR}|$ & mean $|\Delta\mathrm{action}|$ & $\Delta\mathrm{SR}{>}0$ (seed 42) & Wilcoxon $p$ (seed 42) \\
\midrule
BC$^\dagger$ (sampling) & 42, 7, 123 & $-0.136$ [$-0.146$, $-0.128$] & $+0.166$ [$+0.158$, $+0.176$] & 0.190 [0.185, 0.198] & 0.024 [0.024, 0.025] & 11/52 & $5.7\times10^{-5}$ \\
BC$^\dagger$ (argmax) & 42, 7, 123 & $+0.075$ [$+0.060$, $+0.093$] & $-0.127$ [$-0.133$, $-0.121$] & 0.211 [0.207, 0.214] & 0.092 [0.090, 0.093] & 31/52 & $.062$ \\
TTC$^\dagger$ & 42 & $-0.514$ & $+0.545$ & 0.514 & 0.028 & 1/52 & $3.6\times10^{-10}$ \\
Standard AIRL & 42, 7, 123 & $+0.270$ [$+0.264$, $+0.278$] & $-0.251$ [$-0.259$, $-0.243$] & 0.280 [0.272, 0.285] & 0.066 [0.065, 0.066] & 46/52 & $7.1\times10^{-9}$ \\
AIRL (non-shaping) & 42 & $+0.268$ & $-0.243$ & 0.275 & 0.063 & 47/52 & $2.0\times10^{-9}$ \\
GAIL & 42, 7, 123 & $-0.506$ [$-0.508$, $-0.502$] & $+0.534$ [$+0.530$, $+0.536$] & 0.506 [0.503, 0.508] & 0.093 [0.089, 0.097] & 1/52 & $3.7\times10^{-10}$ \\
Option-GAIL & 42, 7, 123 & $-0.506$ [$-0.508$, $-0.505$] & $+0.529$ [$+0.529$, $+0.531$] & 0.507 [0.505, 0.509] & 0.119 [0.118, 0.120] & 0/52 & $5.0\times10^{-10}$ \\
Single-$h$, $h{=}5$ & 42, 7, 123 & $+0.102$ [$+0.096$, $+0.106$] & $-0.079$ [$-0.082$, $-0.075$] & 0.191 [0.184, 0.195] & 0.037 [0.034, 0.038] & 33/52 & $.002$ \\
Episode-AIRL & 42, 7, 123 & $+0.123$ [$+0.119$, $+0.126$] & $-0.097$ [$-0.101$, $-0.093$] & 0.188 [0.182, 0.194] & 0.045 [0.043, 0.046] & 35/52 & $1.3\times10^{-4}$ \\
\textbf{SC-AIRL (ours)} & 42, 7, 123 & $+0.078$ [$+0.069$, $+0.087$] & $-0.048$ [$-0.058$, $-0.039$] & 0.178 [0.172, 0.190] & 0.033 [0.033, 0.033] & 34/52 & $.010$ \\
\bottomrule
\end{tabular}}
\caption{Signed outcome deviations (agent rate $-$ expert rate, mean [min, max] over seeds).
The last two columns count participants with a positive signed $\Delta\mathrm{SR}$ at seed $42$ and give the two-sided Wilcoxon signed-rank $p$ for a zero median at seed $42$.
BC (argmax) decodes the same three BC models with the argmax action instead of sampling.
$^\dagger$Reference rows without reward learning.}
\label{tab:heldout-signed}
\end{table}
Table~\ref{tab:heldout-seeds} gives the state-level scores per training seed.
All four reward-learning rows and BC are complete for $52$ participants at each of the three seeds.
SC-AIRL has lower NLL than Single-$h$ and Standard AIRL for $51$--$52$ of $52$ participants at every seed, and lower NLL than Episode-AIRL for $33$, $40$, and $38$ of $52$ at seeds $42$, $7$, and $123$ (paired Wilcoxon, $p{<}.002$).

\begin{table}[h]\centering\footnotesize\setlength{\tabcolsep}{4pt}
\begin{tabular}{@{}lccccccccc@{}}
\toprule
& \multicolumn{3}{c}{Seed 42} & \multicolumn{3}{c}{Seed 7} & \multicolumn{3}{c}{Seed 123} \\
\cmidrule(lr){2-4}\cmidrule(lr){5-7}\cmidrule(lr){8-10}
Model & NLL & Top-1 & KL & NLL & Top-1 & KL & NLL & Top-1 & KL \\
\midrule
BC$^\dagger$ & 0.940 & 0.688 & 0.037 & 0.949 & 0.691 & 0.037 & 0.955 & 0.688 & 0.036 \\
Standard AIRL & 15.268 & 0.544 & 0.231 & 14.944 & 0.553 & 0.231 & 16.809 & 0.551 & 0.223 \\
GAIL & 15.317 & 0.461 & 0.264 & 12.634 & 0.462 & 0.261 & 14.709 & 0.455 & 0.271 \\
Option-GAIL & 14.309 & 0.373 & 0.639 & 17.441 & 0.356 & 0.637 & 14.879 & 0.359 & 0.591 \\
Single-$h$, $h{=}5$ & 2.105 & 0.607 & 0.120 & 2.163 & 0.611 & 0.115 & 2.059 & 0.611 & 0.112 \\
Episode-AIRL & 1.662 & 0.614 & 0.120 & 1.679 & 0.620 & 0.116 & 1.626 & 0.616 & 0.113 \\
\textbf{SC-AIRL (ours)} & 1.593 & 0.646 & 0.085 & 1.594 & 0.651 & 0.081 & 1.549 & 0.643 & 0.082 \\
\bottomrule
\end{tabular}
\caption{State-level held-out scores per training seed ($N{=}52$).
AIRL (non-shaping) and TTC exist at seed $42$ only.
$^\dagger$Reference rows without reward learning.}
\label{tab:heldout-seeds}
\end{table}

Table~\ref{tab:heldout-diff} gives the participant-level differences between SC-AIRL and each baseline.
The 95\% confidence intervals of the differences exclude zero for every baseline in NLL, Top-1, and KL; SC-AIRL is better than every reward-learning and adversarial imitation baseline and worse than BC.

\begin{table}[h]\centering\footnotesize\setlength{\tabcolsep}{3pt}
\resizebox{\linewidth}{!}{%
\begin{tabular}{@{}llcccccc@{}}
\toprule
& & \multicolumn{2}{c}{NLL} & \multicolumn{2}{c}{Top-1} & \multicolumn{2}{c}{KL} \\
\cmidrule(lr){3-4}\cmidrule(lr){5-6}\cmidrule(lr){7-8}
Baseline & Seeds & SC-AIRL $-$ baseline & SC-AIRL better & SC-AIRL $-$ baseline & SC-AIRL better & SC-AIRL $-$ baseline & SC-AIRL better \\
\midrule
Episode-AIRL & 42, 7, 123 & $-0.077$ [$-0.104$, $-0.049$] & 42/52 & $+0.030$ [$+0.025$, $+0.035$] & 51/52 & $-0.033$ [$-0.041$, $-0.027$] & 49/52 \\
Single-$h$, $h{=}5$ & 42, 7, 123 & $-0.530$ [$-0.596$, $-0.468$] & 52/52 & $+0.037$ [$+0.031$, $+0.043$] & 51/52 & $-0.033$ [$-0.041$, $-0.026$] & 47/52 \\
Standard AIRL & 42, 7, 123 & $-14.09$ [$-16.59$, $-11.70$] & 52/52 & $+0.097$ [$+0.086$, $+0.109$] & 52/52 & $-0.146$ [$-0.170$, $-0.124$] & 50/52 \\
GAIL & 42, 7, 123 & $-12.64$ [$-15.57$, $-9.98$] & 52/52 & $+0.187$ [$+0.177$, $+0.197$] & 52/52 & $-0.183$ [$-0.205$, $-0.161$] & 51/52 \\
Option-GAIL & 42, 7, 123 & $-13.96$ [$-15.82$, $-12.33$] & 52/52 & $+0.284$ [$+0.269$, $+0.299$] & 52/52 & $-0.540$ [$-0.607$, $-0.474$] & 51/52 \\
BC$^\dagger$ & 42, 7, 123 & $+0.631$ [$+0.559$, $+0.700$] & 0/52 & $-0.042$ [$-0.048$, $-0.036$] & 2/52 & $+0.046$ [$+0.038$, $+0.055$] & 1/52 \\
\bottomrule
\end{tabular}}
\caption{Participant-level differences between SC-AIRL and each baseline on held-out pedestrian episodes ($N{=}52$).
Each participant's score is averaged over the listed training seeds before the difference is taken.
Cells give the mean difference with its participant-bootstrap 95\% confidence interval ($10{,}000$ resamples) and the number of participants for whom SC-AIRL is better.
Lower is better for NLL and KL, and higher is better for Top-1.
$^\dagger$Reference row without reward learning.}
\label{tab:heldout-diff}
\end{table}

Table~\ref{tab:heldout-train} scores the same held-out models on the training-split expert states, without refitting.
At seed $42$ the Top-1 gap (test $-$ train, per-participant median) orders the models BC, SC-AIRL, Episode-AIRL, Single-$h$, Standard AIRL, AIRL (non-shaping), and TTC from largest to smallest (GAIL and Option-GAIL, outside this ordering, have gaps of $-0.110$ and $-0.017$).
The same order holds at seed $123$, and at seed $7$ only Episode-AIRL and Single-$h$ swap places. The gaps at seeds $7$ and $123$ are BC $-0.074$ and $-0.077$, SC-AIRL $-0.060$ and $-0.062$, Episode-AIRL $-0.032$ and $-0.037$, Single-$h$ $-0.035$ and $-0.034$, and Standard AIRL $-0.016$ and $-0.018$.
Neither untruncated AIRL arm has a significant NLL gap ($p{=}.52$ and $.19$).

\begin{table}[h]\centering\footnotesize\setlength{\tabcolsep}{3pt}
\resizebox{\linewidth}{!}{%
\begin{tabular}{@{}lccccc|ccccc@{}}
\toprule
& \multicolumn{5}{c}{NLL} & \multicolumn{5}{c}{Top-1} \\
\cmidrule(lr){2-6}\cmidrule(lr){7-11}
Model & Train & Test & median $\Delta$ & worse & $p$ & Train & Test & median $\Delta$ & worse & $p$ \\
\midrule
BC$^\dagger$ & 0.735 & 0.940 & $+0.195$ & 49/52 & $5.0\times10^{-10}$ & 0.765 & 0.688 & $-0.074$ & 48/52 & $1.2\times10^{-9}$ \\
TTC$^\dagger$ & 1.123 & 1.170 & $+0.049$ & 38/52 & $1.6\times10^{-4}$ & 0.487 & 0.485 & $0.000$ & 26/52 & $.899$ \\
Standard AIRL & 14.186 & 15.268 & $-0.001$ & 26/52 & $.524$ & 0.570 & 0.544 & $-0.026$ & 42/52 & $2.5\times10^{-6}$ \\
AIRL (non-shaping) & 15.622 & 17.589 & $+1.049$ & 31/52 & $.190$ & 0.575 & 0.551 & $-0.022$ & 36/52 & $4.9\times10^{-5}$ \\
GAIL & 12.965 & 15.317 & $+0.255$ & 29/52 & $.071$ & 0.572 & 0.461 & $-0.110$ & 52/52 & $3.5\times10^{-10}$ \\
Option-GAIL & 14.892 & 14.309 & $-0.655$ & 22/52 & $.092$ & 0.390 & 0.373 & $-0.017$ & 38/52 & $8.6\times10^{-4}$ \\
Single-$h$, $h{=}5$ & 2.022 & 2.105 & $+0.067$ & 30/52 & $.070$ & 0.642 & 0.607 & $-0.038$ & 43/52 & $2.3\times10^{-7}$ \\
Episode-AIRL & 1.549 & 1.662 & $+0.063$ & 37/52 & $.002$ & 0.649 & 0.614 & $-0.039$ & 41/52 & $3.7\times10^{-7}$ \\
\textbf{SC-AIRL (ours)} & 1.302 & 1.506 & $+0.171$ & 42/52 & $6.0\times10^{-7}$ & 0.708 & 0.650 & $-0.058$ & 49/52 & $9.9\times10^{-10}$ \\
\bottomrule
\end{tabular}}
\caption{Training-split and held-out scores of the same models (seed $42$, $N{=}52$).
Train and Test are participant means; median $\Delta$ is the per-participant median of test $-$ train; worse counts participants whose test score is worse than their training score; $p$ is from a two-sided Wilcoxon signed-rank test.
SC-AIRL in this table uses the early-stopped router of Appendix~\ref{sec:appendix-router-robustness-imitation}, so its Test NLL and Top-1 ($1.506$, $0.650$) differ from Table~\ref{tab:heldout-seeds} ($1.593$, $0.646$), which uses the $500$-step router.
$^\dagger$Reference rows without reward learning.}
\label{tab:heldout-train}
\end{table}

\section{Effect of excluding potential shaping}\label{sec:appendix-shaping}

Standard AIRL keeps the potential term of \S\ref{sec:method-reward}, which SC-AIRL removes.
The AIRL (non-shaping) arm removes that term from Standard AIRL and changes nothing else, so the pair isolates shaping from the depth structure.
On neither task does the shaping term significantly change the NLL of the untruncated baseline.

On the pedestrian task ($N{=}52$, held-out episodes), the matched seed-$42$ NLL is $15.27$ for Standard AIRL against $17.59$ for AIRL (non-shaping).
The paired difference is not significant ($29$ of $52$ participants, $p{=}.148$), and the other metrics of both arms are in Tables~\ref{tab:heldout-full} and~\ref{tab:heldout-signed}.
SC-AIRL stays an order of magnitude below both on NLL ($1.593$ at seed $42$).

On the highway task ($N{=}47$, two runs per arm), the two arms do not differ on NLL ($4.422$ against $4.375$; $23$ of $47$, $p{=}.89$). Nor do they differ on mean $|\Delta\mathrm{action}|$ ($0.031$ each; $24$ of $47$, $p{=}.76$) or $|\Delta\mathrm{cleared}|$ ($0.123$ each; $23$ of $47$, $p{=}.95$).
Top-1 favors the arm without shaping ($0.629$ against $0.644$; $18$ of $47$, $p{=}.050$).
At seed $7$, the only run both arms share, their learned rewards correlate at $0.885\pm0.080$ (Appendix~\ref{sec:appendix-reward}).

\section{Full-data reference runs}\label{sec:appendix-insample}
The main text reports the held-out protocol of \S\ref{sec:setup-holdout}.
This section keeps the earlier runs on all episodes of every participant, with a $10{,}000$-step post-hoc router, as the reference for the seed-stability and depth-grid comparisons below.
Here and in Appendix~\ref{sec:appendix-router-robustness} the categorical, Gaussian, and stick-breaking routers appear side by side, so we write SC-Cat for the categorical variant that the main text calls SC-AIRL.

\subsection{Imitation fidelity}
\begin{table}[h]
  \centering
  \footnotesize
  \setlength{\tabcolsep}{4pt}
  \resizebox{\linewidth}{!}{%
  \begin{tabular}{@{}lcccccccc@{}}
    \toprule
    & \multicolumn{3}{c}{\textit{State-level (expert states)}} & \multicolumn{5}{c}{\textit{Rollout-level}} \\
    \cmidrule(lr){2-4}\cmidrule(lr){5-9}
    Model & NLL$\downarrow$ & KL$\downarrow$ & Top-1$\uparrow$ & $|\Delta\mathrm{SR}|\downarrow$ & $|\Delta\mathrm{CR}|\downarrow$ & $|\Delta\mathrm{TO}|\downarrow$ & $|\Delta\mathrm{UP}|\downarrow$ & $|\Delta\mathrm{NOTHING}|\downarrow$ \\
    \midrule
    TTC                      & $1.127$          & $0.001$          & $0.486$          & $0.562$          & $0.575$          & $0.013$ & $0.007$ & $0.005$ \\
    BC                     & $0.322$          & $0.019$          & $0.887$          & $0.087$          & $0.096$          & $0.013$ & $0.072$ & $0.092$ \\
    AIRL (non-shaping)       & $16.67$          & $0.077$          & $0.585$          & $0.247$          & $0.237$          & $0.013$ & $0.036$ & $0.142$ \\
    GAIL                     & $10.05$          & $0.194$          & $0.568$          & $0.547$          & $0.557$ & $0.012$ & $0.079$ & $0.163$ \\
    Option-GAIL              & $15.21$          & $0.330$          & $0.387$          & $0.552$          & $0.556$ & $0.015$ & $0.072$ & $0.233$ \\
    \addlinespace[2pt]
    Single-$h$, $h{=}5$      & $2.208$          & $0.031$          & $0.648$          & $0.126$          & $0.119$ & $\mathbf{0.011}$ & $0.029$ & $0.057$ \\
    Episode-AIRL                  & $1.580$          & $0.046$          & $0.652$          & $0.145$          & $0.136$          & $0.012$ & $0.027$ & $0.095$ \\
    \addlinespace[2pt]
    \textbf{Ours SC-Cat}        & $1.410$          & $\mathbf{0.018}$ & $\mathbf{0.768}$ & $0.091$          & $0.084$          & $0.013$ & $0.028$ & $\mathbf{0.043}$ \\
    \textbf{Ours SC-Gauss}      & $1.410$          & $0.038$          & $0.724$          & $\mathbf{0.069}$ & $\mathbf{0.072}$ & $0.013$ & $0.037$ & $0.089$ \\
    \textbf{Ours SC-Stickbreaking} & $\mathbf{1.369}$ & $0.019$          & $0.767$          & $0.100$          & $0.092$ & $0.012$ & $0.028$          & $0.047$ \\
    \textbf{Ours SC-Gauss ($\log h$)} & $1.502$       & $0.019$          & $0.758$          & $0.091$          & $0.082$ & $0.013$ & $\mathbf{0.026}$ & $0.044$ \\
    \bottomrule
  \end{tabular}}
  \caption{Full-data (in-sample) imitation fidelity ($N{=}52$, seed $42$; AIRL (non-shaping), Single-$h$, and Episode-AIRL $N{=}51$).
  Bold marks the best value among the reward-learning models.
  NLL and Top-1 evaluate the mixed policy on expert $(s,a)$ pairs. KL, listed with them, compares the action marginal of the policy's rollouts with the expert action marginal.
  Rollout-level $|\Delta X|$ are mean per-participant absolute deviations from
  each participant's expert rates (cohort means: SR $0.569$, CR $0.418$, TO $0.013$, UP $0.429$, NOTHING $0.405$).
  Single-$h$ uses SC-Cat per-depth $\pi^{h=5}$ as a proxy.
  Episode-AIRL shares policies with SC-Cat and uses depth weights fitted once per participant, independent of the state.
  SC-Gauss uses the fixed-width kernel on the depth index ($\sigma{=}0.5$) and SC-Gauss ($\log h$) the kernel on $\log h$ (\S\ref{sec:appendix-router-families}).}
  \label{tab:imitation-insample}
\end{table}

\subsection{Seed stability of the full-data runs}\label{sec:appendix-seed-stability}

Table~\ref{tab:seed-stability} reports the eight Table~\ref{tab:imitation-insample} metrics for each of the five multi-seed conditions at seeds $42$, $7$, and $123$.
Per-participant means use the canonical $N{=}52$ cohort ($51$ for AIRL (non-shaping) and Episode-AIRL at seed $42$).
Within the SC-AIRL family (SC-Cat, SC-Gauss, SC-Stickbreaking), Top-1 agrees within $0.005$ and every rollout gap within $0.013$ across the three seeds (largest: SC-Stickbreaking $|\Delta\mathrm{SR}|$ $0.100$ to $0.087$), and NLL within $0.049$ (SC-Cat $1.361$--$1.410$).
The largest cross-seed deviations sit on the AIRL (non-shaping) and Episode-AIRL rows (AIRL (non-shaping) NLL $14.52$--$16.67$; Episode-AIRL Top-1 $0.652$ at seed $42$ against $0.704$ at seeds $7$ and $123$).

\begin{table}[h]
  \centering
  \scriptsize
  \setlength{\tabcolsep}{3pt}
  \resizebox{\textwidth}{!}{%
  \begin{tabular}{@{}lcccccccc@{}}
    \toprule
    Condition & NLL & KL & Top-1 & $|\Delta\mathrm{SR}|$ & $|\Delta\mathrm{CR}|$ & $|\Delta\mathrm{TO}|$ & $|\Delta\mathrm{UP}|$ & $|\Delta\mathrm{NOTHING}|$ \\
    \midrule
    \multicolumn{9}{l}{\emph{seed 42 / seed 7 / seed 123}} \\[1pt]
    AIRL (non-shaping)      & $16.67/14.52/14.96$ & $0.077/0.074/0.078$ & $0.585/0.588/0.585$ & $0.247/0.247/0.247$ & $0.237/0.238/0.236$ & $0.013/0.013/0.013$ & $0.036/0.035/0.041$ & $0.142/0.136/0.140$ \\
    Episode-AIRL            & $1.580/1.357/1.310$ & $0.046/0.026/0.026$ & $0.652/0.704/0.704$ & $0.145/0.130/0.119$ & $0.136/0.125/0.112$ & $0.012/0.011/0.012$ & $0.027/0.026/0.029$ & $0.095/0.098/0.098$ \\
    \textbf{SC-Cat}         & $1.410/1.383/1.361$ & $0.018/0.019/0.019$ & $0.768/0.768/0.767$ & $0.091/0.084/0.086$ & $0.084/0.078/0.080$ & $0.013/0.012/0.013$ & $0.028/0.026/0.027$ & $0.043/0.047/0.047$ \\
    \textbf{SC-Gauss}       & $1.410/1.392/1.371$ & $0.038/0.038/0.037$ & $0.724/0.724/0.722$ & $0.069/0.073/0.065$ & $0.072/0.074/0.068$ & $0.013/0.013/0.013$ & $0.037/0.036/0.039$ & $0.089/0.082/0.090$ \\
    \textbf{SC-Stickbreaking} & $1.369/1.349/1.375$ & $0.019/0.021/0.019$ & $0.767/0.768/0.763$ & $0.100/0.091/0.087$ & $0.092/0.084/0.082$ & $0.012/0.013/0.013$ & $0.028/0.025/0.024$ & $0.047/0.047/0.047$ \\
    \bottomrule
  \end{tabular}%
  }
  \caption{Three-seed values for the five multi-seed conditions in Table~\ref{tab:imitation-insample}.
  Each cell reports the per-participant mean at seed $42$ / seed $7$ / seed $123$.
  SC-AIRL rows (SC-Cat, SC-Gauss, SC-Stickbreaking) use post-hoc-fit categorical/Gaussian/stick-breaking routers on the seed-specific converged per-depth policies.
  The Gaussian router is the fixed-width kernel on the depth index ($\sigma{=}0.5$) at every seed.}
  \label{tab:seed-stability}
\end{table}

On these full-data runs, the correlation between the SC-AIRL and AIRL (non-shaping) rewards is $r_{\mathrm{cross}}=0.784\pm0.043$ at seed $42$ and $0.764\pm0.039$ at seed $123$ (mean $\pm$ SD, $N{=}52$), and the SC-AIRL reward correlates at $0.798$ between the two seeds.
The corresponding values on the held-out-protocol runs are in Appendix~\ref{sec:appendix-reward}.

\subsection{Depth-grid robustness}\label{sec:appendix-uniform5}

This appendix repeats the pedestrian full-data training with $\Hset=\{2,4,6,8,10\}$ in place of the paper grid $\{2,3,5,10\}$ (five per-depth policies, even spacing) on the $N{=}52$ cohort.
Within $[2,10]$ an evenly spaced integer grid admits $2$, $3$, $5$, or $9$ depths. Three and five are equally close to the four of the main grid, and we use five.
The seed-$42$ runs are complete ($52/52$) and carry the same $10{,}000$-step post-hoc fitting procedure (the categorical router of one participant, P1002, was fitted for $500$ steps) as Table~\ref{tab:imitation-insample}, here for the categorical, stick-breaking, and $H$-space Gaussian routers.
The seed-$7$ runs finished for $50/52$ participants (P1053 and P1054 did not) and carry the same $10{,}000$-step categorical fit; the seed-$123$ runs finished for $38/52$ and have no such fit.
For seed $7$ we report the $10{,}000$-step fit and the light post-hoc fit ($\le 200$ epochs) for both grids on the $50$ matched participants; seed $123$ is not tabulated.
The held-out protocol of \S\ref{sec:setup-holdout} was not run for this grid.
On the seed-$42$ runs the per-participant expected depth $\mathbb{E}[h]$ of the $10{,}000$-step categorical router agrees across the two grids (Pearson $r{=}0.69$, Spearman $\rho{=}0.68$, $p{<}10^{-7}$, $N{=}52$).
The SC-Cat rewards learned on the two grids correlate at a median $r{=}0.85$ over participants ($0.84\pm0.04$, $2{,}000$ sampled expert states per participant).

At seed $42$ with the $10{,}000$-step router (Table~\ref{tab:uniform5}), the uniform grid lowers NLL by $0.135$ ($p{<}.001$; $39/52$ participants) and raises Top-1 by $0.015$ ($p{<}.001$; $43/52$).
The six other metrics differ by at most $0.008$ ($p{\ge}.056$).
At seed $7$ with the same fit the direction repeats: NLL falls by $0.161$ ($p{<}.001$; $42/50$) and Top-1 rises by $0.015$ ($p{<}.001$; $41/50$), and the six other metrics differ by at most $0.005$ ($p{\ge}.079$).
Under the light fit the NLL direction repeats at seeds $42$ and $7$ ($p{<}.001$ each), and the Top-1 differences are $0.003$ and $0.005$ ($p{=}.083$ and $.070$).
Under the light fit at seed $7$ the uniform grid has larger $|\Delta\mathrm{SR}|$ and $|\Delta\mathrm{CR}|$ by $0.015$ ($p{=}.019$ and $.018$) and smaller $|\Delta\mathrm{UP}|$ by $0.006$ ($p{=}.021$); none of the three differences appears with the $10{,}000$-step fit ($p{\ge}.64$).
Table~\ref{tab:uniform5-traits} repeats, under the uniform grid, the seed-$42$ full-data rows of Table~\ref{tab:router-robust-traits}. All $12$ cells keep their sign, and $7$ of $12$ have $p{<}.05$ against $11$ of $12$ on the paper grid.

\begin{table}[h]
  \centering
  \scriptsize
  \setlength{\tabcolsep}{3pt}
  \resizebox{\textwidth}{!}{%
  \begin{tabular}{@{}llcccccccc@{}}
    \toprule
    & & \multicolumn{3}{c}{\textit{State-level (expert states)}} & \multicolumn{5}{c}{\textit{Rollout-level}} \\
    \cmidrule(lr){3-5}\cmidrule(lr){6-10}
    Grid & Router & NLL$\downarrow$ & KL$\downarrow$ & Top-1$\uparrow$ & $|\Delta\mathrm{SR}|\downarrow$ & $|\Delta\mathrm{CR}|\downarrow$ & $|\Delta\mathrm{TO}|\downarrow$ & $|\Delta\mathrm{UP}|\downarrow$ & $|\Delta\mathrm{NOTHING}|\downarrow$ \\
    \midrule
    \multicolumn{10}{l}{\emph{Seed 42, $10{,}000$-step post-hoc router (as Table~\ref{tab:imitation-insample}), $N{=}52$}} \\[1pt]
    $\{2,3,5,10\}$  & SC-Cat            & $1.410$ & $0.018$ & $0.768$ & $0.091$ & $0.084$ & $0.013$ & $0.028$ & $0.043$ \\
    $\{2,4,6,8,10\}$ & SC-Cat           & $1.274$ & $0.018$ & $0.783$ & $0.083$ & $0.079$ & $0.013$ & $0.026$ & $0.050$ \\
    $\{2,4,6,8,10\}$ & SC-Stickbreaking & $1.220$ & $0.019$ & $0.783$ & $0.105$ & $0.096$ & $0.013$ & $0.023$ & $0.048$ \\
    $\{2,4,6,8,10\}$ & SC-Gauss         & $1.345$ & $0.021$ & $0.767$ & $0.090$ & $0.083$ & $0.013$ & $0.026$ & $0.053$ \\
    \multicolumn{2}{l}{\quad Wilcoxon $p$, SC-Cat, uniform vs.\ paper grid} & $<.001$ & $.199$ & $<.001$ & $.355$ & $.524$ & $.878$ & $.560$ & $.056$ \\
        \addlinespace[2pt]
    \multicolumn{10}{l}{\emph{Seed 7, $10{,}000$-step post-hoc router, $N{=}50$}} \\[1pt]
    $\{2,3,5,10\}$  & SC-Cat            & $1.379$ & $0.019$ & $0.770$ & $0.081$ & $0.077$ & $0.011$ & $0.025$ & $0.045$ \\
    $\{2,4,6,8,10\}$ & SC-Cat           & $1.218$ & $0.014$ & $0.785$ & $0.084$ & $0.079$ & $0.011$ & $0.023$ & $0.044$ \\
    \multicolumn{2}{l}{\quad Wilcoxon $p$, uniform vs.\ paper grid} & $<.001$ & $.079$ & $<.001$ & $.660$ & $.731$ & $.947$ & $.646$ & $.566$ \\
    \midrule
    \multicolumn{10}{l}{\emph{Light post-hoc router ($\le 200$ epochs, both grids), SC-Cat}} \\[1pt]
    $\{2,3,5,10\}$  & seed 42, $N{=}52$ & $1.423$ & $0.029$ & $0.691$ & $0.083$ & $0.084$ & $0.013$ & $0.028$ & $0.063$ \\
    $\{2,4,6,8,10\}$ & seed 42, $N{=}52$ & $1.306$ & $0.031$ & $0.695$ & $0.084$ & $0.079$ & $0.012$ & $0.026$ & $0.074$ \\
    \multicolumn{2}{l}{\quad Wilcoxon $p$} & $<.001$ & $.216$ & $.083$ & $.781$ & $.557$ & $.591$ & $.799$ & $.056$ \\
    \addlinespace[2pt]
    $\{2,3,5,10\}$  & seed 7, $N{=}50$ & $1.393$ & $0.029$ & $0.693$ & $0.075$ & $0.070$ & $0.011$ & $0.028$ & $0.064$ \\
    $\{2,4,6,8,10\}$ & seed 7, $N{=}50$ & $1.296$ & $0.026$ & $0.698$ & $0.090$ & $0.086$ & $0.011$ & $0.022$ & $0.064$ \\
    \multicolumn{2}{l}{\quad Wilcoxon $p$} & $<.001$ & $.328$ & $.070$ & $.019$ & $.018$ & $.386$ & $.021$ & $.653$ \\
    \bottomrule
  \end{tabular}%
  }
  \caption{Imitation metrics on the paper depth grid $\{2,3,5,10\}$ and the uniform grid $\{2,4,6,8,10\}$ (pedestrian task, full-data runs).
  Each cell is a per-participant mean; metric definitions and expert reference rates are those of Table~\ref{tab:imitation-insample}.
  The upper block uses the $10{,}000$-step post-hoc router of Table~\ref{tab:imitation-insample} (the uniform-grid categorical router of one participant, P1002, was fitted for $500$ steps), and its seed-$42$ paper-grid row is that table's SC-Cat row.
  The lower block uses the light fit for both grids; its values are not comparable with the upper block.
  Seed-$7$ rows use the $50$ participants with a finished uniform-grid run.
  $p$: two-sided Wilcoxon signed-rank test over participants, uniform grid vs.\ paper grid, same seed and fit.
  SC-Gauss uses the H-space kernel.}
  \label{tab:uniform5}
\end{table}

\begin{table}[h]
  \centering
  \footnotesize
  \setlength{\tabcolsep}{3pt}
  \resizebox{\linewidth}{!}{%
  \begin{tabular}{@{}llcccccc@{}}
    \toprule
    & & \multicolumn{2}{c}{SC-Cat} & \multicolumn{2}{c}{SC-Stickbreaking} & \multicolumn{2}{c}{SC-Gauss} \\
    \cmidrule(lr){3-4}\cmidrule(lr){5-6}\cmidrule(lr){7-8}
    Measure $\times$ scale & Grid & raw & score-controlled & raw & score-controlled & raw & score-controlled \\
    \midrule
    $I(s;h)\times$STAI        & $\{2,3,5,10\}$   & $-0.358$ (.009) & $-0.284$ (.041) & $-0.306$ (.027) & $-0.223$ (.112) & $-0.333$ (.016) & $-0.294$ (.035) \\
                              & $\{2,4,6,8,10\}$ & $-0.336$ (.015) & $-0.317$ (.022) & $-0.400$ (.003) & $-0.399$ (.003) & $-0.262$ (.061) & $-0.247$ (.077) \\
    \addlinespace[2pt]
    $\Delta h\times$BIS-motor & $\{2,3,5,10\}$   & $-0.375$ (.006) & $-0.380$ (.005) & $-0.283$ (.042) & $-0.289$ (.037) & $-0.362$ (.008) & $-0.361$ (.009) \\
                              & $\{2,4,6,8,10\}$ & $-0.370$ (.007) & $-0.379$ (.006) & $-0.259$ (.064) & $-0.296$ (.033) & $-0.251$ (.073) & $-0.252$ (.072) \\
    \bottomrule
  \end{tabular}%
  }
  \caption{Trait correlations (Spearman $\rho$, $p$ in parentheses, $N{=}52$) at seed $42$ on the two depth grids, full-data runs with the $10{,}000$-step router fit (the uniform-grid categorical router of one participant, P1002, was fitted for $500$ steps).
  Score-controlled: partial correlation controlling the participant's mean task score.
  Measures as in Table~\ref{tab:router-robust-traits}.
  SC-Gauss uses the H-space kernel.}
  \label{tab:uniform5-traits}
\end{table}

\section{Robustness to the router family}\label{sec:appendix-router-robustness}

The main text reports one router, the categorical router of SC-AIRL, written SC-Cat here where it stands beside the other two families.
This section replaces its output family with the stick-breaking and Gaussian families of \S\ref{sec:appendix-router-families} and repeats the analyses of \S\ref{sec:exp-imitation} and \S\ref{sec:exp-clinical}.
In every comparison the per-depth policies $\{\pi^h\}$ and the reward $r_\theta$ of SC-Cat are frozen, and the router alone is re-fitted with the mixture NLL of Eq.~\ref{eq:router-nll}.
Table~\ref{tab:imitation-insample} makes this comparison on the earlier full-data pedestrian runs with a $10{,}000$-step fit.
Table~\ref{tab:router-robust-imitation} uses the held-out protocol of \S\ref{sec:setup-holdout}, a short fit, and both tasks.
The short fit runs for at most $200$ epochs with early stopping on the pedestrian task (learning rate $10^{-3}$, weight decay $10^{-4}$, batch size $256$) and as in Table~\ref{tab:highway-hparams} on the highway task.
The Gaussian family has three kernels, and each analysis below uses a single one.
The held-out pedestrian imitation comparison and the held-out pedestrian trait block use a Gaussian kernel on $\log h$ with a state-dependent width.
The highway comparisons and the full-data pedestrian trait block use a Gaussian kernel on the depth values $h$ with a learned state-dependent width $\sigma(s)\in[0.3,10]$ (H-space), and this kernel is the same at every training seed.
Table~\ref{tab:imitation-insample} reports the fixed-width index kernel and the $\log h$ kernel.

\subsection{Imitation fidelity}\label{sec:appendix-router-robustness-imitation}

Across the three router families, held-out NLL stays within $1.502$--$1.536$ on the pedestrian task and $1.465$--$1.537$ on the highway task (Table~\ref{tab:router-robust-imitation}).
SC-Stickbreaking matches SC-Cat on the pedestrian task ($\Delta\mathrm{NLL}{=}{-}0.003$, paired Wilcoxon $p{=}.53$, and $\Delta\text{Top-1}{=}{-}0.0005$, $p{=}.62$) and on the highway task ($\Delta\mathrm{NLL}{=}{-}0.002$ against the re-fitted categorical router).
SC-Gauss ranks third on the pedestrian task ($\Delta\mathrm{NLL}{=}{+}0.032$, $p{<}10^{-6}$, and $\Delta\text{Top-1}{=}{-}0.005$, $p{=}.009$) and on the highway task ($\Delta\mathrm{NLL}{=}{+}0.070$ and $\Delta\text{Top-1}{=}{-}0.010$ against the re-fitted categorical router).
Every family keeps the NLL margin over the baselines on both tasks, and the rollout margin on the highway task.
On the pedestrian task each family has lower NLL than Standard AIRL for $52$ of $52$ participants, and lower NLL than Episode-AIRL for $52$, $51$, and $45$ of $52$ participants for the re-fit SC-Cat, SC-Stickbreaking, and SC-Gauss routers (paired Wilcoxon, $p{<}10^{-7}$).
On the highway task every family is below Episode-AIRL and AIRL (non-shaping) on NLL, mean action-rate deviation, and outcome gap.
On the pedestrian task no family differs from the re-fitted categorical router on either rollout measure (paired Wilcoxon, $p{\ge}.23$).

\begin{table}[h]\centering\footnotesize\setlength{\tabcolsep}{4pt}
\resizebox{\linewidth}{!}{%
\begin{tabular}{@{}lccccc@{}}
\toprule
& \multicolumn{3}{c}{\textit{State-level}} & \multicolumn{2}{c}{\textit{Rollout}} \\
\cmidrule(lr){2-4}\cmidrule(lr){5-6}
Router & NLL$\downarrow$ & KL$\downarrow$ & Top-1$\uparrow$ & mean $|\Delta\mathrm{action}|\downarrow$ & outcome gap$\downarrow$ \\
\midrule
\multicolumn{6}{l}{\emph{Pedestrian crossing ($N{=}52$, training seed $42$, outcome = success rate)}} \\
SC-Cat      & 1.502 & 0.078 & 0.651 & 0.037 & 0.183 \\
SC-Cat, re-fit             & 1.505 & 0.078 & 0.650 & 0.037 & 0.187 \\
SC-Stickbreaking           & 1.502 & 0.079 & 0.650 & 0.036 & 0.195 \\
SC-Gauss ($\log h$ kernel) & 1.536 & 0.083 & 0.645 & 0.038 & 0.181 \\
\addlinespace[2pt]
Episode-AIRL (reference)   & 1.662 & 0.120 & 0.614 & 0.046 & 0.194 \\
Standard AIRL (reference)  & 15.27 & 0.231 & 0.544 & 0.066 & 0.285 \\
\midrule
\multicolumn{6}{l}{\emph{Highway overtaking ($N{=}47$, mean [min, max] over two training runs, outcome = cleared rate)}} \\
SC-Cat      & 1.470 [1.442, 1.499] & 0.0444 [0.0425, 0.0463] & 0.719 [0.719, 0.720] & 0.0249 [0.0237, 0.0260] & 0.115 [0.108, 0.123] \\
SC-Cat, re-fit             & 1.467 [1.438, 1.496] & 0.0447 [0.0429, 0.0465] & 0.719 [0.718, 0.720] & 0.0250 [0.0240, 0.0259] & 0.116 [0.107, 0.125] \\
SC-Stickbreaking           & 1.465 [1.437, 1.493] & 0.0441 [0.0425, 0.0457] & 0.721 [0.720, 0.721] & 0.0244 [0.0234, 0.0254] & 0.119 [0.111, 0.127] \\
SC-Gauss (H-space kernel)  & 1.537 [1.500, 1.573] & 0.0499 [0.0482, 0.0516] & 0.709 [0.709, 0.710] & 0.0258 [0.0249, 0.0267] & 0.116 [0.115, 0.116] \\
\addlinespace[2pt]
Episode-AIRL (reference)   & 1.719 [1.676, 1.762] & 0.0693 [0.0691, 0.0694] & 0.674 [0.674, 0.674] & 0.0281 [0.0271, 0.0291] & 0.126 [0.122, 0.130] \\
AIRL (non-shaping) (reference) & 4.375 [4.375, 4.375] & 0.0924 [0.0912, 0.0936] & 0.644 [0.642, 0.646] & 0.0313 [0.0310, 0.0317] & 0.123 [0.115, 0.131] \\
\bottomrule
\end{tabular}}
\caption{\textbf{Held-out imitation fidelity by router family.}
All SC rows of a task share the same frozen per-depth policies and reward, and the router is the single re-fitted component.
The main-text fit is the categorical router behind Table~\ref{tab:imitation}, and the re-fit rows use one common fitting procedure and initialization seed for the three families.
Pedestrian rows use training seed $42$ with a fit of at most $200$ epochs and rollouts on the eight held-out map seeds ($200$ episodes per participant) with each family's own router.
Highway rows average the unfixed and seed-$7$ training runs ($47$ of $47$ participants in each), with rollouts on the four held-out scenarios ($150$ episodes per participant) for every router row.
Reference rows are computed on the same runs as the router rows (pedestrian seed $42$; highway unfixed and seed-$7$ runs).
Metrics follow Table~\ref{tab:imitation}.}
\label{tab:router-robust-imitation}
\end{table}

\subsection{Associations with psychometric measures}\label{sec:appendix-router-robustness-traits}

With the $10{,}000$-step fit used throughout this subsection for the pedestrian task, both pedestrian associations stay negative in all three router families of the held-out block, at seeds $42$ and $7$ and in their two-seed means (Table~\ref{tab:router-robust-traits}).
The main text averages three training seeds for SC-Cat, the only family with a seed-$123$ fit in this block, and reports its three-seed means.
For the held-out three-seed $\Delta h\times$BIS-motor mean, the participant-bootstrap 95\% interval excludes zero while the two-sided test gives $p{=}.058$, and we report the test.
Leaving out any one participant keeps the held-out three-seed $I(s;h)\times$STAI correlation between $-0.41$ and $-0.31$, with $p{<}.05$ in all $52$ cases ($51$ of $52$ with the score controlled).
The score-controlled $\Delta h\times$BIS-motor correlation stays between $-0.36$ and $-0.30$, with $p{<}.05$ in all $52$ cases.
The raw $\Delta h\times$BIS-motor correlation stays between $-0.30$ and $-0.23$ and has $p{<}.05$ in $14$ of $52$ cases.
Fit length changes the trait correlations more than the router family does.
With the ${\le}200$-epoch fit of Table~\ref{tab:router-robust-imitation} the two-seed mean $I(s;h)$ of the categorical router falls from $1.12$ to $0.30$ nats and the three seed-$42$ $I(s;h)\times$STAI estimates become $-0.002$, $+0.111$, and $+0.166$ ($p{\ge}.23$), with the score-controlled Gaussian estimate at $+0.294$ ($p{=}.035$), consistent with the parameter recovery of Appendix~\ref{sec:appendix-recovery}.

The full-data three-seed block keeps every sign. Its $I(s;h)\times$STAI magnitudes are larger in all three families, and its $\Delta h\times$BIS-motor magnitudes are similar (seed-mean rows).
In its nine seed-by-family cells raw $p{<}.05$ holds in $8$ for $I(s;h)\times$STAI and $4$ for $\Delta h\times$BIS-motor, and all $18$ are negative.
Participant-bootstrap 95\% intervals ($2{,}000$ resamples) exclude zero in $8$ of these $9$ cells for $I(s;h)\times$STAI (all but SC-Gauss at seed $7$, $[-0.39,+0.09]$) and in $3$ of $9$ for $\Delta h\times$BIS-motor.
With the task score controlled they exclude zero in $4$ of $9$ cells for each measure, and they exclude zero in every seed-mean cell of both measures, raw and controlled.

On the highway task the score-controlled two-run mean for $\Delta h$ and BIS-11 motor impulsivity has $p{<}.05$ in all three families.
The raw two-run mean and the seed-$7$ run alone do not reach $p{<}.05$, and the three families were re-fitted on the first two runs only.
For $d_{\mathrm{car}}$ the score-controlled two-run mean has $p{<}.05$ for SC-Cat and SC-Stickbreaking and not for SC-Gauss.
The per-participant highway shift correlates $0.20$ on average over pairs of the three runs ($+0.12$ to $+0.38$ per pair), against $+0.75$ for the pedestrian shift between seeds $42$ and $7$, so both tasks report the seed average.

\begin{table}[h]\centering\footnotesize\setlength{\tabcolsep}{3pt}
\resizebox{\linewidth}{!}{%
\begin{tabular}{@{}llcccccc@{}}
\toprule
& & \multicolumn{2}{c}{SC-Cat} & \multicolumn{2}{c}{SC-Stickbreaking} & \multicolumn{2}{c}{SC-Gauss} \\
\cmidrule(lr){3-4}\cmidrule(lr){5-6}\cmidrule(lr){7-8}
Measure $\times$ scale & Estimate & raw & score-controlled & raw & score-controlled & raw & score-controlled \\
\midrule
\multicolumn{8}{l}{\emph{Pedestrian ($N{=}52$), full-data runs, $10{,}000$-step fit, H-space Gaussian at every seed}} \\
$I(s;h)\times$STAI & seed 42   & $-0.358$ (.009) & $-0.284$ (.041) & $-0.306$ (.027) & $-0.223$ (.112) & $-0.333$ (.016) & $-0.294$ (.035) \\
                   & seed 7    & $-0.298$ (.032) & $-0.228$ (.104) & $-0.307$ (.027) & $-0.240$ (.087) & $-0.168$ (.233) & $-0.114$ (.421) \\
                   & seed 123  & $-0.410$ (.003) & $-0.359$ (.009) & $-0.360$ (.009) & $-0.297$ (.032) & $-0.392$ (.004) & $-0.350$ (.011) \\
                   & seed mean & $-0.418$ (.002) & $-0.344$ (.013) & $-0.400$ (.003) & $-0.321$ (.020) & $-0.333$ (.016) & $-0.282$ (.043) \\
\addlinespace[2pt]
$\Delta h\times$BIS-motor & seed 42   & $-0.375$ (.006) & $-0.380$ (.005) & $-0.283$ (.042) & $-0.289$ (.037) & $-0.362$ (.008) & $-0.361$ (.009) \\
                          & seed 7    & $-0.196$ (.165) & $-0.211$ (.133) & $-0.224$ (.111) & $-0.258$ (.064) & $-0.172$ (.222) & $-0.194$ (.168) \\
                          & seed 123  & $-0.295$ (.033) & $-0.387$ (.005) & $-0.201$ (.154) & $-0.271$ (.052) & $-0.194$ (.169) & $-0.288$ (.038) \\
                          & seed mean & $-0.343$ (.013) & $-0.387$ (.005) & $-0.314$ (.023) & $-0.368$ (.007) & $-0.287$ (.039) & $-0.332$ (.016) \\
\midrule
\multicolumn{8}{l}{\emph{Pedestrian ($N{=}52$), routers fitted on the training split, $\log h$ Gaussian$^{\dagger}$$^{\ddagger}$}} \\
$I(s;h)\times$STAI & seed 42, $10{,}000$ steps   & $-0.302$ (.030) & $-0.296$ (.033) & $-0.147$ (.300) & $-0.121$ (.394) & $-0.225$ (.110) & $-0.166$ (.239) \\
                   & seed 7, $10{,}000$ steps    & $-0.277$ (.047) & $-0.251$ (.072) & $-0.273$ (.050) & $-0.242$ (.085) & $-0.285$ (.041) & $-0.247$ (.077) \\
                   & seed 123, $10{,}000$ steps  & $-0.284$ (.042) & $-0.223$ (.112) & --- & --- & --- & --- \\
                   & two-seed mean (42, 7)       & $-0.318$ (.021) & $-0.307$ (.027) & $-0.242$ (.084) & $-0.204$ (.146) & $-0.276$ (.048) & $-0.228$ (.104) \\
                   & three-seed mean (SC-Cat only) & $-0.339$ (.014) & $-0.303$ (.029) & --- & --- & --- & --- \\
                   & seed 42, ${\le}200$ epochs  & $-0.002$ (.988) & $+0.055$ (.701) & $+0.111$ (.435) & $+0.134$ (.345) & $+0.166$ (.238) & $+0.294$ (.035) \\
\addlinespace[2pt]
$\Delta h\times$BIS-motor & seed 42, $10{,}000$ steps & $-0.306$ (.027) & $-0.336$ (.015) & $-0.253$ (.070) & $-0.317$ (.022) & $-0.289$ (.037) & $-0.336$ (.015) \\
                          & seed 7, $10{,}000$ steps  & $-0.215$ (.125) & $-0.284$ (.041) & $-0.365$ (.008) & $-0.415$ (.002) & $-0.210$ (.136) & $-0.264$ (.058) \\
                          & seed 123, $10{,}000$ steps & $-0.140$ (.324) & $-0.164$ (.245) & --- & --- & --- & --- \\
                          & two-seed mean (42, 7)     & $-0.278$ (.046) & $-0.334$ (.016) & $-0.339$ (.014) & $-0.402$ (.003) & $-0.275$ (.049) & $-0.342$ (.013) \\
                          & three-seed mean (SC-Cat only) & $-0.265$ (.058) & $-0.323$ (.019) & --- & --- & --- & --- \\
\midrule
\multicolumn{8}{l}{\emph{Highway ($N{=}47$), held-out states, $600$-step re-fit, H-space Gaussian in both runs}} \\
$\Delta h\times$BIS-motor & unfixed & $+0.069$ (.645) & $+0.251$ (.089) & $+0.123$ (.408) & $+0.245$ (.098) & $+0.204$ (.170) & $+0.343$ (.018) \\
                          & seed 7           & $+0.173$ (.246) & $+0.155$ (.300) & $+0.103$ (.490) & $+0.156$ (.294) & $-0.004$ (.981) & $+0.021$ (.891) \\
                          & two-run mean     & $+0.186$ (.210) & $+0.306$ (.036) & $+0.197$ (.185) & $+0.321$ (.028) & $+0.212$ (.152) & $+0.333$ (.022) \\
\addlinespace[2pt]
$d_{\mathrm{car}}\times$BIS-motor & unfixed & $+0.250$ (.091) & $+0.344$ (.018) & $+0.312$ (.032) & $+0.370$ (.011) & $+0.240$ (.105) & $+0.332$ (.023) \\
                                  & seed 7           & $+0.089$ (.552) & $+0.015$ (.921) & $+0.028$ (.854) & $+0.016$ (.917) & $-0.051$ (.731) & $-0.098$ (.512) \\
                                  & two-run mean     & $+0.273$ (.063) & $+0.325$ (.026) & $+0.284$ (.053) & $+0.359$ (.013) & $+0.192$ (.195) & $+0.242$ (.102) \\
\bottomrule
\end{tabular}}
\caption{\textbf{Trait associations by router family.}
Cells give Spearman $r_s$ with the two-sided $p$ in parentheses, and the score-controlled columns give the partial Spearman correlation with the mean task score controlled.
On both tasks $\Delta h = \mathbb{E}[h\,|\,\text{danger}] - \mathbb{E}[h\,|\,\text{safe}]$.
On the pedestrian task danger states are on-road states with an adjacent vehicle and safe states are off-road states.
On the highway task danger states are the nearest $20\%$ of lead-vehicle gaps and safe states are states without a lead vehicle within $105$ distance units, and $d_{\mathrm{car}}$ contrasts states with and without a lead vehicle.
$I(s;h)$ follows Eq.~\ref{eq:mi-state-depth} and is computed on the full expert state set of each participant.
Seed-mean rows average the per-participant measure over training seeds before the correlation is computed, and single-seed rows use one training run.
Routers of different families in one row share the same frozen policies, so the three families' estimates in a row are dependent and are reported without multiplicity correction.
The highway SC-Cat column is the categorical router re-fitted with the procedure of the other two families.
The separately initialized categorical fit used in the main text gives the raw and score columns of Table~\ref{tab:router-risk-control-hw} for $\Delta h$, and $+0.260$ (.078) raw and $+0.338$ (.020) score-controlled for the three-run mean of $d_{\mathrm{car}}$.
$^{\dagger}$The Gaussian router of this block uses a kernel on $\log h$, a different model class from the H-space kernel of the other two blocks.
$^{\ddagger}$The seed-$7$ stick-breaking and $\log h$ Gaussian routers of this block, and six of its seed-$42$ stick-breaking routers, were fitted with fit seed $0$, and its other routers, including the seed-$123$ categorical routers, were fitted without a fixed fit seed. The stick-breaking and $\log h$ Gaussian families have no seed-$123$ fit (---).}
\label{tab:router-robust-traits}
\end{table}

\paragraph{Immediate risk and difficulty as controls.}
A router that only read the immediate risk of a state would give a participant who visits riskier states a different $I(s;h)$ and $\Delta h$ for that reason alone.
We therefore repeat the two pedestrian tests of \S\ref{sec:exp-clinical} with each participant's mean exposure to immediate risk and to task difficulty partialled out, by the same rank-residual partial Spearman used for the task-score control.
Exposure is the mean over the participant's expert states of five raw state channels (Table~\ref{tab:state-schema}).
Three measure risk: the closeness of the nearest same-row vehicle (maximum of head and tail closeness, indices 89, 90, 93, 94), vehicle density (17), and the number of approaching vehicles (sum of 18--23).
Two measure difficulty: the time remaining (0) and the closeness to unreachable tiles (sum of 1--4).

Table~\ref{tab:router-risk-control} reports the three-seed mean of the $10{,}000$-step categorical routers.
Partialling out risk changes $I(s;h)\times$STAI by $0.03$ ($r_s{=}{-}0.339\to{-}0.310$), and adding difficulty strengthens it ($-0.383$, $p{=}.005$, bootstrap 95\% CI $[-0.61,-0.11]$). The risk-only cell is above $\alpha{=}.05$ in each seed taken alone ($p{=}.052$--$.075$), and the risk-plus-difficulty cell is below it in all three ($p{\le}.025$).
$\Delta h\times$BIS-motor moves from $-0.265$ ($p{=}.058$) to $-0.329$ with risk and $-0.332$ ($p{=}.016$) with risk and difficulty. Its bootstrap CI includes zero ($[-0.59,+0.02]$), and the seed-123 cell is non-significant with and without controls.
The controls carry between-participant variance: they remove $8$--$11\%$ of the between-participant rank variance of $I(s;h)$ and $25$--$31\%$ of that of $\Delta h$ (three-seed means).
The five controls correlate weakly with STAI ($|r_s|{\le}0.17$). Vehicle density correlates with BIS-motor ($r_s{=}{-}0.34$, $p{=}.015$) despite a between-participant range of only $0.37$--$0.40$.

\begin{table}[h]\centering\footnotesize\setlength{\tabcolsep}{3pt}
\resizebox{\linewidth}{!}{%
\begin{tabular}{@{}llcccccc@{}}
\toprule
Measure $\times$ scale & Estimate & raw & score & risk & risk $+$ difficulty & risk $+$ difficulty $+$ score & 95\% CI (risk $+$ difficulty) \\
\midrule
$I(s;h)\times$STAI & seed 42 & $-0.302$ (.030) & $-0.296$ (.033) & $-0.267$ (.055) & $-0.324$ (.019) & $-0.302$ (.029) & [$-0.55$, $-0.03$] \\
 & seed 7 & $-0.277$ (.047) & $-0.251$ (.072) & $-0.271$ (.052) & $-0.343$ (.013) & $-0.310$ (.025) & [$-0.60$, $-0.05$] \\
 & seed 123 & $-0.284$ (.042) & $-0.223$ (.112) & $-0.249$ (.075) & $-0.311$ (.025) & $-0.253$ (.071) & [$-0.55$, $-0.01$] \\
 & three-seed mean & $-0.339$ (.014) & $-0.303$ (.029) & $-0.310$ (.025) & $-0.383$ (.005) & $-0.342$ (.013) & [$-0.61$, $-0.11$] \\
\addlinespace[2pt]
$\Delta h\times$BIS-motor & seed 42 & $-0.306$ (.027) & $-0.336$ (.015) & $-0.341$ (.013) & $-0.341$ (.013) & $-0.355$ (.010) & [$-0.59$, $-0.01$] \\
 & seed 7 & $-0.215$ (.125) & $-0.284$ (.041) & $-0.274$ (.050) & $-0.273$ (.050) & $-0.314$ (.024) & [$-0.55$, $+0.06$] \\
 & seed 123 & $-0.140$ (.324) & $-0.164$ (.245) & $-0.164$ (.246) & $-0.163$ (.248) & $-0.168$ (.234) & [$-0.41$, $+0.13$] \\
 & three-seed mean & $-0.265$ (.058) & $-0.323$ (.019) & $-0.329$ (.017) & $-0.332$ (.016) & $-0.362$ (.008) & [$-0.59$, $+0.02$] \\
\bottomrule
\end{tabular}}
\caption{\textbf{Immediate-risk and difficulty controls for the pedestrian trait associations ($N{=}52$).}
Held-out-protocol runs with the $10{,}000$-step post-hoc categorical router; cells give the Spearman $r_s$ (partial Spearman in the controlled columns) with the two-sided $p$ in parentheses, and each hypothesis is tested at $\alpha{=}.05$.
Risk and difficulty are defined in the text; score is the mean task score.
Three-seed rows average the per-participant measures over training seeds $42$, $7$, and $123$ before the correlation.
The last column is the participant-bootstrap 95\% interval ($2{,}000$ resamples) of the risk-plus-difficulty cell.}
\label{tab:router-risk-control}
\end{table}

\paragraph{Highway task.}
We repeat the highway test of \S\ref{sec:exp-clinical} ($\Delta h\times$BIS-motor, $N{=}47$) with the analogous controls, computed on the held-out expert states on which $\Delta h$ is measured (Table~\ref{tab:highway-state}).
Risk has three components. The first is the closeness of the nearest car in the participant's own lane, ahead or behind (the larger of $1-s_{5+\mathrm{lane}}$ and $1-s_{8+\mathrm{lane}}$). The other two are the share of the six forward and rear slots with a car in range and the number of lanes with a car ahead that is slower than the participant. Difficulty is the fuel remaining ($1-t/299$ at step $t$ of the episode) and the share of states in an edge lane, where one lane change is unavailable.
Table~\ref{tab:router-risk-control-hw} reports the three training runs and their mean.
Partialling out risk and difficulty lowers the association from $+0.290$ ($p{=}.048$) to $+0.230$ ($p{=}.121$, bootstrap 95\% CI $[-0.10,+0.54]$), and no single run is below $\alpha{=}.05$ ($p{=}.25$--$.36$).
With the task score also partialled out, it is $+0.442$ ($p{=}.002$), against $+0.423$ ($p{=}.003$) for the score control alone.
The controls remove $14\%$ of the between-participant rank variance of $\Delta h$ (three-run mean).
Of the five controls, only the number of slower cars ahead correlates with BIS-motor ($r_s{=}{-}0.324$, $p{=}.026$). Without it, the risk-plus-difficulty cell is $+0.257$ ($p{=}.081$).

\begin{table}[h]\centering\footnotesize\setlength{\tabcolsep}{3pt}
\resizebox{\linewidth}{!}{%
\begin{tabular}{@{}lcccccc@{}}
\toprule
Estimate & raw & score & risk & risk $+$ difficulty & risk $+$ difficulty $+$ score & 95\% CI (risk $+$ difficulty) \\
\midrule
unfixed & $+0.228$ (.123) & $+0.343$ (.018) & $+0.172$ (.247) & $+0.163$ (.274) & $+0.308$ (.035) & [$-0.17$, $+0.45$] \\
seed 7 & $+0.182$ (.221) & $+0.177$ (.235) & $+0.138$ (.357) & $+0.171$ (.252) & $+0.211$ (.155) & [$-0.16$, $+0.52$] \\
seed 123 & $+0.200$ (.178) & $+0.365$ (.012) & $+0.123$ (.412) & $+0.136$ (.362) & $+0.358$ (.013) & [$-0.16$, $+0.45$] \\
three-run mean & $+0.290$ (.048) & $+0.423$ (.003) & $+0.215$ (.147) & $+0.230$ (.121) & $+0.442$ (.002) & [$-0.10$, $+0.54$] \\
\bottomrule
\end{tabular}}
\caption{\textbf{Immediate-risk and difficulty controls for the highway trait association ($\Delta h\times$BIS-motor, $N{=}47$).}
Rows use the categorical post-hoc routers of the main text for the three highway training runs. Cells, controls, and the last column are as in Table~\ref{tab:router-risk-control}.
The three-run row averages the per-participant $\Delta h$ over the three runs before the correlation.}
\label{tab:router-risk-control-hw}
\end{table}

\paragraph{Simple action statistics as controls.}
The depth shift could restate how often a participant waits or moves in danger states, so we also partial out summaries of each participant's own actions on the state set of each measure.
The first set has the rates of three actions in danger states and the action entropy there (pedestrian: NOTHING, UP, and lateral moves; highway: SLOWER, FASTER, and lane changes), and the second has the danger-minus-safe differences in the same three rates.
These summaries explain $22$--$35\%$ of the between-participant rank variance of $\Delta h$ on the pedestrian task, $8$--$30\%$ on the highway task, and at most $6\%$ of that of $I(s;h)$ on the pedestrian task, but none of them correlates with BIS-motor on either task ($p{\ge}.10$).
With the task score also controlled, both pedestrian associations and the highway association stay significant under every set (Table~\ref{tab:behavior-control}).
Without the score control, $I(s;h)\times$STAI stays significant under every set, whereas $\Delta h\times$BIS-motor, at $p{=}.058$ and $.048$ without controls, loses significance under the difference sets on the pedestrian task and under every action set on the highway task.
If deeper planning itself produces these action differences, part of what the controls remove is signal rather than confound. These controls were defined after the main analysis.

\begin{table}[h]\centering\footnotesize\setlength{\tabcolsep}{3pt}
\resizebox{\linewidth}{!}{%
\begin{tabular}{@{}lcccccc@{}}
\toprule
& \multicolumn{2}{c}{Pedestrian $I(s;h)\times$STAI} & \multicolumn{2}{c}{Pedestrian $\Delta h\times$BIS-motor} & \multicolumn{2}{c}{Highway $\Delta h\times$BIS-motor} \\
\cmidrule(lr){2-3}\cmidrule(lr){4-5}\cmidrule(lr){6-7}
Controls & raw & score & raw & score & raw & score \\
\midrule
none (main text)               & $-0.339$ (.014) & $-0.303$ (.029) & $-0.265$ (.058) & $-0.323$ (.019) & $+0.290$ (.048) & $+0.423$ (.003) \\
risk $+$ difficulty            & $-0.383$ (.005) & $-0.342$ (.013) & $-0.332$ (.016) & $-0.362$ (.008) & $+0.230$ (.121) & $+0.442$ (.002) \\
danger-state action rates      & $-0.336$ (.020) & $-0.296$ (.043) & $-0.330$ (.022) & $-0.403$ (.005) & $+0.268$ (.082) & $+0.395$ (.010) \\
danger-minus-safe differences  & $-0.365$ (.010) & $-0.333$ (.021) & $-0.241$ (.095) & $-0.297$ (.040) & $+0.253$ (.098) & $+0.422$ (.005) \\
both action sets               & $-0.349$ (.019) & $-0.316$ (.037) & $-0.276$ (.067) & $-0.352$ (.019) & $+0.254$ (.114) & $+0.423$ (.007) \\
\bottomrule
\end{tabular}}
\caption{\textbf{Trait associations with immediate risk, task difficulty, or simple action statistics controlled.}
Cells give the Spearman $r_s$ (partial Spearman when controls are present) of the three-seed or three-run mean measure, with the two-sided $p$ in parentheses computed with $n{-}2{-}k$ degrees of freedom for $k$ controls (the first row repeats the values of \S\ref{sec:exp-clinical}); the score columns add the mean task score as a control ($N{=}52$ pedestrian, $N{=}47$ highway).
The risk and difficulty rows repeat Tables~\ref{tab:router-risk-control} and~\ref{tab:router-risk-control-hw}.
The action-rate set has four controls (three danger-state action rates and the danger-state action entropy), the difference set has three, and the combined set has seven.}
\label{tab:behavior-control}
\end{table}

\section{Reward comparison details}\label{sec:appendix-reward}

This appendix extends the reward comparison of \S\ref{sec:exp-reward} to held-out expert states, to the AIRL (non-shaping) arm of Remark~1 (Appendix~\ref{sec:appendix-theory}), and to the highway task.
$r_{\mathrm{cross}}$ is the per-participant Pearson correlation between two reward heads over all five actions at every state of a set (median $4{,}535$ training states and $362$ held-out states per participant).

\paragraph{Cross-method and between-seed correlations.}
On held-out states the SC-AIRL versus Standard AIRL value is below both methods' between-seed values at every seed (Table~\ref{tab:reward-rcross}).
Averaged over the three seeds, the cross-method value is below the SC-AIRL between-seed mean for $46$ of $52$ participants on training states and for $43$ of $52$ on held-out states (per seed and pair: $38$--$47$ and $36$--$44$).
It is below the Standard AIRL between-seed mean for all $52$ on both state sets.

\begin{table}[h]
  \centering
  \footnotesize
  \setlength{\tabcolsep}{4pt}
  \resizebox{\linewidth}{!}{%
  \begin{tabular}{@{}llcccccc@{}}
    \toprule
    & & \multicolumn{3}{c}{Training states} & \multicolumn{3}{c}{Held-out states} \\
    \cmidrule(lr){3-5}\cmidrule(lr){6-8}
    Pair & Seed(s) & Mean $\pm$ SD & Median & Min & Mean $\pm$ SD & Median & Min \\
    \midrule
    \multicolumn{8}{l}{\emph{Pedestrian ($N{=}52$; held-out-protocol runs)}} \\
    SC-AIRL vs.\ Standard AIRL              & $42$            & $0.754\pm0.042$ & $0.757$ & $0.674$ & $0.733\pm0.051$ & $0.742$ & $0.592$ \\
    SC-AIRL vs.\ Standard AIRL              & $7$             & $0.750\pm0.043$ & $0.759$ & $0.646$ & $0.724\pm0.057$ & $0.730$ & $0.525$ \\
    SC-AIRL vs.\ Standard AIRL              & $123$           & $0.748\pm0.037$ & $0.749$ & $0.661$ & $0.722\pm0.052$ & $0.724$ & $0.569$ \\
    SC-AIRL vs.\ Standard AIRL              & 3-seed mean     & $0.751\pm0.031$ & $0.754$ & $0.677$ & $0.727\pm0.043$ & $0.730$ & $0.569$ \\
    SC-AIRL vs.\ AIRL (non-shaping)       & $42$            & $0.778\pm0.047$ & $0.786$ & $0.662$ & $0.751\pm0.060$ & $0.761$ & $0.592$ \\
    Standard AIRL vs.\ AIRL (non-shaping) & $42$            & $0.807\pm0.040$ & $0.806$ & $0.704$ & $0.800\pm0.043$ & $0.802$ & $0.704$ \\
    SC-AIRL, between seeds                  & $42$ vs.\ $7$   & $0.798\pm0.037$ & $0.802$ & $0.712$ & $0.778\pm0.051$ & $0.773$ & $0.658$ \\
    SC-AIRL, between seeds                  & $42$ vs.\ $123$ & $0.799\pm0.040$ & $0.802$ & $0.717$ & $0.772\pm0.058$ & $0.783$ & $0.593$ \\
    SC-AIRL, between seeds                  & $7$ vs.\ $123$  & $0.796\pm0.032$ & $0.791$ & $0.722$ & $0.765\pm0.038$ & $0.762$ & $0.685$ \\
    SC-AIRL, between seeds                  & 3-pair mean     & $0.798\pm0.031$ & $0.798$ & $0.739$ & $0.772\pm0.040$ & $0.770$ & $0.665$ \\
    Standard AIRL, between seeds            & $42$ vs.\ $7$   & $0.847\pm0.031$ & $0.846$ & $0.735$ & $0.835\pm0.038$ & $0.839$ & $0.682$ \\
    Standard AIRL, between seeds            & $42$ vs.\ $123$ & $0.843\pm0.029$ & $0.844$ & $0.785$ & $0.838\pm0.034$ & $0.848$ & $0.764$ \\
    Standard AIRL, between seeds            & $7$ vs.\ $123$  & $0.850\pm0.029$ & $0.852$ & $0.780$ & $0.837\pm0.035$ & $0.843$ & $0.727$ \\
    Standard AIRL, between seeds            & 3-pair mean     & $0.847\pm0.024$ & $0.843$ & $0.795$ & $0.837\pm0.028$ & $0.840$ & $0.760$ \\
    \midrule
    \multicolumn{8}{l}{\emph{Highway ($N{=}47$; $2{,}000$-state sample of all expert states; left column group)}} \\
    SC-AIRL vs.\ AIRL (non-shaping)       & unfixed          & $0.834\pm0.076$ & $0.849$ & $0.593$ & -- & -- & -- \\
    SC-AIRL vs.\ AIRL (non-shaping)       & $7$              & $0.876\pm0.059$ & $0.886$ & $0.564$ & -- & -- & -- \\
    SC-AIRL, between runs                   & unfixed vs.\ $7$ & $0.876\pm0.041$ & $0.879$ & $0.731$ & -- & -- & -- \\
    SC-AIRL, between runs                   & unfixed vs.\ $123$ & $0.868\pm0.052$ & $0.885$ & $0.742$ & -- & -- & -- \\
    SC-AIRL, between runs                   & $7$ vs.\ $123$ & $0.873\pm0.065$ & $0.891$ & $0.579$ & -- & -- & -- \\
    SC-AIRL, between runs                   & 3-pair mean & $0.873\pm0.044$ & $0.882$ & $0.701$ & -- & -- & -- \\
    AIRL (non-shaping), between runs      & unfixed vs.\ $7$ & $0.854\pm0.056$ & $0.855$ & $0.685$ & -- & -- & -- \\
    SC-AIRL vs.\ Standard AIRL              & $7$              & $0.857\pm0.074$ & $0.854$ & $0.544$ & -- & -- & -- \\
    SC-AIRL vs.\ Standard AIRL              & $123$            & $0.857\pm0.063$ & $0.871$ & $0.701$ & -- & -- & -- \\
    Standard AIRL, between runs             & $7$ vs.\ $123$   & $0.833\pm0.085$ & $0.856$ & $0.415$ & -- & -- & -- \\
    Standard AIRL vs.\ AIRL (non-shaping)   & $7$              & $0.885\pm0.080$ & $0.902$ & $0.529$ & -- & -- & -- \\
    \bottomrule
  \end{tabular}}
  \caption{\textbf{Per-participant reward correlation $r_{\mathrm{cross}}$ between two reward heads $\rth(s,a)$ on the same states.}
  Mean, population SD, median, and minimum over participants.
  Pedestrian rows use the held-out-protocol runs of \S\ref{sec:setup-holdout} and evaluate all training states and all held-out states of each participant; Standard AIRL is the arm with potential shaping, and AIRL (non-shaping) the arm without a potential, trained for seed $42$ only.
SC-AIRL and Standard AIRL are trained for seeds $42$, $7$, and $123$; ``3-seed mean'' and ``3-pair mean'' rows aggregate each participant's mean over the three seeds or the three seed pairs.
  Highway rows use a $2{,}000$-state sample of all expert states per participant, so the held-out columns do not apply; ``unfixed'' is the training run without a fixed seed.}
  \label{tab:reward-rcross}
\end{table}

\begin{figure}[htbp]
  \centering
  \includegraphics[width=\linewidth]{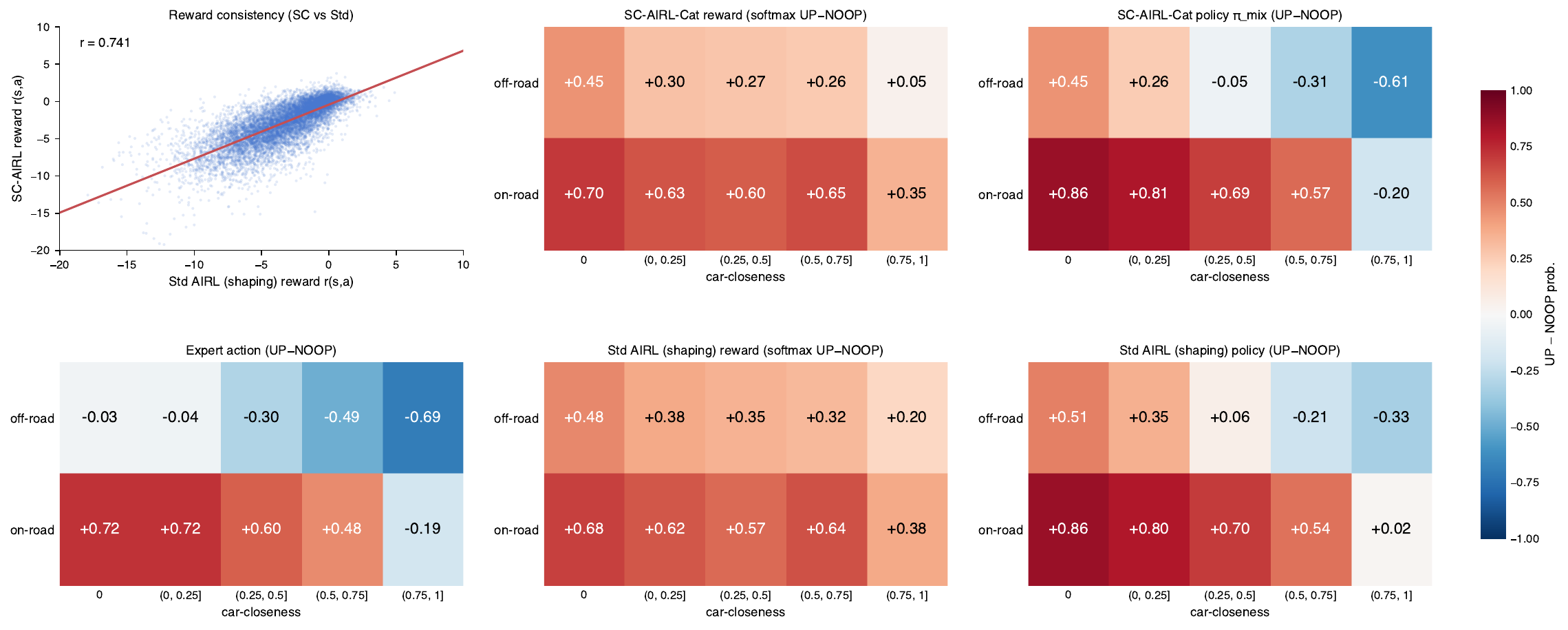}
  \caption{\textbf{Reward, policy, and expert UP$-$NOTHING preferences on the training states of the held-out-protocol runs (pedestrian task, seed $42$, $N{=}52$).}
  Top left: SC-AIRL against Standard AIRL reward head $\rth(s,a)$ over all training states and actions, pooled across participants ($1{,}195{,}050$ pairs; pooled Pearson $r{=}0.741$, a different statistic from the per-participant $r_{\mathrm{cross}}$ of Table~\ref{tab:reward-rcross}).
  Remaining panels, in the cells of Table~\ref{tab:reward-cells}: UP$-$NOTHING preference of the SC-AIRL reward, of the SC-AIRL mixed policy $\pi_{\mathrm{mix}}$ under the post-hoc categorical router, of participants' actions, of the Standard AIRL reward, and of the Standard AIRL policy.}
  \label{fig:reward-heatmap}
\end{figure}

\paragraph{Cell-level comparison.}
Table~\ref{tab:reward-cells} gives the UP$-$NOTHING preference of each reward in the $10$ cells of Figure~\ref{fig:reward-policy}b, and Figure~\ref{fig:reward-heatmap} shows the training-state grids beside the policy and expert grids.
Both rewards prefer UP in all $10$ cells on both state sets, for seed $42$ (Table~\ref{tab:reward-cells}) and for the seed-$7$ and seed-$123$ rewards (reward columns only; policy columns are computed for seed $42$).
On training states, $8$ of $10$ cells differ between methods by at most $0.08$, with a maximum of $0.158$ in the off-road cell with car-closeness in $(0.75,1]$. On held-out states, $8$ of $10$ cells differ by at most $0.08$, with a maximum of $0.136$ in the same cell.
On held-out states the SC-AIRL policy chooses NOTHING more often than UP in $2$ cells, the Standard AIRL policy in $2$, and participants in $4$ (training states: $4$, $2$, and $6$).

\begin{table}[h]
  \centering
  \footnotesize
  \setlength{\tabcolsep}{5pt}
  \resizebox{\linewidth}{!}{%
  \begin{tabular}{@{}llcccccc@{}}
    \toprule
    & & \multicolumn{3}{c}{Training states} & \multicolumn{3}{c}{Held-out states} \\
    \cmidrule(lr){3-5}\cmidrule(lr){6-8}
    Road zone & Car-closeness & SC-AIRL & Standard AIRL & $|\Delta|$ & SC-AIRL & Standard AIRL & $|\Delta|$ \\
    \midrule
    off-road & $0$           & $0.450$ & $0.483$ & $0.033$ & $0.455$ & $0.521$ & $0.066$ \\
    off-road & $(0,0.25]$    & $0.296$ & $0.380$ & $0.084$ & $0.235$ & $0.339$ & $0.104$ \\
    off-road & $(0.25,0.5]$  & $0.275$ & $0.348$ & $0.073$ & $0.456$ & $0.438$ & $0.018$ \\
    off-road & $(0.5,0.75]$  & $0.261$ & $0.322$ & $0.061$ & $0.380$ & $0.389$ & $0.009$ \\
    off-road & $(0.75,1]$    & $0.046$ & $0.204$ & $0.158$ & $0.114$ & $0.250$ & $0.136$ \\
    on-road  & $0$           & $0.704$ & $0.684$ & $0.020$ & $0.741$ & $0.717$ & $0.024$ \\
    on-road  & $(0,0.25]$    & $0.625$ & $0.621$ & $0.004$ & $0.654$ & $0.636$ & $0.018$ \\
    on-road  & $(0.25,0.5]$  & $0.602$ & $0.565$ & $0.037$ & $0.608$ & $0.582$ & $0.026$ \\
    on-road  & $(0.5,0.75]$  & $0.654$ & $0.640$ & $0.015$ & $0.712$ & $0.704$ & $0.008$ \\
    on-road  & $(0.75,1]$    & $0.347$ & $0.383$ & $0.036$ & $0.505$ & $0.527$ & $0.021$ \\
    \bottomrule
  \end{tabular}}
  \caption{\textbf{UP$-$NOTHING preference of the two rewards in the road-zone $\times$ car-closeness cells (pedestrian task, seed $42$, held-out-protocol runs).}
  Each cell is the mean over participants of the participant's state-averaged $\mathrm{softmax}(\rth(s,\cdot))[\mathrm{UP}]-\mathrm{softmax}(\rth(s,\cdot))[\mathrm{NOTHING}]$; a participant enters a cell with at least $5$ states.
  All $52$ participants enter every training-state cell; on held-out states, $35$ enter off-road $(0,0.25]$, $51$ enter on-road $(0,0.25]$ and $(0.25,0.5]$, and $52$ enter the remaining cells.
  $|\Delta|$ is the absolute difference between the two method columns.}
  \label{tab:reward-cells}
\end{table}

\paragraph{Relation between $\fth$ and $\rth$ in Standard AIRL.}
We evaluate $\fth(s,a,s')=\rth(s,a)+\gamma\Phi(s')-\Phi(s)$ and $\rth(s,a)$ on every expert transition of each participant, separately for the training episodes and the held-out episodes of the held-out-protocol runs ($N{=}52$; seeds $42$, $7$, and $123$).
For seed $42$, the per-participant Pearson correlation between $\fth$ and $\rth$ has median $0.805$ on training transitions (mean $0.768\pm0.123$, minimum $0.344$) and $0.823$ on held-out transitions (mean $0.789\pm0.109$, minimum $0.317$). Over the three seeds the medians lie between $0.773$ and $0.823$.
In the decomposition $\mathrm{Var}(\fth)=\mathrm{Var}(\rth)+\mathrm{Var}(\gamma\Phi(s')-\Phi(s))+2\,\mathrm{Cov}$, the potential term accounts for a median $41\%$ of $\mathrm{Var}(\fth)$ on training transitions (mean $45\%$; range $19\%$ to $108\%$) and $40\%$ on held-out transitions (mean $42\%$; range $9\%$ to $95\%$) at seed $42$. The reward term accounts for $96\%$ and $107\%$, and the covariance term for $-39\%$ and $-42\%$. Over the three seeds, the medians of the potential share lie between $35\%$ and $44\%$.

\paragraph{Highway task.}
The highway block of Table~\ref{tab:reward-rcross} gives both untruncated arms ($N{=}47$).
The two untruncated arms agree with each other at $0.885\pm0.080$, the highest mean in the highway block, which is what the shaping term dropping out of $\rth$ predicts.

\section{Router fit length and the individual-difference measures}\label{sec:appendix-router-fitlength}

\begin{figure}[h]\centering
\includegraphics[width=0.80\linewidth]{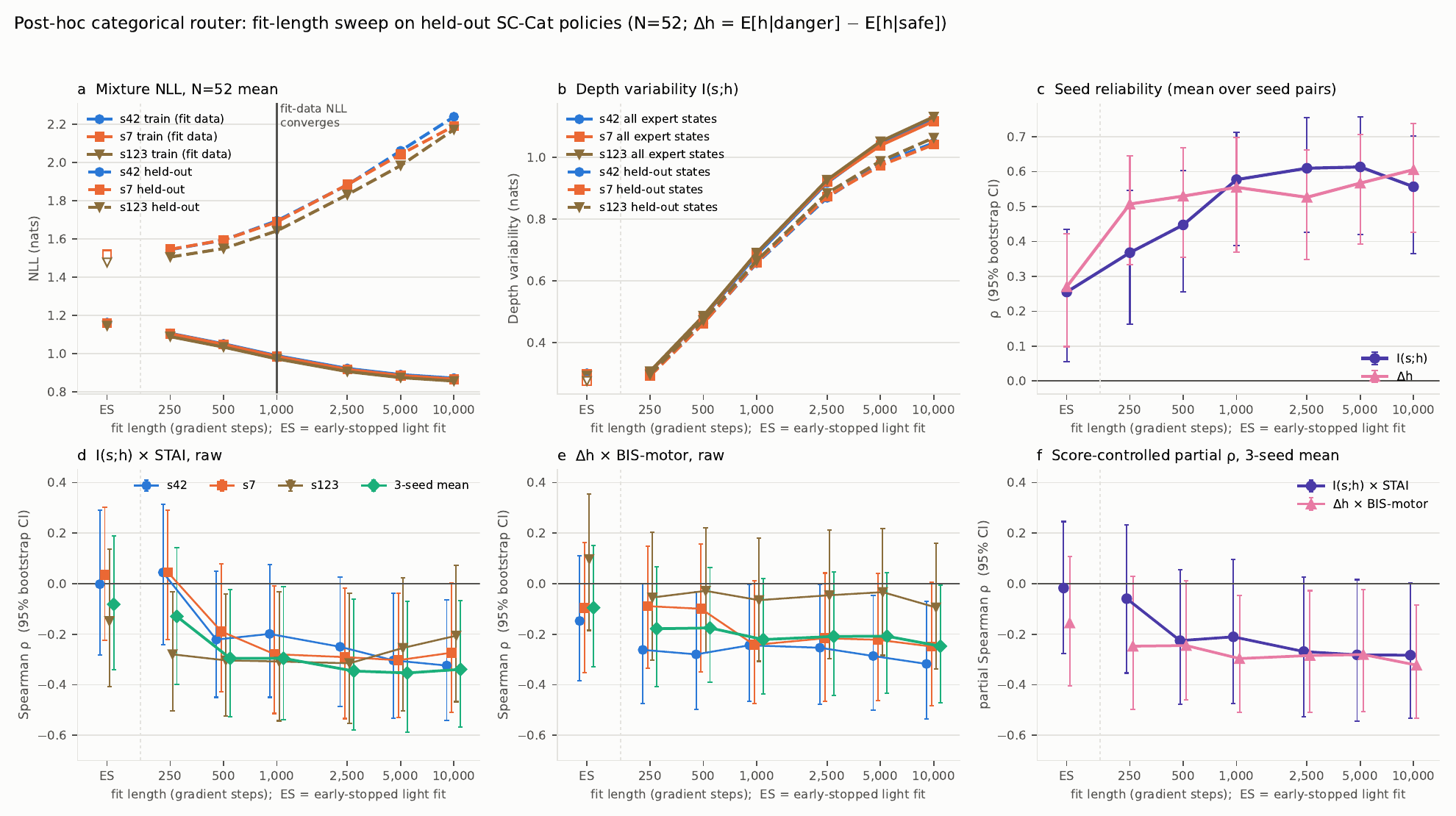}
\caption{\textbf{Router fit-length sweep on the held-out SC-AIRL policies ($N{=}52$).}
The horizontal axis is the fit length in gradient steps on a log scale, and the column marked ES to the left of the dotted break is the early-stopped router.
\textbf{(a)} $N{=}52$ mean mixture NLL on the fit data (solid) and on the held-out map seeds (dashed) for each training seed, with the vertical line at $1{,}000$ steps, where the decrease of the fit-data NLL falls below $2\times10^{-4}$ nats per step.
\textbf{(b)} Mean depth variability $I(s;h)$ on all expert states (solid) and on held-out states (dashed).
\textbf{(c)} Per-participant Spearman reliability for $I(s;h)$ and $\Delta h$, averaged over the three pairs of training seeds, with $95\%$ participant-bootstrap intervals.
\textbf{(d, e)} Spearman correlations of $I(s;h)$ with STAI and of $\Delta h$ with BIS-11 motor impulsivity, per training seed and for the three-seed mean.
\textbf{(f)} The same two correlations for the three-seed mean with the mean task score controlled.
Panels (a)--(f) use the sweep's own $10{,}000$-step snapshot at the right edge, while Table~\ref{tab:router-fitlength} uses the router fits of Appendix~\ref{sec:appendix-router-robustness-traits} at that length.}
\label{fig:router-fitlength}
\end{figure}

Both trait correlations grow in magnitude as the router fit runs longer, up to about $5{,}000$ steps, while the held-out mixture NLL and mean $I(s;h)$ rise. The seed reliability of $I(s;h)$ rises to $0.58$ at $1{,}000$ steps and stays between $0.56$ and $0.61$ up to $10{,}000$, and that of $\Delta h$ stays between $0.51$ and $0.61$ from $250$ steps on (Table~\ref{tab:router-fitlength}, Figure~\ref{fig:router-fitlength}).
These rows come from refitting the categorical router on the held-out training split for $250$ to $10{,}000$ Adam steps of batch size $512$ at three training seeds, with the per-depth policies and the reward frozen throughout.
The early-stopped router of the held-out evaluation is a separate procedure rather than a shorter run, stopping on a validation split after at most $200$ epochs, with weight decay and without mirror augmentation or no-op removal.

The training objective settles earlier than any of these curves.
The decrease of the $N{=}52$ mean train NLL on the fit data falls below $2\times10^{-4}$ nats per step at $1{,}000$ steps in all three training seeds, where that NLL is $0.972$ to $0.988$ nats, against $1.033$ to $1.050$ at $500$ steps and $0.856$ to $0.871$ at $10{,}000$.
The last $9{,}000$ steps move it by about $0.12$ nats while the held-out NLL rises by about $0.5$ nats.

For the sweep's $10{,}000$-step snapshot, a refit that differs only in the router's initialization seed reproduces the per-participant measures, with rank agreement of $0.910$ for $I(s;h)$ at training seeds $42$ and $7$ and $0.908$ and $0.940$ for $\Delta h$. The reliability between these two training seeds is the same whether the two fits share that seed ($0.570$ and $0.708$) or not ($0.577$ and $0.690$).
For the sweep's seeded $10{,}000$-step snapshot routers at the same two training seeds, controlling the number of expert transitions each router was fitted on leaves the anxiety correlation at $\rho = -0.318$ ($p = .022$), against $\rho = -0.346$ ($p = .012$) without that control.

The two cross pairings, $I(s;h)$ with BIS-motor and $\Delta h$ with STAI, do not reach $\alpha{=}.05$ at $10{,}000$ steps.
Over the three training seeds of \S\ref{sec:exp-clinical}, the first is $-0.229$ ($p = .102$) raw and $-0.202$ ($p = .150$) with the task score controlled, and the second is $+0.019$ ($p = .896$) raw and $-0.056$ ($p = .693$) controlled (two-seed values: $-0.219$, $-0.207$, $+0.040$, $-0.029$).

\begin{table}[h]\centering\footnotesize\setlength{\tabcolsep}{3.5pt}
\resizebox{\linewidth}{!}{%
\begin{tabular}{@{}lcccccccc@{}}
\toprule
& & & \multicolumn{2}{c}{\textit{Seed reliability}} & \multicolumn{2}{c}{$I(s;h)\times$STAI} & \multicolumn{2}{c}{$\Delta h\times$BIS-motor} \\
\cmidrule(lr){4-5}\cmidrule(lr){6-7}\cmidrule(lr){8-9}
Fit length & Held-out NLL & $I(s;h)$ (nats) & $I(s;h)$ & $\Delta h$ & raw & controlled & raw & controlled \\
\midrule
Early-stopped      & 1.500 & 0.30 & 0.26 & 0.27 & $-0.082$ (.565) & $-0.017$ (.903) & $-0.095$ (.502) & $-0.156$ (.271) \\
250 steps          & 1.531 & 0.31 & 0.37 & 0.51 & $-0.130$ (.359) & $-0.060$ (.675) & $-0.178$ (.206) & $-0.247$ (.077) \\
500 steps          & 1.579 & 0.48 & 0.45 & 0.53 & $-0.295$ (.033) & $-0.225$ (.108) & $-0.175$ (.214) & $-0.245$ (.080) \\
$1{,}000$ steps     & 1.675 & 0.69 & 0.58 & 0.55 & $-0.296$ (.033) & $-0.210$ (.134) & $-0.221$ (.115) & $-0.296$ (.033) \\
$2{,}500$ steps     & 1.865 & 0.92 & 0.61 & 0.53 & $-0.346$ (.012) & $-0.269$ (.054) & $-0.209$ (.137) & $-0.284$ (.041) \\
$5{,}000$ steps     & 2.029 & 1.04 & 0.61 & 0.57 & $-0.353$ (.010) & $-0.282$ (.043) & $-0.208$ (.139) & $-0.281$ (.044) \\
$10{,}000$ steps    & 2.197 & 1.13 & 0.58 & 0.61 & $-0.339$ (.014) & $-0.303$ (.029) & $-0.265$ (.058) & $-0.323$ (.019) \\
\bottomrule
\end{tabular}}
\caption{\textbf{Post-hoc router fit length, held-out likelihood, and the two trait correlations ($N{=}52$).}
Rows are fit lengths of the categorical router on frozen per-depth policies, and the early-stopped row is the router of Appendix~\ref{sec:appendix-router-robustness-imitation}.
Held-out NLL and $I(s;h)$ are $N{=}52$ means averaged over the three training seeds.
Seed reliability is the mean of the per-participant Spearman correlations over the three pairs of training seeds ($42$, $7$, $123$).
The rule of \S\ref{sec:method-algo} uses ICC(A,k) of the three-seed measures instead, which is $0.68$ and $0.75$ for $I(s;h)$ and $\Delta h$ at $250$ steps and $0.71$ and $0.76$ at $500$ steps.
Correlation cells give the Spearman $\rho$ of the three-seed mean of the per-participant measure with the two-sided $p$ in parentheses, and the controlled columns give the partial Spearman correlation with the mean task score controlled.
$\Delta h$ and $I(s;h)$ are defined as in Table~\ref{tab:router-robust-traits}.
The $500$-step row is the fit selected by the rule of \S\ref{sec:method-algo}, and the $10{,}000$-step row is the set of router fits of Appendix~\ref{sec:appendix-router-robustness-traits}. The sweep's own seeded $10{,}000$-step snapshot, which the figure plots, gives held-out NLL $2.200$, $I(s;h)$ $1.12$ nats, and reliability $0.56$ and $0.61$.
Its correlations are $-0.339$ (.014) and $-0.283$ (.042) for $I(s;h)\times$STAI and $-0.248$ (.077) and $-0.322$ (.020) for $\Delta h\times$BIS-motor, and the per-participant rank agreement between the two fits is $0.82$ to $0.95$.
Per-seed mean $I(s;h)$ at $10{,}000$ steps is $1.126$ at seed $42$, $1.122$ at seed $7$, and $1.136$ at seed $123$, and at the early-stopped fit it is $0.301$, $0.298$, and $0.293$.}
\label{tab:router-fitlength}
\end{table}

The held-out evaluation and the trait analysis both use the $500$-step fit.
Rolling out the $10{,}000$-step routers on the held-out map seeds ($200$ episodes per participant, three training seeds, $N{=}52$) leaves every rollout gap within $0.009$ of the early-stopped router rolled out on the same seeds: $|\Delta\mathrm{SR}|$ $0.191$ vs.\ $0.182$ ($p{=}.17$), $|\Delta\mathrm{CR}|$ $0.178$ vs.\ $0.172$ ($p{=}.28$), $|\Delta\mathrm{TO}|$ $0.031$ vs.\ $0.031$ ($p{=}.37$), $|\Delta\mathrm{NOTHING}|$ $0.067$ vs.\ $0.075$ ($p{=}.051$), and $|\Delta\mathrm{UP}|$ $0.040$ vs.\ $0.038$ ($p{=}.31$; paired Wilcoxon).
The state-level scores are worse, with NLL $2.197$ vs.\ $1.500$, Top-1 $0.639$ vs.\ $0.652$, and KL $0.088$ vs.\ $0.077$ ($p{<}10^{-4}$ each).

On the highway task, the router of both the held-out evaluation and the trait analysis is fit by full-batch Adam steps on each participant's training episodes, and the reported fit has $600$ steps (Table~\ref{tab:router-fitlength-highway}).
Refitting it for $50$ to $2{,}000$ steps at three training runs shows that the seed reliability of $I(s;h)$ rises to $0.71$ at $600$ steps and $0.77$ at $1{,}000$ and falls to $0.55$ at $2{,}000$, while that of $\Delta h$ stays between $0.17$ and $0.28$ at every length.
Held-out NLL is lowest at $400$ to $600$ steps.
The score-controlled correlation of $\Delta h$ with BIS-motor is $+0.32$ to $+0.40$ ($p{<}.05$) from $400$ to $2{,}000$ steps in these refits, and the raw correlation reaches $\alpha{=}.05$ only in the reported fit.

\begin{table}[h]\centering\footnotesize\setlength{\tabcolsep}{4pt}
\begin{tabular}{@{}lcccccc@{}}
\toprule
& & & \multicolumn{2}{c}{\textit{Seed reliability}} & \multicolumn{2}{c}{$\Delta h\times$BIS-motor} \\
\cmidrule(lr){4-5}\cmidrule(lr){6-7}
Fit length & Held-out NLL & $I(s;h)$ (nats) & $I(s;h)$ & $\Delta h$ & raw & controlled \\
\midrule
50 steps            & 1.542 & 0.08 & 0.07 & 0.17 & $+0.105$ (.483) & $+0.198$ (.182) \\
100 steps           & 1.530 & 0.17 & 0.19 & 0.20 & $+0.165$ (.268) & $+0.264$ (.073) \\
200 steps           & 1.519 & 0.24 & 0.26 & 0.21 & $+0.168$ (.258) & $+0.287$ (.050) \\
400 steps           & 1.500 & 0.39 & 0.53 & 0.23 & $+0.244$ (.099) & $+0.394$ (.006) \\
600 steps           & 1.502 & 0.55 & 0.71 & 0.28 & $+0.247$ (.095) & $+0.386$ (.007) \\
$1{,}000$ steps     & 1.527 & 0.76 & 0.77 & 0.22 & $+0.250$ (.090) & $+0.398$ (.006) \\
$2{,}000$ steps     & 1.583 & 0.96 & 0.55 & 0.22 & $+0.196$ (.187) & $+0.315$ (.031) \\
\midrule
600 steps, reported & 1.505 & 0.56 & 0.71 & 0.20 & $+0.290$ (.048) & $+0.423$ (.003) \\
\bottomrule
\end{tabular}
\caption{\textbf{Post-hoc router fit length on the highway task ($N{=}47$).}
Rows above the rule refit the categorical router with a fixed initialization seed for the given number of full-batch Adam steps on frozen per-depth policies, and the last row is the fit reported in \S\ref{sec:exp-clinical}, whose initialization was not seeded.
Held-out NLL is scored on the four held-out scenarios, and all columns average the three training runs (unfixed seed, seeds $7$ and $123$).
Seed reliability is the mean of the per-participant Spearman correlations over the three pairs of training runs.
The rule of \S\ref{sec:method-algo} uses ICC(A,k) of $\Delta h$, which is highest at $600$ steps ($0.50$, against $0.44$--$0.49$ at the other lengths).
Correlation cells give the Spearman $\rho$ of the three-run mean of $\Delta h$ with BIS-11 motor impulsivity, with the two-sided $p$ in parentheses, and the controlled column gives the partial Spearman correlation with the mean task score controlled.
The highway cohort did not complete the STAI, so $I(s;h)\times$STAI is not tested on this task.}
\label{tab:router-fitlength-highway}
\end{table}

\section{Parameter recovery of the individual-difference measures}\label{sec:appendix-recovery}

Parameter recovery tests whether a measure derived from a fitted model returns the value that generated the data~\citep{wilson2019ten}.
We apply it to the router stage of both tasks.
For each participant, the per-depth policies of the trained model and the participant's fitted router form the generating model.
A synthetic agent replays the participant's map seeds (pedestrian) or traffic scenarios (highway) in the participant's episode order, drawing at every step a depth from the router and an action from that depth's policy.
The router is refit to the synthetic behavior with the policies held fixed, with the fit length of \S\ref{sec:method-algo}, and the measures of the refit router are compared with those of the generating router on the same synthetic states.
On the pedestrian task the generating routers are $500$-step fits on the per-depth policies of the full-data seed-$42$ runs, and the refit uses the same length.
On the highway task the generating routers are the $600$-step fits of the three held-out-protocol runs, measures are computed on the held-out scenarios, and the depth shift is averaged over the three runs as in \S\ref{sec:exp-clinical}.

Three contrasts replace each generating router by one whose routing is the same in every state, so that the generating $I(s;h)$ and $\Delta h$ are zero.
Contrast (a) uses the participant's own marginal depth distribution $\bar{p}_i(h)$, keeping it unchanged.
Contrast (b) uses the group-average distribution for every participant, and contrast (c) assigns each participant another participant's $\bar{p}(h)$ under three random permutations; we apply (b) and (c) on the highway task.
Contrasts (b) and (c) thus also remove the link between each participant and their marginal depth distribution.

\paragraph{Recovery.}
The depth shift is recovered in both tasks (Table~\ref{tab:recovery}).
The rank correlation between generating and recovered values is $0.90$ on the pedestrian task and $0.72$--$0.84$ across the three highway runs ($0.77$ for their mean).
Under contrast (a) the recovered shift is smaller than under the generating routers (median $|\Delta h|$ $0.18$ against $0.73$ on the pedestrian task and $0.43$ against $0.97$ on the highway task, where each highway value rests on eight held-out episodes).
For $I(s;h)$ the rank recovery is $0.59$, and under contrast (a) the refit still returns about half of the generating value (median $0.25$ against $0.43$).
$H(\bar{p})$ is recovered with rank correlation $0.83$.

\begin{table}[h]\centering\footnotesize\setlength{\tabcolsep}{3.5pt}
\resizebox{\linewidth}{!}{%
\begin{tabular}{@{}lllccc@{}}
\toprule
Task & Measure & Generating fit & Refit & Rank recovery & Median (generating / contrast (a)) \\
\midrule
Pedestrian & $|\Delta h|$ & $500$ steps & $500$ steps & $0.90$ & $0.73$ / $0.18$ \\
 & $I(s;h)$ & $500$ steps & $500$ steps & $0.59$ & $0.43$ / $0.25$ \\
 & $H(\bar{p})$ & $500$ steps & $500$ steps & $0.83$ & -- \\
\midrule
Highway & $|\Delta h|$ & $600$ steps & $600$ steps & $0.72$, $0.81$, $0.84$ (mean $0.77$) & $0.97$ / $0.43$ \\
 & & $600$ steps & $100$ steps & $0.60$, $0.72$, $0.62$ (mean $0.66$) & $0.49$ / $0.16$ \\
\bottomrule
\end{tabular}}
\caption{\textbf{Parameter recovery of the router-derived measures.}
Rank recovery is the Spearman correlation across participants ($N{=}52$ pedestrian, $N{=}47$ highway) between the measure of the generating router and that of the refit router, computed on the signed $\Delta h$; highway entries give the unfixed, seed-$7$, and seed-$123$ runs and the correlation of their mean.
The last column gives the median over participants under the generating router and under contrast (a), which removes state dependence.
Pedestrian generating routers are $500$-step fits on the full-data seed-$42$ policies; highway values use the held-out scenarios.}
\label{tab:recovery}
\end{table}

\paragraph{Associations with questionnaire scores.}
Table~\ref{tab:recovery-traits} gives the correlations of the recovered measures with the questionnaire scores under the generating routers and the contrasts.
On the pedestrian task the anxiety correlation of the recovered $I(s;h)$ is $-0.17$ under the generating routers and $-0.24$ under contrast (a), so a correlation of this size can arise from the marginal depth distribution alone.
The recovered shift correlates with BIS-motor at $-0.23$ under the generating routers and $-0.13$ under contrast (a).
On the highway task, with the task score controlled, the recovered shift under contrasts (a)--(c) lies between $-0.19$ and $+0.11$, against $+0.28$ under the generating routers and $+0.42$ in the data.
On the highway task the generating routers give $+0.27$ on the synthetic states and the recovered shift $+0.28$; both stay below the data value because the synthetic agents visit a different set of states.

\begin{table}[h]\centering\footnotesize\setlength{\tabcolsep}{4pt}
\resizebox{\linewidth}{!}{%
\begin{tabular}{@{}lccc@{}}
\toprule
& Pedestrian $I(s;h)\times$STAI & Pedestrian $\Delta h\times$BIS-motor & Highway $\Delta h\times$BIS-motor \\
Generating fit & $500$ steps & $500$ steps & $600$ steps, raw / controlled \\
\midrule
Data (\S\ref{sec:exp-clinical}) & $-0.295$ & $-0.175$ & $+0.290$ / $+0.423$ \\
Generating routers & $-0.17$ & $-0.23$ & $+0.16$ / $+0.28$ \\
(a) own average distribution & $-0.24$ & $-0.13$ & $-0.10$ / $-0.01$ \\
(b) group-average distribution & -- & -- & $+0.01$ / $-0.03$ \\
(c) permuted distributions & -- & -- & $-0.20$ to $+0.01$ / $-0.19$ to $+0.11$ \\
\bottomrule
\end{tabular}}
\caption{\textbf{Associations of the recovered measures with questionnaire scores.}
Spearman correlations across participants of the measures of the router refit to synthetic behavior ($500$ steps pedestrian, $600$ steps highway), under the generating routers and the three contrasts; pedestrian entries are raw correlations.
The data row repeats the three-seed held-out values of \S\ref{sec:exp-clinical}. Contrasts (b) and (c) were run on the highway task only; highway contrast (c) gives the range over three permutations.}
\label{tab:recovery-traits}
\end{table}

\paragraph{Scope.}
The analysis tests the router stage with the per-depth policies fixed, and the generating agents use the model's own policies.
Choice noise that changes only near a vehicle can also move the recovered depth shift at a fixed depth.
Parameter recovery here covers the router fit with the per-depth policies fixed.
Recovery through the full pipeline with independently generated agents would separate the depth shift from state-dependent choice noise.

\section{Cohort-level depth shift and action profiles of the per-depth policies}\label{sec:appendix-depth-profile}

This appendix reports the cohort-level depth shift of the router on both tasks and the actions each per-depth policy $\pi^h$ produces on the pedestrian task (\S\ref{sec:exp-clinical}).
For every participant we freeze the four converged policies, evaluate them on that participant's held-out expert states (\S\ref{sec:setup-holdout}), and average the action probabilities $\pi^h(a|s)$ over states.
The action-profile analysis is teacher-forced on expert states, so it is independent of the router and of rollouts.
For the action profiles we report $N{=}52$ participants and two training seeds ($42$ and $7$).
Danger and safe states are those of Table~\ref{tab:router-robust-traits}.
Each participant contributes a median of $56.5$ held-out danger states (minimum $26$, maximum $96$).

The cohort-level depth shift below uses the $10{,}000$-step routers on the pedestrian task (cohort mean $-0.40$ at the $500$-step fit) and the $600$-step routers on the highway task.
On the pedestrian task the expected depth, computed over all expert states of a participant, is lower in danger than in safe states. The per-participant safe/danger ratio is $1.085 \pm 0.110$ at seed $42$, $1.078 \pm 0.113$ at seed $7$, and $1.116 \pm 0.101$ at seed $123$ ($1.091 \pm 0.092$ for the three-seed mean, $42$ of $52$ participants above $1$, paired Wilcoxon $p{<}.001$ in each case).
On the highway task the expected depth is higher in the nearest lead-vehicle gap quintile than without a lead vehicle in all three runs ($N{=}47$). The cohort means are $+0.33$, $+0.62$, and $+1.11$ for the unfixed, seed-$7$, and seed-$123$ runs (Wilcoxon $p{=}.23$, $.031$, and $1.1\times10^{-6}$) and $+0.69$ for their mean ($p{=}1.0\times10^{-4}$).

The association with motor impulsivity in \S\ref{sec:exp-clinical} enlarges this cohort-level shift rather than reversing it.
We split each cohort into thirds by BIS-11 motor score, breaking ties by participant order ($17$, $17$, and $18$ participants on the pedestrian task and $16$, $15$, and $16$ on the highway task).
Within each third we average the seed-mean $\Delta h$ of the $10{,}000$-step routers.
On the pedestrian task the mean $\Delta h$ is $-0.28$, $-0.41$, and $-0.57$ from the lowest to the highest third, and $11$ of $17$, $14$ of $17$, and $17$ of $18$ participants have a negative shift.
On the highway task it is $+0.40$, $+0.73$, and $+0.93$, and $10$ of $16$, $11$ of $15$, and $13$ of $16$ participants have a positive shift.
In both tasks the lowest third already shifts in the cohort's direction on average, and the shift grows with motor impulsivity.

Table~\ref{tab:depth-profile-danger} gives the mean action probabilities in danger states.
The NOTHING probability is $0.35$--$0.42$ for $h\in\{2,3\}$ and $0.22$--$0.25$ for $h\in\{5,10\}$.
The DOWN, RIGHT, and LEFT probabilities are $0.005$--$0.024$ for $h\in\{2,3\}$ and $0.044$--$0.095$ for $h\in\{5,10\}$.
The UP probability stays within $0.54$--$0.60$ at every depth.

\begin{table}[h]
  \centering
  \footnotesize
  \setlength{\tabcolsep}{5pt}
  \begin{tabular}{@{}lccccc@{}}
    \toprule
    Action & Expert & $h{=}2$ & $h{=}3$ & $h{=}5$ & $h{=}10$ \\
    \midrule
    NOTHING & $0.358$ & $0.347/0.374$ & $0.399/0.419$ & $0.239/0.250$ & $0.229/0.221$ \\
    UP      & $0.572$ & $0.602/0.580$ & $0.555/0.537$ & $0.586/0.586$ & $0.540/0.561$ \\
    DOWN    & $0.028$ & $0.007/0.005$ & $0.007/0.005$ & $0.059/0.054$ & $0.077/0.076$ \\
    RIGHT   & $0.025$ & $0.024/0.023$ & $0.024/0.023$ & $0.070/0.066$ & $0.095/0.084$ \\
    LEFT    & $0.017$ & $0.019/0.018$ & $0.015/0.018$ & $0.046/0.044$ & $0.060/0.057$ \\
    \midrule
    DOWN{+}RIGHT{+}LEFT & $0.070$ & $0.051/0.046$ & $0.046/0.045$ & $0.175/0.164$ & $0.231/0.218$ \\
    \bottomrule
  \end{tabular}
  \caption{\textbf{Deeper per-depth policies wait less and move back or sideways more in vehicle-present states (pedestrian task, $N{=}52$).}
  Each cell is the participant mean of the state-averaged action probability $\pi^h(a|s)$ on held-out danger states, reported as seed $42$ / seed $7$.
  The Expert column is the empirical action frequency in the same states and is identical for the two seeds.
  The last row sums the three back and lateral actions.}
  \label{tab:depth-profile-danger}
\end{table}

Table~\ref{tab:depth-profile-contrast} tests the contrast between the deep pair and the shallow pair within participants.
For each participant and action we take the mean probability over $h\in\{5,10\}$ minus the mean probability over $h\in\{2,3\}$.
The NOTHING contrast is $-0.139$ at seed $42$ and $-0.160$ at seed $7$, and it is negative for $50$ and $49$ of $52$ participants.
The DOWN, RIGHT, and LEFT contrasts are positive for $46$ to $52$ of $52$ participants in both seeds.
The UP contrast changes sign between seeds ($-0.016$ and $+0.015$), with two-sided $p{=}.055$ and $p{=}.423$.

\begin{table}[h]
  \centering
  \footnotesize
  \setlength{\tabcolsep}{5pt}
  \begin{tabular}{@{}llccccc@{}}
    \toprule
    Seed & Action & Shallow & Deep & Deep $-$ shallow & Positive & Wilcoxon $p$ \\
    \midrule
    $42$ & NOTHING & $0.373$ & $0.234$ & $-0.139$ & $2/52$  & $2.2\times10^{-9}$ \\
         & UP      & $0.579$ & $0.563$ & $-0.016$ & $20/52$ & $.055$ \\
         & DOWN    & $0.007$ & $0.068$ & $+0.061$ & $51/52$ & $4.4\times10^{-10}$ \\
         & RIGHT   & $0.024$ & $0.082$ & $+0.058$ & $50/52$ & $2.9\times10^{-9}$ \\
         & LEFT    & $0.017$ & $0.053$ & $+0.036$ & $50/52$ & $2.4\times10^{-8}$ \\
    \midrule
    $7$  & NOTHING & $0.396$ & $0.236$ & $-0.160$ & $3/52$  & $7.9\times10^{-10}$ \\
         & UP      & $0.558$ & $0.574$ & $+0.015$ & $25/52$ & $.423$ \\
         & DOWN    & $0.005$ & $0.065$ & $+0.060$ & $52/52$ & $3.5\times10^{-10}$ \\
         & RIGHT   & $0.023$ & $0.075$ & $+0.052$ & $49/52$ & $8.4\times10^{-10}$ \\
         & LEFT    & $0.018$ & $0.051$ & $+0.033$ & $46/52$ & $7.0\times10^{-8}$ \\
    \bottomrule
  \end{tabular}
  \caption{\textbf{The deep-minus-shallow contrast has the same sign for at least $46$ of $52$ participants on four of five actions, in both seeds (held-out danger states).}
  Shallow and Deep are participant means of the action probability averaged over $h\in\{2,3\}$ and over $h\in\{5,10\}$.
  Deep $-$ shallow is the mean within-participant difference.
  Positive counts participants with a difference above zero, and every remaining participant has a difference below zero.
  $p$ values are raw values from two-sided Wilcoxon signed-rank tests across participants, and the largest of the eight NOTHING, DOWN, RIGHT, and LEFT values is $7.0\times10^{-8}$.}
  \label{tab:depth-profile-contrast}
\end{table}

Three observations bound the reading of these tables.
First, the NOTHING contrast is specific to vehicle-present states.
In held-out states with all adjacent cells free of vehicles, the NOTHING contrast is positive, at $+0.066$ for seed $42$ ($47$ of $52$ participants) and $+0.044$ for seed $7$ ($42$ of $52$).
Restricting further to on-road states gives $+0.056$ ($49$ of $52$) and $+0.033$ ($42$ of $52$).
In both of these state sets the deeper policies move probability away from UP ($-0.165$ and $-0.150$ in vehicle-free states, $-0.116$ and $-0.096$ in vehicle-free on-road states), and the DOWN, RIGHT, and LEFT contrasts remain positive ($+0.011$ to $+0.041$).
The back and lateral contrast therefore holds in vehicle-present and vehicle-free states, and the waiting contrast reverses with vehicle presence.
Second, the deeper policies place more probability on back and lateral moves than the experts do.
In danger states their summed DOWN, RIGHT, and LEFT probability exceeds the expert value (last row of Table~\ref{tab:depth-profile-danger}), and the expert NOTHING rate lies between the deep-pair and shallow-pair means (Table~\ref{tab:depth-profile-contrast}).
Third, the same directions appear when the training states are added to the held-out states (median $814$ danger states per participant, minimum $384$).
There the NOTHING contrast is $-0.147$ and $-0.164$, positive for $1$ and $0$ of $52$ participants, and the DOWN, RIGHT, and LEFT contrasts are $+0.041$ to $+0.072$, positive for $50$ to $52$ of $52$ participants.

\section{Broader Impact}
\label{sec:appendix-broader-impact}

\paragraph{Positive applications.}
SC-AIRL aims to predict individual human decisions with higher held-out likelihood than Standard AIRL and to capture individual differences in how people plan.
We expect the approach to extend from naturalistic laboratory tasks to real-world decision tasks, although this remains to be tested (see \emph{Simulator-only validation} below).

\textbf{Cognitive and behavioral science.}
SC-AIRL agents fitted to individual participants can serve as simulation models for studying bounded foresight across environments and populations.
The associations of depth variability with anxiety and of depth shift with motor impulsivity (§\ref{sec:exp-clinical}) indicate that the inferred depth can carry information about individual traits, so the allocation of planning depth and its shift near hazards can serve as behavior-based individual-difference measures.

\textbf{Clinical assessment.}
Planning-depth estimates could complement self-report anxiety and impulsivity scales, which rest on what respondents report about themselves.
They could serve as behavior-based proxy measures alongside self-report scales where short cognitive tasks are already administered (e.g., neuropsychological assessment batteries).

\textbf{Decision-support applications.}
Potential downstream domains include pedestrian intent modeling for autonomous vehicles and personalized decision support, both of which need models that predict what a particular person will do rather than what an optimal agent would do.
More broadly, SC-AIRL uses inverse reinforcement learning to build models that behave like humans rather than models that behave optimally, which may help narrow the gap between computational models of decision-making and human behavior.

\paragraph{Risks and responsible deployment.}
Several constraints bound the responsible use of these applications.

\textbf{Simulator-only validation.} All results derive from two simulator tasks. Generalization to clinical populations, other task domains, or real-world deployments is unvalidated. Depth estimates are not screening instruments and should not be used as diagnostic markers without independent replication in the target domain.

\textbf{Fairness and group differences.}
Correlations with BIS and STAI are population-level associations in convenience samples ($N{=}52$ adults recruited through a public advertisement and the $N{=}47$ participants of the public dataset of \citet{lee2024bridging}).
The model does not account for demographic variation in planning style.
Deployment in contexts with differential representation risks amplifying existing disparities in clinical screening or AV pedestrian modeling.

\end{document}